\documentclass{article}

\usepackage{etoolbox}

\PassOptionsToPackage{dvipsnames}{xcolor}

\newcommand{\loose}{\looseness=-1}

\usepackage[utf8]{inputenc} %
\usepackage[T1]{fontenc}    %
\usepackage{url}            %
\usepackage{booktabs}       %
\usepackage{amsfonts}       %
\usepackage{nicefrac}       %
\usepackage{microtype}      %

\usepackage{tocloft}            %

\usepackage{enumitem}
\usepackage{breakcites}

\usepackage{wrapfig}
\usepackage{placeins}
\newcommand{\fitpaperfigure}[1]{#1}

\usepackage{mathrsfs}

\usepackage{algorithm}
\usepackage{verbatim}
\usepackage[noend]{algpseudocode}

\usepackage{multicol}
\usepackage{siunitx}
\usepackage{colortbl}
\usepackage{xcolor}
\definecolor{hl}{RGB}{245,245,245}
\newcolumntype{H}{>{\columncolor{hl}}} %

\definecolor{sd}{RGB}{232, 233, 251}
\definecolor{fm}{RGB}{252, 232, 216}

\usepackage{setspace}

\usepackage{transparent}

\usepackage{inconsolata}
\usepackage[scaled=.90]{helvet}
\usepackage{xspace}
\usepackage{tcolorbox}
\tcbuselibrary{breakable}

\usepackage{pifont}

\usepackage[a4paper, margin=2.5cm]{geometry}
\PassOptionsToPackage{hypertexnames=false}{hyperref}  %
\usepackage[dvipsnames]{xcolor}
\usepackage[colorlinks]{hyperref}
\hypersetup{
    citecolor=[RGB]{50,100,170},
    linkcolor=[RGB]{50,100,170},
    urlcolor=[RGB]{255,102,178}}
\usepackage{fancyhdr}

\newcommand{\toptitlebar}{%
  {\color{black}\hrule height 1pt}%
  \vskip 0.25in%
}

\makeatletter
\renewcommand{\maketitle}{%
  \begin{center}%
    \toptitlebar
    \vskip 0.1in%
    {\LARGE\bfseries \@title \par}%
    \vskip 0.3in%
    {\large \@author \par}%
  \end{center}%
  \par
  \vskip 0.3in%
}

\renewcommand\section{\@startsection {section}{1}{\z@}{-2.0ex plus
    -0.5ex minus -.2ex}{1.5ex plus 0.3ex minus .2ex}{\large\bfseries\raggedright}}
\renewcommand\subsection{\@startsection{subsection}{2}{\z@}{-1.8ex plus
    -0.5ex minus -.2ex}{0.8ex plus .2ex}{\normalsize\bfseries\raggedright}}
\renewcommand\subsubsection{\@startsection{subsubsection}{3}{\z@}{-1.5ex plus
   -0.5ex minus -.2ex}{0.5ex plus .2ex}{\normalsize\bfseries\raggedright}}
\renewcommand\paragraph{\@startsection{paragraph}{4}{\z@}%
   {1ex plus .2ex minus .1ex}{-.5em}{\normalfont\normalsize\bfseries}}

\renewenvironment{abstract}%
  {\centerline{\large\bfseries Abstract}%
   \begin{list}{}%
      {\setlength{\rightmargin}{0.6cm}%
       \setlength{\leftmargin}{0.6cm}}%
    \item[]\ignorespaces}%
  {\unskip\end{list}}

\makeatother

\usepackage{microtype}
\usepackage{hhline}

\makeatletter
\newcommand{\neutralize}[1]{\expandafter\let\csname c@#1\endcsname\count@}
\makeatother

\usepackage{algorithm}

\usepackage{natbib}
\bibpunct{(}{)}{;}{a}{,}{,}

\usepackage{amsthm}
\usepackage{mathtools}
\usepackage{amsmath}
\usepackage{bbm}
\usepackage{amsfonts}
\usepackage{amssymb}

\usepackage{xpatch}

\usepackage{thmtools}
\usepackage{thm-restate}
\declaretheorem[name=Theorem,parent=section]{theorem}
\declaretheorem[name=Lemma,parent=section]{lemma}
\declaretheorem[name=Assumption, parent=section]{assumption}
\declaretheorem[name=Condition, parent=section]{condition}

\declaretheorem[name=Proposition, parent=section]{proposition}

\makeatletter
  \renewenvironment{proof}[1][Proof]%
  {%
   \par\noindent{\bfseries\upshape {#1.}\ }%
  }%
  {\qed\newline}
  \makeatother

\theoremstyle{definition}  %

\newtheorem{corollary}{Corollary}[section]

\theoremstyle{plain}
\newtheorem{definition}{Definition}[section]

\xpatchcmd{\proof}{\itshape}{\normalfont\proofnameformat}{}{}
\newcommand{\proofnameformat}{\bfseries}

\usepackage[nameinlink,capitalize]{cleveref}

\AddToHook{cmd/appendix/after}{%
  \crefalias{section}{appendix}%
  \crefalias{subsection}{subappendix}%
  \crefalias{subsubsection}{subsubappendix}%
}

\newcommand{\pref}[1]{\cref{#1}}

\renewcommand{\eqref}[1]{\texorpdfstring{\hyperref[#1]{(\ref*{#1})}}{(\ref*{#1})}}

\crefformat{equation}{#2Eq. (#1)#3}
\Crefformat{equation}{#2Eq. (#1)#3}

\Crefformat{figure}{#2Figure #1#3}
\Crefname{assumption}{Assumption}{Assumptions}
\Crefformat{assumption}{#2Assumption #1#3}
\Crefname{subsubsection}{Section}{Sections}
\crefformat{subsubsection}{#2Section #1#3}
\Crefformat{subsubsection}{#2Section #1#3}
\Crefname{alg}{Alg.}{Algs.}

\crefname{experimentquestion}{RQ}{RQs}
\Crefname{experimentquestion}{RQ}{RQs}
\crefformat{experimentquestion}{#2RQ#1#3}
\Crefformat{experimentquestion}{#2RQ#1#3}
\crefmultiformat{experimentquestion}{#2RQ#1#3}{ and~#2RQ#1#3}{, #2RQ#1#3}{, and~#2RQ#1#3}
\Crefmultiformat{experimentquestion}{#2RQ#1#3}{ and~#2RQ#1#3}{, #2RQ#1#3}{, and~#2RQ#1#3}

\usepackage{crossreftools}
\usepackage{xparse}

\ExplSyntaxOn
\DeclareDocumentCommand{\XDeclarePairedDelimiter}{mm}
 {
  \__egreg_delimiter_clear_keys: %
  \keys_set:nn { egreg/delimiters } { #2 }
  \use:x %
   {
    \exp_not:n {\NewDocumentCommand{#1}{sO{}m} }
     {
      \exp_not:n { \IfBooleanTF{##1} }
       {
        \exp_not:N \egreg_paired_delimiter_expand:nnnn
         { \exp_not:V \l_egreg_delimiter_left_tl }
         { \exp_not:V \l_egreg_delimiter_right_tl }
         { \exp_not:n { ##3 } }
         { \exp_not:V \l_egreg_delimiter_subscript_tl }
       }
       {
        \exp_not:N \egreg_paired_delimiter_fixed:nnnnn 
         { \exp_not:n { ##2 } }
         { \exp_not:V \l_egreg_delimiter_left_tl }
         { \exp_not:V \l_egreg_delimiter_right_tl }
         { \exp_not:n { ##3 } }
         { \exp_not:V \l_egreg_delimiter_subscript_tl }
       }
     }
   }
 }

\keys_define:nn { egreg/delimiters }
 {
  left      .tl_set:N = \l_egreg_delimiter_left_tl,
  right     .tl_set:N = \l_egreg_delimiter_right_tl,
  subscript .tl_set:N = \l_egreg_delimiter_subscript_tl,
 }

\cs_new_protected:Npn \__egreg_delimiter_clear_keys:
 {
  \keys_set:nn { egreg/delimiters } { left=.,right=.,subscript={} }
 }

\cs_new_protected:Npn \egreg_paired_delimiter_expand:nnnn #1 #2 #3 #4
 {%
  \mathopen{}
  \mathclose\c_group_begin_token
   \left#1
   #3
   \group_insert_after:N \c_group_end_token
   \right#2
   \tl_if_empty:nF {#4} { \c_math_subscript_token {#4} }
 }
\cs_new_protected:Npn \egreg_paired_delimiter_fixed:nnnnn #1 #2 #3 #4 #5
 {
  \mathopen{#1#2}#4\mathclose{#1#3}
  \tl_if_empty:nF {#5} { \c_math_subscript_token {#5} }
 }
\ExplSyntaxOff

\XDeclarePairedDelimiter{\supnorm}{
  left=\lVert,
  right=\rVert,
  subscript=\infty
  }

\newcommand{\KL}{\mathrm{KL}}%

\newcommand{\grpo}{\textsc{GRPO}\xspace}
\newcommand{\maxrl}{\textsc{MaxRL}\xspace}
\newcommand{\tailrl}{\textsc{TailRL}\xspace}
\newcommand{\reinforce}{\textsc{REINFORCE}\xspace}

\newcommand{\ppo}{\textsc{PPO}\xspace}

\renewcommand{\epsilon}{\varepsilon}

\DeclarePairedDelimiter{\brk}{[}{]}

\DeclarePairedDelimiter{\prn}{(}{)}

\DeclarePairedDelimiter{\floor}{\lfloor}{\rfloor}

\let\Pr\undefined

\DeclareMathOperator{\Pr}{Pr}

\def\ddefloop#1{\ifx\ddefloop#1\else\ddef{#1}\expandafter\ddefloop\fi}
\def\ddef#1{\expandafter\def\csname bb#1\endcsname{\ensuremath{\mathbb{#1}}}}
\ddefloop ABCDEFGHIJKLMNOPQRSTUVWXYZ\ddefloop
\def\ddefloop#1{\ifx\ddefloop#1\else\ddef{#1}\expandafter\ddefloop\fi}
\def\ddef#1{\expandafter\def\csname b#1\endcsname{\ensuremath{\mathbf{#1}}}}
\ddefloop ABCDEFGHIJKLMNOPQRSTUVWXYZ\ddefloop
\def\ddef#1{\expandafter\def\csname sf#1\endcsname{\ensuremath{\mathsf{#1}}}}
\ddefloop ABCDEFGHIJKLMNOPQRSTUVWXYZ\ddefloop
\def\ddef#1{\expandafter\def\csname c#1\endcsname{\ensuremath{\mathcal{#1}}}}
\ddefloop ABCDEFGHIJKLMNOPQRSTUVWXYZ\ddefloop
\def\ddef#1{\expandafter\def\csname h#1\endcsname{\ensuremath{\widehat{#1}}}}
\ddefloop ABCDEFGHIJKLMNOPQRSTUVWXYZ\ddefloop
\def\ddef#1{\expandafter\def\csname hc#1\endcsname{\ensuremath{\widehat{\mathcal{#1}}}}}
\ddefloop ABCDEFGHIJKLMNOPQRSTUVWXYZ\ddefloop
\def\ddef#1{\expandafter\def\csname t#1\endcsname{\ensuremath{\widetilde{#1}}}}
\ddefloop ABCDEFGHIJKLMNOPQRSTUVWXYZ\ddefloop
\def\ddef#1{\expandafter\def\csname tc#1\endcsname{\ensuremath{\widetilde{\mathcal{#1}}}}}
\ddefloop ABCDEFGHIJKLMNOPQRSTUVWXYZ\ddefloop
\def\ddefloop#1{\ifx\ddefloop#1\else\ddef{#1}\expandafter\ddefloop\fi}
\def\ddef#1{\expandafter\def\csname scr#1\endcsname{\ensuremath{\mathscr{#1}}}}
\ddefloop ABCDEFGHIJKLMNOPQRSTUVWXYZ\ddefloop

\newcommand{\ldef}{\vcentcolon=}

\makeatletter
\let\OldStatex\Statex
\renewcommand{\Statex}[1][3]{%
  \setlength\@tempdima{\algorithmicindent}%
  \OldStatex\hskip\dimexpr#1\@tempdima\relax}
\makeatother

\usepackage{accents}
\usepackage{tikz}
\usetikzlibrary{decorations.pathreplacing,positioning,arrows.meta,calc,fit}
\definecolor{PaperUnified}{HTML}{2A78D6}
\definecolor{PaperStaged}{HTML}{777777}
\definecolor{PaperBase}{HTML}{A6ACB5}
\definecolor{PaperPretraining}{HTML}{DCE4ED}
\definecolor{PaperMidtraining}{HTML}{F0B66E}
\definecolor{PaperInk}{HTML}{252A31}
\definecolor{PaperSecondary}{HTML}{555B63}
\definecolor{PaperBorder}{HTML}{B9BEC6}
\newcommand{\paperfigfont}{\fontfamily{DejaVuSans-TLF}\fontsize{8bp}{9.6bp}\selectfont}
\newcommand{\paperfigsmall}{\fontfamily{DejaVuSans-TLF}\fontsize{7bp}{8.4bp}\selectfont}
\newcommand{\paperfigtiny}{\fontfamily{DejaVuSans-TLF}\fontsize{6.5bp}{7.8bp}\selectfont}
\tikzset{
    paper figure/.style={font=\paperfigfont, text=PaperInk, line width=0.65pt},
    paper legend/.style={draw=PaperBorder, fill=white, line width=0.6pt},
    paper arrow/.style={-{Latex[length=1.6mm]}, draw=PaperSecondary, line width=0.9pt},
    paper label/.style={font=\paperfigsmall, text=PaperInk, fill=white, inner sep=1.5pt},
}

\usepackage{varwidth}

\addtocontents{toc}{\protect\setcounter{tocdepth}{0}}

\makeatletter
\newif\if@anycite
\AtBeginDocument{%
  \pretocmd{\@citex}{\global\@anycitetrue}{}{}%
  \pretocmd{\nocite}{\global\@anycitetrue}{}{}%
}
\newcommand{\safebibliography}[1]{\if@anycite\bibliography{#1}\fi}
\makeatother

\let\oldparagraph\paragraph
\renewcommand{\paragraph}[1]{%
        \oldparagraph{#1.}%
}

\algrenewcommand\algorithmicrequire{\textbf{require}}

\usepackage{yfonts}

\title{Training Foundation Models with a Single Algorithm}
    \newcommand{\paperauthors}{%
      {\fontsize{10.95}{15}\selectfont
      \textbf{Yuda Song}$^{*,\dagger,1}$\qquad
      \textbf{Alvin Zhang}\textsuperscript{*,2}\qquad
      \textbf{Fahim Tajwar}\textsuperscript{*,1}\\[2pt]
      \textbf{Daman Arora}\textsuperscript{1}\quad
      \textbf{Jeff Schneider}\textsuperscript{1}\quad
      \textbf{Daniel Khashabi}\textsuperscript{2}\quad
      \textbf{Ruslan Salakhutdinov}\textsuperscript{1}\\[2pt]
      \textbf{J. Andrew Bagnell}$^{\ddag,1,3}$\qquad
      \textbf{Aarti Singh}$^{\ddag,1}$\qquad
      \textbf{Andrea Zanette}$^{\ddag,1}$\par}
      \vspace{6pt}
      {\normalsize
      \textsuperscript{1}\,Carnegie Mellon University\quad
      \textsuperscript{2}\,Johns Hopkins University\quad
      \textsuperscript{3}\,Aurora Innovation\par}
      \vspace{4pt}
      {\small\ttfamily
      \{yudas,ftajwar,aarti,azanette\}@andrew.cmu.edu\quad
      bzhang90@jh.edu\quad dbagnell@aurora.tech\par}
    }
    \author{\paperauthors}

\begin{document}

\maketitle
\renewcommand{\thefootnote}{}
\footnotetext{$^*$Equal contribution, $^\dagger$project lead, $^\ddag$equal advising.}

\thispagestyle{fancy}
\fancyhead{}
\lhead{\raisebox{-0.7cm}{\includegraphics[height=0.52cm]{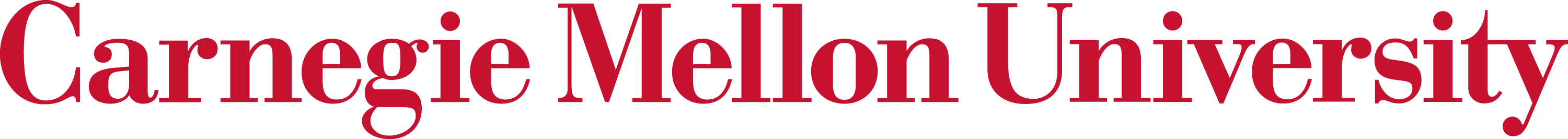}}}
\renewcommand{\headrulewidth}{0pt}
\setlength{\headheight}{29pt} %
\setlength{\headsep}{3mm}
 
\begin{abstract}

Foundation models are typically trained in stages: pretraining, midtraining, supervised fine-tuning (SFT), and reinforcement learning (RL), with changes in both data and learning algorithms. We show that this algorithmic separation is not fundamental. By deriving suitable RL objectives as maximum-likelihood fitting to model-generated target distributions, we place all of these stages under a single likelihood-based training algorithm, \emph{unified training}. This perspective turns the foundation-model training lifecycle into a problem of choosing which data to train on, in what proportions, and when, with conventional staged training as a special case. We instantiate this framework in a unified training system that mixes static data and RL generations within a single trainer. Across controlled and real-world experiments, unified training improves the accuracy--compute tradeoff and task coverage while preserving broader capabilities, and provides a stronger starting point for continual learning. This work lays the groundwork for a unified science of foundation-model training, connecting mathematical principles, training infrastructure, and the development of model capabilities.

\end{abstract}

\par
\begingroup
\centering\normalsize\sffamily\bfseries
\hypersetup{urlcolor=[RGB]{50,100,170}}
\href{https://yudasong.github.io/unified-training/}{Website}
\hspace{1.2em}{\normalfont\color{black!35}\textbullet}\hspace{1.2em}%
\href{https://github.com/yudasong/unified-training}{GitHub}
\hspace{1.2em}{\normalfont\color{black!35}\textbullet}\hspace{1.2em}%
\href{https://huggingface.co/sfanm/unified-training}{Hugging Face}
\par
\endgroup

\suppressfloats[t]
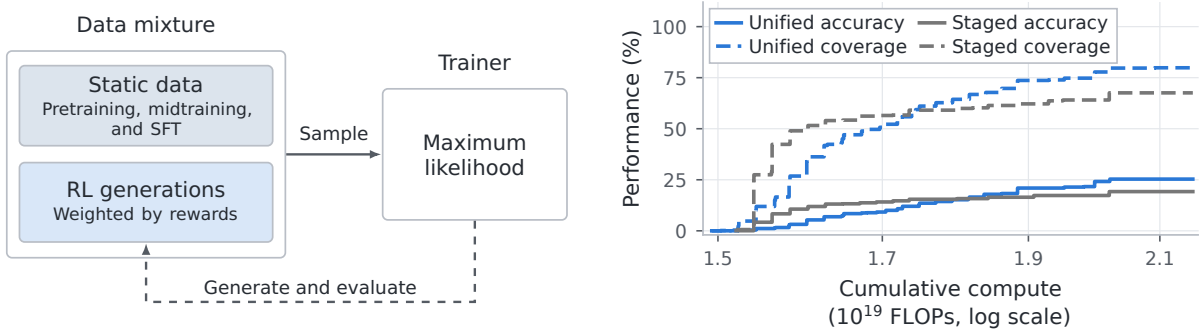
\begin{figure}[t]
        \centering
    \begin{minipage}[c]{0.49\linewidth}
        \centering
        \resizebox{\linewidth}{!}{\begin{tikzpicture}[
    paper figure,
    box/.style={draw=PaperBorder, rounded corners=2pt, fill=white,
        align=center, inner sep=4pt},
    source/.style={box, text width=3.10cm, minimum height=1.06cm},
]
    \path[use as bounding box] (0,0) rectangle (7.80,4.445);
    \node[box, minimum width=3.72cm, minimum height=2.76cm]
        (mixture) at (1.99,2.42) {};
    \node[above=3pt of mixture.north] {Data mixture};
    \node[source, fill=PaperPretraining] (static) at (1.99,3.07)
        {Static data\\[-1pt]
        {\paperfigtiny Pretraining, midtraining,}\\[-2pt]
        {\paperfigtiny and SFT}};
    \node[source, fill=PaperUnified!18] (rl) at (1.99,1.78)
        {RL generations\\[-1pt]
        {\paperfigtiny Weighted by rewards}};
    \node[box, text width=2.18cm, minimum height=1.75cm]
        (trainer) at (6.38,2.42)
        {Maximum\\likelihood};
    \node[above=3pt of trainer.north] {Trainer};
    \draw[paper arrow] (mixture.east)
        -- node[paper label, above=2pt] {Sample} (trainer.west);
    \draw[paper arrow, dashed] (trainer.south)
        -- (6.38,0.45)
        -- node[paper label, above=1pt] {Generate and evaluate} (1.99,0.45)
        -- (rl.south);
\end{tikzpicture}}
    \end{minipage}\hfill
    \begin{minipage}[c]{0.49\linewidth}
        \centering
                \includegraphics[width=\linewidth]{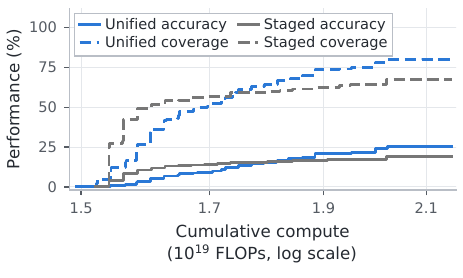}
    \end{minipage}\par
    \caption{\textbf{Unified training framework.}
    \textbf{Left:} For suitable RL objectives, static data and RL generations share a weighted likelihood loss; the updated model supplies new generations that are added to the data mixture.
    \textbf{Right:} Unified training reaches higher accuracy and coverage on a reasoning task within the same total compute budget.
Each curve shows the best observed accuracy (solid) or coverage (dashed) attainable within a compute budget, across all tested unified (blue) or staged (gray) configurations.
Compute includes pretraining, midtraining, and RL.
\pref{fig:e2e-compute} shows the individual training trajectories. \loose}
    \label{fig:teaser}
\end{figure}

\section{Introduction}
\label{sec:intro}

Foundation models are conventionally trained through a sequence of stages: pretraining, midtraining, supervised fine-tuning (SFT), and reinforcement learning (RL)~\citep{deepseek2025v3,olmo2025olmo3}. These stages have become embedded in specialized training recipes and infrastructure, and are often treated as separate learning problems. In this paper, we argue that this separation is not fundamental to the training process. Instead, the full training pipeline admits a strikingly simple view: \loose
\begin{center}
\emph{Foundation-model training can be viewed as a single learning objective \\ driven by an evolving distribution of training data.}
\end{center}

This shift changes the central design problem. Conventional pipelines organize design decisions around separate stages, each with its own training recipe.
Under a unified view, the learning algorithm stays fixed from start to end, and the central design choices concern the evolving distribution of training data: which data the model learns from, in what proportions, and when.
Pretraining text, demonstrations, and model generations selected with verifiable rewards become different data sources for the same underlying learning process.
Conventional staged training is then one particular choice, in which the proportion of each data source abruptly shifts from one stage to the next.

The unified view also changes the scientific question. Our understanding of foundation-model training has historically developed one stage at a time: pretraining scaling laws characterize the relationship between data, compute, and model performance~\citep{kaplan2020scaling,hoffmann2022computeoptimal}, while a growing literature studies scaling during RL~\citep{nimmaturi2025predictive,khatri2026art,tan2026rlscaling}.
Research on transfer and interactions across stages shows that earlier training decisions also affect subsequent learning~\citep{hernandez2021transfer,isik2025downstream,runwal2026prism,shen2026understanding}.
As a result, we know much less about how decisions made throughout training jointly determine the capabilities of the final model. A unified formulation lets us study these decisions together and judge them by their effect on the final model rather than by stage-specific metrics. This opens the door to a science of training trajectories: principles for how data and compute should evolve over the lifetime of a foundation model.

Making this view concrete requires bringing RL under the maximum-likelihood formulation already used for pretraining, midtraining, and SFT. Building on Maximum Likelihood Reinforcement Learning (\maxrl)~\citep{tajwar2026maxrl}, we extend its likelihood interpretation to arbitrary fixed generation policies and show how the resulting targets can be combined with static data in a single likelihood objective.
This provides the theoretical foundation for \emph{unified training},
a framework for learning from evolving mixtures of static and generated
data under a common training rule, including when generation and
optimization proceed independently. We implement this formulation through a common data representation, weighted likelihood loss, and trainer for both static and generated data.
\pref{fig:teaser} visualizes the unified training loop and shows that unified training can improve performance over staged training within the same total compute budget.
Concretely, our contributions are as follows: \loose

\begin{itemize}[leftmargin=*,itemsep=5pt,topsep=4pt]
    \item \textbf{A unified formulation of foundation-model training.} We establish a common maximum likelihood formulation for pretraining, midtraining, SFT, and suitable RL objectives. This provides a theoretical basis for learning from changing mixtures of static and generated data, with staged training as a special case. The formulation also extends to other RL objectives and continuous rewards (\pref{sec:framework}), and it supports a unified training infrastructure (\pref{sec:implementation}). \loose
    \item \textbf{Empirical benefits across training settings.} Through controlled and real-world experiments, we demonstrate improvements of unified training over staged training in accuracy and coverage. We also show that unified training can preserve broader capabilities (\pref{sec:experiments}).
    \item \textbf{Data mixing and compute allocation.} We characterize how mixing proportions and schedules shape the tradeoffs among accuracy, coverage, and retention, and study compute allocation across pretraining, midtraining, and RL (\pref{sec:experiments}).
\end{itemize}

By placing data composition, timing, and compute allocation within a common framework, unified training provides a basis for developing principles that govern the full training process. We believe it has both scientific and practical value as a starting point for studying foundation-model training as a whole.

\section{Preliminaries}
\label{sec:prelim}

We write $\pi_\theta(y\mid x)$ for the model's distribution over completions $y$ given a prompt $x$, with parameters $\theta$.

\subsection{Maximum Likelihood Estimation}
\label{sec:prelim-mle}
Maximum likelihood estimation (MLE) fits the model to training examples by minimizing their negative log-likelihood.
We represent each example as a prompt-completion pair $(x,y)$, where the prompt may be empty.
Let $\rho(x)$ denote the prompt distribution and $\mu(y\mid x)$ the conditional completion distribution in the training data.
Their joint distribution is denoted by $\rho\circ\mu$, with $(\rho\circ\mu)(x,y)\ldef\rho(x)\mu(y\mid x)$.
MLE minimizes the loss
\begin{align*}
    \cL(\theta;\rho\circ\mu)
    \ldef -\bbE_{x\sim\rho}\,\bbE_{y\sim\mu(\cdot\mid x)}\brk*{\log\pi_\theta(y\mid x)}.
\end{align*}

Pretraining, midtraining, and SFT all use this objective with different training data.
Pretraining and midtraining fit text from general and curated corpora, respectively, using empty prompts; SFT fits demonstrations paired with instructions. 
We refer to the fixed corpora and demonstrations used in these stages as \emph{static data}.

For a fixed training data distribution, this loss differs from the expected forward KL divergence $\bbE_{x\sim\rho}[\KL(\mu(\cdot\mid x)\,\|\,\pi_\theta(\cdot\mid x))]$ only by a constant independent of $\theta$.
Both objectives therefore have the same gradient:
\begin{align*}
    \nabla_\theta\cL(\theta;\rho\circ\mu)
    = -\bbE_{x\sim\rho}\,\bbE_{y\sim\mu(\cdot\mid x)}\brk*{\nabla_\theta\log\pi_\theta(y\mid x)}.
\end{align*}

\subsection{Reinforcement Learning}
\label{sec:prelim-rl}
Reinforcement learning (RL) uses reward feedback to optimize model-generated completions.
Given a prompt $x\sim\rho$, the model samples a completion $y\sim\pi_\theta(\cdot\mid x)$ and receives a reward $r(x,y) \in\{0,1\}$.\footnote{We begin with binary rewards for clarity. \pref{sec:unified} discusses continuous rewards through \tailrl~\citep{ramasubramanian2026tailrl}; \pref{app:continuous-rewards} gives details.}
The expected-reward objective is
\begin{align*}
    J(\theta) \ldef \bbE_{x\sim\rho}\,\bbE_{y\sim\pi_\theta(\cdot\mid x)}\brk*{r(x,y)},
\end{align*}
with policy gradient:
\begin{align}
    \nabla_\theta J(\theta)
    = \bbE_{x\sim\rho}\,\bbE_{y\sim\pi_\theta(\cdot\mid x)}\brk*{r(x,y)\,\nabla_\theta\log\pi_\theta(y\mid x)}.
    \label{eq:policy-gradient}
\end{align}

\reinforce~\citep{williams1992reinforce} estimates this expectation from sampled completions.
Other methods, including \ppo~\citep{schulman2017ppo} and \grpo~\citep{shao2024deepseekmath}, introduce choices of advantage estimation, reward normalization, and constraints on policy updates.

\section{A Unified Likelihood View of Foundation-Model Training}
\label{sec:framework}

Pretraining, midtraining, and SFT already share a common maximum-likelihood training rule, differing primarily in the data on which the model is trained, and when to apply each data. 
At first glance, RL seems to break this pattern, as it is typically formulated as policy optimization. To unify these stages, we
build on the likelihood interpretation of \maxrl~\citep{tajwar2026maxrl}
to develop a common framework for these stages that also accommodates
other generation policies, RL objectives, and continuous rewards.

\subsection{\maxrl as Likelihood Fitting}
\label{sec:maxrl}

For binary rewards, let $p_\theta(x)\ldef\bbE_{y\sim\pi_\theta(\cdot\mid x)}[r(x,y)]$
be the probability of success on prompt $x$, and let
$\pi_\theta^+(y\mid x)\ldef\pi_\theta(y\mid x,r=1)$ be the success-conditioned distribution.
\citet{tajwar2026maxrl} show that the \maxrl gradient can be written as
\[
\nabla_\theta \log p_\theta(x)
=
\bbE_{y\sim\pi_\theta^+(\cdot\mid x)}
\left[
\nabla_\theta \log \pi_\theta(y\mid x)
\right].
\]
\par
Thus, at iteration $t$, a \maxrl update can be viewed as maximum-likelihood
fitting to the current policy's success-conditioned distribution
$\pi_{\theta_t}^+$, held fixed during the update:
\[
\cL(\theta;\rho\circ\pi_{\theta_t}^+)
=
-\bbE_{x\sim\rho}
\bbE_{y\sim\pi_{\theta_t}^+(\cdot\mid x)}
\left[
\log \pi_\theta(y\mid x)
\right].
\]
\par
This places RL in the same likelihood-fitting form as pretraining,
midtraining, and SFT.

\subsection{Unified Training}
\label{sec:unified-training}
\label{sec:unified}

Pretraining, midtraining, SFT, and \maxrl can therefore all be expressed as
likelihood fitting. We represent each source $s$ at iteration $t$ by a prompt
distribution $\rho_{s,t}$ and target distribution $\mu_{s,t}$, as summarized
in \pref{tab:stages}.

Let $\lambda_{s,t}\geq 0$, with $\sum_s\lambda_{s,t}=1$, denote the
mixture weight of source $s$ at iteration $t$. Together, the sources define
a single joint distribution over training examples,
\[
q_t
\ldef
\sum_s
\lambda_{s,t}(\rho_{s,t}\circ\mu_{s,t}).
\]
Unified training applies the same likelihood loss to this mixture:
\begin{align*}
\cL(\theta;q_t) = -\bbE_{(x,y)\sim q_t}\brk*{\log \pi_\theta(y\mid x)}.
\end{align*}
\begin{wraptable}{r}{0.43\textwidth}
        \centering
    \small
    \setlength{\tabcolsep}{3pt}
    \begin{tabular*}{\linewidth}{@{\extracolsep{\fill}}lcc@{}}
        \toprule
        Source & Prompts $\rho_{s,t}$ & Target $\mu_{s,t}$ \\
        \midrule
        Pretraining & empty & $\mu_{\mathrm{pre}}$ \\
        Midtraining & empty & $\mu_{\mathrm{mid}}$ \\
        SFT & $\rho_{\mathrm{sft}}$ & $\mu_{\mathrm{sft}}$ \\
        RL & $\rho_{\mathrm{rl}}$ & $\pi^{+}_{\theta_t}$ or $\nu^+$ \\
        \bottomrule
    \end{tabular*}
    \caption{Data sources for the shared likelihood objective.
    For binary rewards, $\nu^+$ is the success-conditioned distribution of the generation policy $\nu$; $\nu=\pi_{\theta_t}$ recovers on-policy \maxrl.}
    \label{tab:stages}
\end{wraptable}

The source distributions and mixture weights are held fixed when differentiating the loss.
Equivalently, one can sample source $s$ with probability $\lambda_{s,t}$, draw an example
from that source, and apply the same likelihood loss.

Both the source distributions and their mixture weights may evolve throughout
training. The learning rule itself, however, remains unchanged: what changes
is the distribution of data presented to the model. Conventional staged
training is the special case in which all mixture weight is placed on one
source at a time; unified training additionally allows sources to overlap
and their proportions to vary throughout training.

The formulation above contains the core ingredients of unified training,
and we turn to its implementation in the next section. Before doing so, we
establish two extensions of the framework: it can incorporate generations
produced by policies other than the model currently being optimized, and it
can accommodate other policy-gradient objectives through different choices
of target and effective prompt distributions.

\paragraph{Likelihood targets from arbitrary generators}
The likelihood formulation does not require RL data to be generated by the
model currently being optimized. Suppose completions are generated by a policy
$\nu$, held fixed while constructing the training target. Under binary rewards,
its successful completions define
\[
\nu^+(\cdot\mid x)
\ldef
\nu(\cdot\mid x, r=1).
\]
The trainer can fit this distribution using the same maximum-likelihood loss,
\[
\cL(\theta;\rho\circ\nu^+)
=
-\bbE_{x\sim\rho}
 \bbE_{y\sim\nu^+(\cdot\mid x)}
 \left[\log \pi_\theta(y\mid x)\right].
\]
When $\nu=\pi_{\theta_t}$, this recovers the on-policy \maxrl update of
\pref{sec:maxrl}. For a different generator, the trainer instead fits
that generator's success-conditioned distribution. Thus, generation and
optimization can proceed independently while retaining the same likelihood
training rule; the resulting update, however, is generally not the
current-policy \maxrl gradient.

\paragraph{Other RL objectives}
Under binary rewards, \reinforce fits the same success-conditioned target
$\pi_{\theta_t}^+$ as \maxrl, but uses the effective prompt distribution
\[
\rho_{\mathrm{RF}}(x)
\propto
\rho(x)p_{\theta_t}(x).
\]
This gives greater weight to prompts with higher success probability, whereas \maxrl retains $\rho$.
The framework also accommodates rewards normalized to $[0,1]$, including
continuous-reward \reinforce and \tailrl~\citep{ramasubramanian2026tailrl}.
\pref{tab:rl-objectives} summarizes the targets and effective prompt distributions for on-policy updates; \pref{app:continuous-rewards} gives the derivations.%

\begin{table}[!ht]
    \centering
    \small
    \setlength{\tabcolsep}{5pt}
    \begin{tabular}{llll}
        \toprule
        Method & Reward & Target $\mu_s$ & Effective prompts $\rho_s$ \\
        \midrule
        \maxrl
        & Binary
        & $\pi_{\theta_t}^+$
        & $\rho(x)$ \\

        \reinforce
        & Binary
        & $\pi_{\theta_t}^+$
        & $\propto \rho(x)p_{\theta_t}(x)$ \\

        \tailrl
        & $[0,1]$
        & $\displaystyle\int_0^1
            \pi_{\theta_t}(\cdot\mid x,r>\tau)\,d\tau$
        & $\rho(x)$ \\

        \reinforce
        & $[0,1]$
        & $\displaystyle
            \frac{
                \pi_{\theta_t}(y\mid x)r(x,y)
            }{
                \bbE_{y'\sim\pi_{\theta_t}(\cdot\mid x)}[r(x,y')]
            }$
        & $\displaystyle
            \propto
            \rho(x)
            \bbE_{y\sim\pi_{\theta_t}(\cdot\mid x)}[r(x,y)]$ \\
        \bottomrule
    \end{tabular}
    \caption{RL objectives within the unified likelihood framework.
    Different objectives correspond to different target and effective prompt
    distributions.}
    \label{tab:rl-objectives}
\end{table}

\section{Practical Implementation}
\label{sec:implementation}

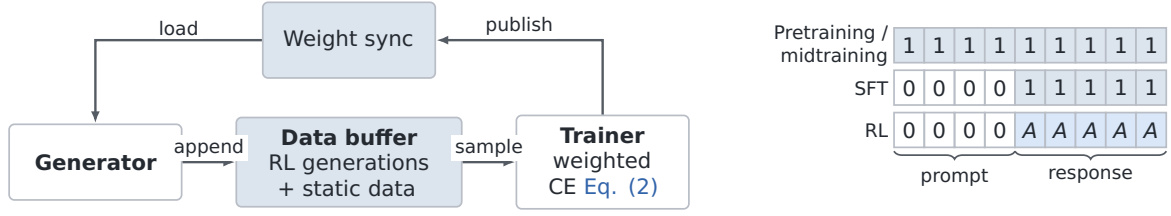
\begin{figure}[t]
    \centering
    \begin{minipage}[c]{0.60\linewidth}
        \centering
        \fitpaperfigure{\begin{tikzpicture}[
            paper figure,
            proc/.style={draw=PaperBorder, rounded corners=2pt, align=center, inner sep=4pt, minimum height=1.0cm, text width=2.0cm},
            store/.style={proc, fill=PaperPretraining},
            arr/.style={paper arrow},
            lab/.style={paper label},
            node distance=0.5cm and 0.7cm,
        ]
            \node[proc] (gen) {\textbf{Generator}};
            \node[store, right=of gen, text width=2.7cm] (buf) {\textbf{Data buffer}\\ \mbox{RL generations}\\ + static data};
            \node[proc, right=of buf] (trn) {\textbf{Trainer}\\ weighted CE \pref{eq:weighted-ce}};
            \node[store, above=of buf] (pol) {Weight sync};
            \draw[arr] (gen) -- node[lab, above]{append} (buf);
            \draw[arr] (buf) -- node[lab, above]{sample} (trn);
            \draw[arr] (trn.north) |- node[lab, pos=0.75, above]{publish} (pol.east);
            \draw[arr] (pol.west) -| node[lab, pos=0.25, above]{load} (gen.north);
        \end{tikzpicture}}
    \end{minipage}\hfill
    \begin{minipage}[c]{0.38\linewidth}
        \centering
        \fitpaperfigure{\begin{tikzpicture}[paper figure, cell/.style={draw=PaperBorder, minimum width=0.40cm, minimum height=0.46cm, inner sep=0pt}]
            \foreach \row/\name/\cells in {
                0/{Pretraining /\\midtraining}/{1,1,1,1,1,1,1,1,1},
                1/{SFT}/{0,0,0,0,1,1,1,1,1},
                2/{RL}/{0,0,0,0,A,A,A,A,A}} {
                \node[anchor=east, align=right, font=\paperfigsmall] at (-0.15, -0.56*\row) {\name};
                \foreach \w [count=\i from 0] in \cells {
                    \ifnum\pdfstrcmp{\w}{1}=0
                        \node[cell, fill=PaperPretraining] at (0.40*\i, -0.56*\row) {1};
                    \else\ifnum\pdfstrcmp{\w}{0}=0
                        \node[cell] at (0.40*\i, -0.56*\row) {0};
                    \else
                        \node[cell, fill=PaperUnified!18] at (0.40*\i, -0.56*\row) {\textit{A}};
                    \fi\fi
                }
            }
            \draw[draw=PaperSecondary, decorate, decoration={brace, mirror, amplitude=3pt}] (-0.20, -1.4) -- (1.40, -1.4) node[midway, below=3pt, font=\paperfigsmall]{prompt};
            \draw[draw=PaperSecondary, decorate, decoration={brace, mirror, amplitude=3pt}] (1.40, -1.4) -- (3.40, -1.4) node[midway, below=3pt, font=\paperfigsmall]{response};
        \end{tikzpicture}}
    \end{minipage}
    \caption{\textbf{Left:} the training system. The generator produces RL data from a recent copy of the weights and appends it to the data buffer, which also holds static data; the trainer samples from the buffer and periodically publishes its weights, which the generator loads. \textbf{Right:} the data format. Each cell is a token and the number is its weight in \pref{eq:weighted-ce}; the stages differ only in the weights.}
    \label{fig:system}
\end{figure}

Building on the theoretical foundation in \pref{sec:unified}, which allows static data and RL generations to be combined under a common likelihood objective, we implement a unified training system with a shared data format, loss, and trainer.
A generator supplies RL generations and their weights to a shared data buffer, which also holds static data.
Generation and optimization run asynchronously, with periodic synchronization of the model weights (\pref{fig:system}).

\subsection{Unified Loss and Data Format}
\label{sec:impl-loss}
Each training example is a pair $(z,w)$, where $z=(z_1,\dots,z_L)$ is a token sequence and $w\in\bbR^L_{\ge0}$ specifies the weights on its token negative log-likelihoods.
Pretraining and midtraining use unit weights on corpus tokens; SFT uses unit weights on demonstration tokens and zero weights on prompt tokens.
RL uses the same prompt mask and assigns each completion a response weight $A$, computed from its reward and the sampled pass rate (\pref{sec:impl-generator}).
These assignments give a common data format for all sources (\pref{fig:system}, right). In practice, we use a token mean over a batch $B$:
\begin{align}
    \hat\cL(\theta;B) \ldef -\,\frac{1}{T_B}\sum_{(z,w)\in B}\sum_{\ell=1}^{L} w_\ell\log\pi_\theta(z_\ell\mid z_{<\ell}),
    \label{eq:weighted-ce}
\end{align}
where $T_B$ counts the total number of non-prompt tokens in the batch.

\subsection{Data Generation}
\label{sec:impl-generator}
The generator samples $N$ completions $y_1,\dots,y_N\sim\pi_{\theta_s}(\cdot\mid x)$ per prompt $x$, where $\pi_{\theta_s}$ is a recent copy of the model.
For binary rewards, it estimates the pass rate by $\hat p\ldef\frac1N\sum_i r(x,y_i)$.
Each correct completion receives weight $A\ldef1/\hat p$ on its response tokens; incorrect completions receive zero.

This weighting is the finite-sample form of \maxrl.
Averaging the weighted score function over the $N$ samples for a prompt gives
\begin{align*}
    \frac{1}{N}\sum_{i=1}^{N} \frac{r(x,y_i)}{\hat p}\,\nabla_\theta\log\pi_\theta(y_i\mid x)
    = \frac{1}{N\hat p}\sum_{i:\,r(x,y_i)=1}\nabla_\theta\log\pi_\theta(y_i\mid x).
\end{align*}
This is the average score over the correct completions, with zero update when all completions are incorrect~\citep{tajwar2026maxrl}.
For generation policy $\nu$, its expectation is $-\nabla_\theta\cL(\theta;\rho\circ\nu^+)$ from \pref{sec:unified}, with prompt $x$ weighted by $1-(1-p_\nu(x))^N$, where $p_\nu(x)$ is the generation policy's success probability.
This weight is the probability that at least one of the $N$ samples is correct.
Finite $N$ therefore leaves the target $\nu^+$ unchanged and only down-weights prompts that the generating policy rarely solves; at $N=1$ the weight is $p_\nu(x)$, recovering \reinforce as in \pref{sec:unified}, and as $N\to\infty$ the weight tends to one.

\subsection{Data Mixing and Asynchronous Training}
\label{sec:impl-buffer}
The trainer implements the data mixture by sampling static data and RL generations from a shared buffer.
For a target token fraction $f_s$ from source $s\in S$, examples are sampled at a rate proportional to $f_s/T_s$, where $T_s$ is the mean number of corpus or response tokens per example.
The mixture can also be specified by the number of examples from each source.
\pref{app:data-mixing} collects the mixing and loss-weighting conventions used in the experiments.

Generation and optimization proceed asynchronously: generators append completed batches to the buffer, while the trainer samples from it and updates the model.
The trainer publishes its weights every $k$ updates, and generators load them between batches.
\pref{app:math-zero-mixing} reports the resulting policy staleness and compares training throughput.

\section{Experiments}
\label{sec:experiments}
The experiments aim to study the practical benefits of unified training, and its value for studying the full training process. Specifically, we examine how data composition and data schedule affect model capabilities, the use of data and compute, and continual learning:
\begin{enumerate}[leftmargin=*,itemsep=3pt,parsep=0pt,topsep=3pt]
    \item \label[experimentquestion]{q:performance}%
        \textbf{Performance and generalization.}
    How do unified and staged training compare in task accuracy, coverage, and generalization beyond the training distribution?
    \item \label[experimentquestion]{q:reuse}\label[experimentquestion]{q:mixing}\textbf{Data mixing and reuse.}
    How do data proportions, timing, and reuse of a fixed corpus affect performance during training and in the final model?
    \item \label[experimentquestion]{q:compute}\textbf{Compute allocation.}
        How does unified training compare with staged training under compute-matched comparisons?
    \item \label[experimentquestion]{q:retention}\textbf{Retention.}
    How do unified and staged training compare in preserving capabilities acquired earlier in training?
    \item \label[experimentquestion]{q:continual}\textbf{Continual learning.}
        Does unified training provide a better starting point for learning new tasks when subsequent training follows the same procedure for all models?
\end{enumerate}

We first compare unified and staged training across the full pipeline of pretraining, midtraining, and RL, in a controlled setting (\pref{sec:e2e}).
Then, with real-world heterogeneous midtraining data, we study the resulting capability tradeoffs, how different data mixture and scheduling can change model's performance, and their consequences for continual learning (\pref{sec:real}).
Unless otherwise specified, staged baselines use \maxrl~\citep{tajwar2026maxrl} for RL for a cleaner comparison against unified training. We defer the study of other RL objectives as future work with a preliminary analysis in \pref{app:math-weighting-details}. 

In the following sections, we sometimes still state the results in terms of stages (pretraining, midtraining, and RL) for clarity and cleaner comparison with staged training, but the unified formulation allows these stages to overlap.

\subsection{Full Unified Training under Controlled Reasoning Tasks}
\label{sec:e2e}

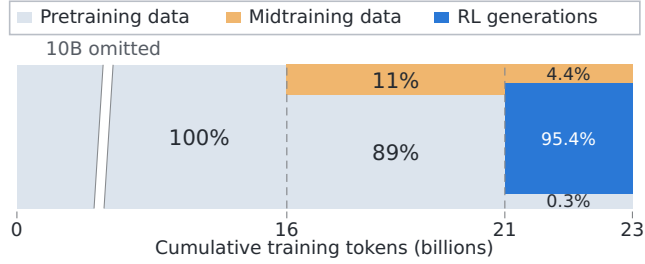
\begin{figure}[t]
    \centering
    \fitpaperfigure{\begin{tikzpicture}[
    x=0.60cm, y=1cm,
    paper figure,
    phase boundary/.style={draw=mixInk!55, densely dashed, line width=0.45pt},
    share/.style={font=\paperfigfont, inner sep=0pt},
    replay share/.style={share, font=\paperfigtiny},
]
    \colorlet{mixPre}{PaperPretraining}
    \colorlet{mixMid}{PaperMidtraining}
    \colorlet{mixRL}{PaperUnified}
    \colorlet{mixInk}{PaperInk}
    \tikzset{every node/.style={text=mixInk}}
    \def\chartHeight{1.9}
    \def\mixStart{5.94864}
    \def\rlStart{10.753936128}
    \def\chartWidth{13.565744459}
    \pgfmathsetmacro{\midBoundary}{\chartHeight-0.40}
    \def\replayBoundary{0.20}
    \pgfmathsetmacro{\demoBoundary}{\chartHeight-0.24}

    \fill[mixPre] (0,0) rectangle (\mixStart,\chartHeight);
    \fill[mixPre] (\mixStart,0) rectangle (\rlStart,\midBoundary);
    \fill[mixMid] (\mixStart,\midBoundary) rectangle (\rlStart,\chartHeight);
    \fill[mixPre] (\rlStart,0) rectangle (\chartWidth,\replayBoundary);
    \fill[mixRL] (\rlStart,\replayBoundary) rectangle (\chartWidth,\demoBoundary);
    \fill[mixMid] (\rlStart,\demoBoundary) rectangle (\chartWidth,\chartHeight);

    \fill[white] (1.70,-0.02) -- (1.92,-0.02)
        -- (2.12,1.92) -- (1.90,1.92) -- cycle;
    \draw[draw=mixInk!55, line width=0.4pt] (1.70,0) -- (1.90,\chartHeight);
    \draw[draw=mixInk!55, line width=0.4pt] (1.92,0) -- (2.12,\chartHeight);
    \node[font=\paperfigsmall, text=PaperSecondary, anchor=south, inner sep=0pt]
        at (1.91,2.02) {10B omitted};

    \draw[phase boundary] (\mixStart,0) -- (\mixStart,\chartHeight);
    \draw[phase boundary] (\rlStart,0) -- (\rlStart,\chartHeight);

    \draw[paper legend] (-0.16,2.30) rectangle (13.73,2.74);
    \foreach \x/\col/\lab in {0/mixPre/Pretraining data,
        4.60/mixMid/Midtraining data,9.20/mixRL/RL generations} {
        \fill[\col] (\x,2.45) rectangle +(0.33,0.14);
        \node[anchor=west, font=\paperfigsmall, inner sep=0pt] at (\x+0.49,2.52) {\lab};
    }

    \node[share] at (4.02,0.95) {100\%};
    \node[share] at ({(\mixStart+\rlStart)/2},{\midBoundary/2}) {89\%};
    \node[share] at ({(\mixStart+\rlStart)/2},{(\midBoundary+\chartHeight)/2}) {11\%};
    \node[replay share, text=white] at ({(\rlStart+\chartWidth)/2},{(\replayBoundary+\demoBoundary)/2}) {95.4\%};
    \node[replay share] at ({(\rlStart+\chartWidth)/2},{(\demoBoundary+\chartHeight)/2}) {4.4\%};
    \node[replay share]
        at ({(\rlStart+\chartWidth)/2},{\replayBoundary/2}) {0.3\%};

    \foreach \x/\lab in {0/0,\mixStart/16,\rlStart/21,\chartWidth/23} {
        \draw[draw=mixInk!60, line width=0.4pt] (\x,-0.04) -- (\x,-0.13);
        \node[anchor=north, font=\paperfigsmall, inner sep=0pt]
            at (\x,-0.17) {\lab};
    }
    \node[font=\paperfigsmall, inner sep=0pt]
        at ({\chartWidth/2},-0.53) {Cumulative training tokens (billions)};
\end{tikzpicture}}
    \caption{\textbf{Data composition throughout training.}
    The illustrated run from \pref{app:e2e-pretraining-replay} mixes midtraining tokens into pretraining, then combines RL generations, midtraining data, and pretraining data.
    Token counts include prompts and are rounded estimates; the axis break omits 10B pretraining tokens, and thin bands are enlarged for legibility.}
    \label{fig:e2e-mixing}
\end{figure}

\paragraph{Experimental setup}
We study how data composition and schedule affect the full training pipeline in a controlled setting.
We train a 124M-parameter model from scratch using a fixed 20B-token pretraining corpus and a fixed 510M-token midtraining corpus across runs.
Staged training uses these sources sequentially before RL; unified training mixes midtraining data into the final part of pretraining and throughout RL (\pref{fig:e2e-mixing}).
Varying epochs over this fixed corpus allows us to study the effect of data reuse through the training process.

We use natural-language reasoning puzzles in the style of Reasoning Gym~\citep{stojanovski2025reasoninggym}, with verifiable answers and controllable difficulty.
The midtraining corpus contains oracle-generated solutions that demonstrate the underlying reasoning operations.
RL introduces new combinations of these operations, allowing us to study how midtraining supports learning beyond the demonstrated patterns.
In \emph{Base}, puzzle sizes remain within the range represented in midtraining.
In \emph{Extended}, we add larger puzzles to RL training and evaluate on held-out puzzles of these larger sizes.
\emph{Extended generalization} evaluates the same Extended models on puzzles larger than those encountered during either midtraining or RL.

Both approaches use \maxrl with 6000 updates for Base and 12000 for Extended, and the same number of RL generations per update.
We report accuracy (pass@1) and coverage (pass@16) on held-out questions, averaging RL results over two training seeds.
Compute comparisons include pretraining, midtraining, and RL.
\pref{app:nltower-construction} describes the puzzle construction and provides examples; \pref{app:e2e-setup} gives the training and evaluation details.

\subsubsection{Unified Training can Obtain Better Task Performance and Generalization}

\begin{figure}[t]
        \centering
        \includegraphics[width=\linewidth]{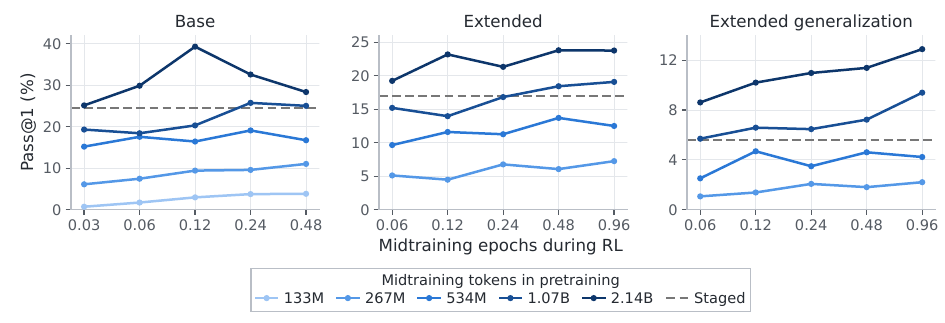}
    \caption{%
    \textbf{Final performance and generalization.}
    Mean pass@1 over two seeds after 6000 RL updates in Base and 12000 in Extended.
    Colors indicate midtraining tokens mixed during pretraining; the horizontal axis measures passes over the same fixed corpus during RL.
    Dashed lines show the best staged configuration on each setting's primary test (\pref{app:e2e-staged-sweep}).
    Extended generalization evaluates the same Extended models on a larger puzzles excluded from RL training (\pref{app:nltower-setup}).}
    \label{fig:e2e-demo-rate}
\end{figure}

In \pref{fig:e2e-demo-rate}, we compare final accuracy across different data configurations for unified training.
For task performance and generalization (\pref{q:performance}), the best unified configurations outperform staged training in both Base and Extended.
This advantage also holds on Extended generalization, which tests generalization by evaluating the models on larger puzzles that are unseen in both midtraining and RL.
Unified training therefore can improve both performance on the RL tasks and generalization beyond the training distribution.

For data mixing and reuse (\pref{q:mixing}), unified training benefits from mixing more midtraining tokens during pretraining, as indicated by the fact that higher intensity curves achieve better performance in \pref{fig:e2e-demo-rate}. The benefit of mixing midtraining data during RL is similar, but less consistent: with the largest pretraining-midtraining mixture (2.14B), the best Base accuracy is achieved at an intermediate RL mixture, whereas in Extended and Extended generalization, the performance overall improves with more midtraining data mixed during RL.
This differs from staged training, where extending midtraining beyond the best-performing budget reduces final task accuracy (\pref{app:e2e-staged-sweep}). 

This suggests that unified training can make better data reuse (as the two best-performing configurations perform 2 and 4 epochs of training over the midtraining corpus during pretraining), while staged training is more sensitive to the amount of midtraining data used.

\subsubsection{Unified Training can be Compute-optimal in More Settings}

\begin{figure}[t]
        \centering
        \includegraphics[width=\linewidth]{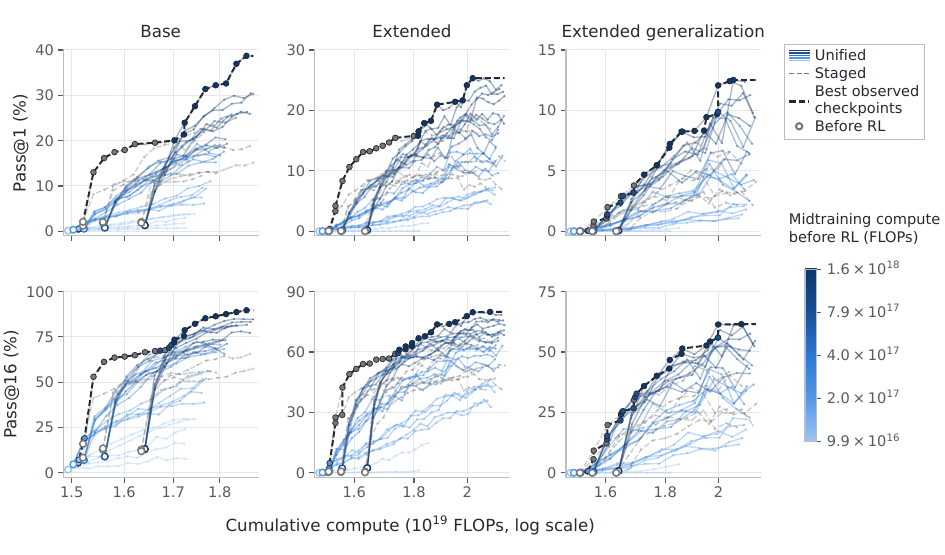}
    \caption{%
    \textbf{Accuracy and coverage across training compute budgets.}
    The top and bottom rows show pass@1 and pass@16 against cumulative compute, including pretraining, midtraining, and RL.
    Colored curves show unified configurations: darker colors indicate more midtraining compute before RL, while curves of the same color vary the midtraining mixture during RL.
    Gray dashed curves show all six staged configurations, spanning three midtraining budgets and two learning rates (\pref{app:e2e-staged-sweep}).
    Open circles mark the pre-RL evaluations in \pref{app:e2e-initialization}.
    The black curve connects the best observed checkpoints for each metric, with colored or gray markers identifying unified or staged training.
    RL points average performance and estimated compute over two seeds.
    Each column ends at the largest final compute budget among its staged runs.
    Extended generalization evaluates the same Extended models on 12-person puzzles.
    \pref{app:compute-accounting} gives the compute estimates.}
    \label{fig:e2e-compute}
\end{figure}

The previous comparison fixes the number of RL updates within each setting and compares final performance, but different data mixtures require different amounts of compute.
In \pref{fig:e2e-compute}, we therefore compare the accuracy and coverage attainable within the same total training budget (\pref{q:compute}).
Each curve follows one configuration through RL, with the horizontal axis including all compute spent on pretraining, midtraining, and RL.
Colored curves show unified training, while gray curves show staged training.
The black curve connects the best observed checkpoints within each compute budget, allowing both the training configuration and the duration of RL to vary; its colored or gray markers indicate whether the unified or staged approach achieves that performance.

Staged training is compute-optimal in terms of accuracy at intermediate compute budget, whereas unified training achieves higher accuracy at larger budgets and the smallest budget in both Base and Extended. Staged training is more compute-optimal in terms of coverage, and the benefit is even more prominent when testing generalization, where unified training achieves the best accuracy and coverage for almost all compute budgets. \pref{app:e2e-staged-compute} shows the same comparison after subtracting the shared pretraining cost, making the compute allocated to midtraining and RL more visible.
Thus, unified training can be more compute-optimal in terms of accuracy, coverage and generalization, where it achieves better performance under a larger compute budget, especially under the most important setting with more compute. 

\subsubsection{Unified Training can Retain General Capabilities Better}
\begin{figure}[!htbp]
    \centering
    \includegraphics[width=0.90\linewidth]{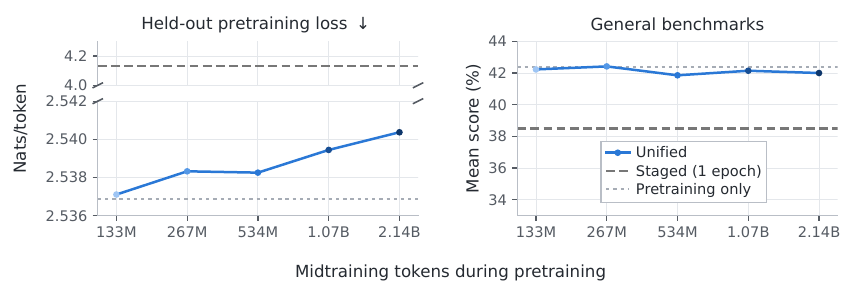}
    \caption{\textbf{Pretraining retention before RL.}
    Unified initializations vary the amount of midtraining data mixed during pretraining.
    Horizontal references show pretraining alone and staged midtraining with one epoch at a peak learning rate of $5\times10^{-5}$, selected for the best retention on both measures.
    Loss is measured on held-out pretraining data; mean score averages five general benchmarks.
    The loss axis is broken, with the lower interval expanded for readability.
    \pref{app:e2e-retention} reports individual benchmarks and all staged configurations.}
    \label{fig:e2e-cpt-retention}
\end{figure}

We next ask whether learning from the midtraining reasoning corpus preserves the general capabilities acquired from pretraining data (\pref{q:retention}).
In \pref{fig:e2e-cpt-retention}, we compare the unified and staged initializations before RL using held-out pretraining loss and mean performance on five general benchmarks.
For this comparison, we select the staged training configuration with the best retention among all six tested variants: it has both the lowest held-out loss and the highest mean benchmark score among all the staged training configurations.

Across all five unified mixtures between pretraining and midtraining, unified training maintains similar retention as the pretrain-only model, even as the amount of midtraining data increases.
On the other hand, the best staged configuration has substantially higher held-out loss and lower mean benchmark performance.
Also note that the staged configuration with the best final RL accuracy performs even worse on both retention measures (\pref{app:e2e-retention}).
These results suggest that mixing midtraining data during pretraining can substantially reduce forgetting while the model learns reasoning skills.

\subsubsection{Additional Analyses}

\paragraph{Reasoning capability before RL}
To understand how the final performance develops across training, we evaluate the models before RL (\pref{app:e2e-initialization}).
Both approaches perform better on the reasoning patterns demonstrated in midtraining than on new combinations of those patterns, and generalization to larger puzzles remains limited.
The staged reference also has higher Base accuracy before RL than any unified initialization, yet the best unified models outperform it after RL.
These results highlight the importance of evaluating midtraining by how it supports subsequent learning, beyond its immediate effect on accuracy.

\paragraph{Combining midtraining and RL losses}
We examine whether the way the two data sources are combined affects learning from the same data mixture (\pref{app:e2e-sftrl}).
Holding the initialization and example counts fixed, we compare unified training with separately normalized midtraining and RL losses combined using a fixed coefficient.
Unified training achieves the highest observed final accuracy and coverage in Base and Extended generalization, while the separately normalized alternative often performs better in Extended.
However, the best coefficient varies across configurations, which suggests that it requires more tuning than the unified loss.
The shared loss therefore provides a competitive way to combine the sources, with their relative contributions following the token mixture.

\paragraph{Retaining pretraining data during RL}
We also examine whether adding pretraining data during RL preserves general capabilities while supporting further reasoning improvements (\pref{app:e2e-pretraining-replay}).
In the evaluated configurations, adding pretraining text lowers held-out pretraining loss, while reasoning accuracy and coverage decline in most comparisons, including on Extended generalization.
General benchmark performance improves in most configurations, although the gains are small.

\FloatBarrier
\subsection{Unified Training with Real-world Midtraining Data}
\label{sec:real}
The controlled setting uses midtraining data that is task-specific and RL tasks whose difficulties can be precisely controlled. To obtain a better understanding for real-world training, we now use a heterogeneous midtraining corpus to examine how unified training during mathematical RL affects task performance, broader capabilities, and continual learning.
We use a 0.76B-parameter model that we train from scratch through pretraining and midtraining.
Unified training starts from this base model, trains on Math12K, and mixes data from a 240M-token corpus with the same composition as the midtraining corpus, including web text, mathematics, code, and reasoning traces.
Staged training uses the additional corpus for midtraining before RL.
\pref{app:math-design} gives the setup.
We report \emph{Math} as the average of MATH-500~\citep{hendrycks2021math,lightman2023verify} and GSM8K~\citep{cobbe2021training}, and \emph{Coding} as the average of HumanEval+ and MBPP+~\citep{chen2021evaluating,austin2021program,liu2023evalplus}, with equal benchmark weights.

\subsubsection{Data Mixtures Tradeoff Accuracy, Coverage, and Retention}

\begin{figure}[t]
        \centering
        \includegraphics[width=0.9\linewidth]{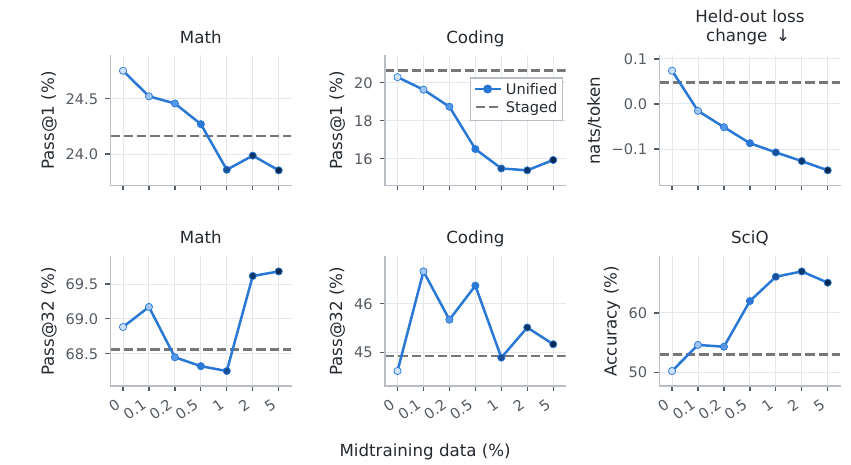}
    \caption{\textbf{Effect of the midtraining data proportion.}
    Final results after 12000 updates.
        Fractions are token proportions estimated before training.
    Loss changes are relative to the base model.
    \pref{app:math-design} gives the training and evaluation details.}
    \label{fig:math-dose}
\end{figure}

We first ask how increasing the midtraining data proportion affects the balance between mathematical learning and retention (\pref{q:performance,q:mixing,q:retention}).
In \pref{fig:math-dose}, we vary the constant midtraining proportion from 0 to 5\% and evaluate both mathematical performance and capabilities beyond the RL task.
The 0\% ablation uses the same asynchronous trainer as the other unified runs, allowing us to assess the effect of adding midtraining data with the implementation and infrastructure held fixed.

Math accuracy remains similar across proportions, with very small degradation with increasing proportions of midtraining data, and the largest proportions achieve the highest coverage. 
The retention benefit is clearer: all nonzero mixtures reduce held-out midtraining loss below the base-model value, with larger proportions yielding lower loss, while staged training increases it;
SciQ~\citep{welbl2017sciq} accuracy also generally improves with mixing.
We also observe lower Coding accuracy with more midtraining data, despite higher Coding coverage than both the base model and staged training.
Thus, unified training from heterogeneous midtraining data improves retention, but its effects on task performance usually trades off accuracy and coverage.
\pref{app:math-dose-details} reports the full benchmark results.

\FloatBarrier
\subsubsection{Data Mixing Schedules}

\begin{figure}[!htbp]
        \centering
        \includegraphics[width=\linewidth]{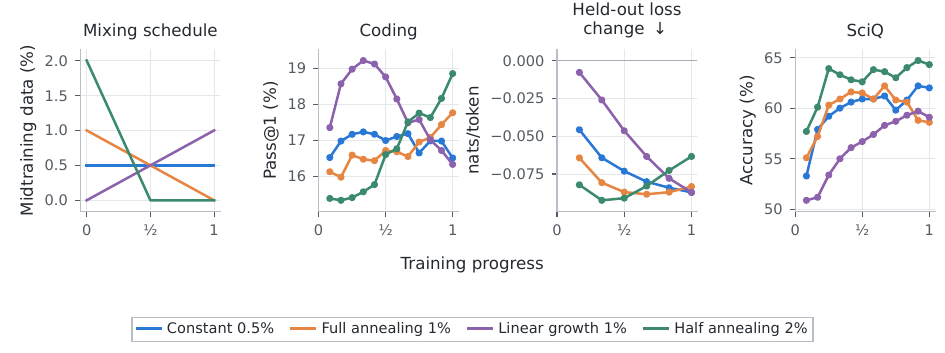}
    \caption{\textbf{Effect of the data mixing schedule.}
        Four schedules with an average configured midtraining proportion of 0.5\%.
    Full and half annealing reach zero at the end and midpoint, respectively; linear growth starts at zero.
    Training details are given in \pref{app:math-schedules}.}
    \label{fig:math-schedules}
\end{figure}

The preceding comparison varies how much midtraining data is used, but keeps its proportion constant throughout RL.
We next ask whether changing when this data is used can reduce the coding accuracy cost while preserving the retention benefits (\pref{q:mixing,q:retention}).
In \pref{fig:math-schedules}, we compare constant mixing, linear decay, linear growth, and decay to zero halfway through training, all with an average configured midtraining proportion of 0.5\% (\pref{app:data-mixing}).

When mixing ends halfway through training, Coding accuracy improves and SciQ gains persist, while held-out corpus loss rises.
Compared with constant mixing, this schedule achieves higher final Coding and SciQ accuracy but retains less of the reduction in corpus loss.
Math accuracy and coverage remain similar across the four schedules (\pref{app:math-schedules}).
Thus, the preferred schedule depends on how performance across tasks is balanced; we leave a detailed analysis of these interactions to future work.

\subsubsection{Unified Training can Enable Better Continual Learning}
\begin{figure}[H]
        \centering
        \includegraphics[width=\linewidth]{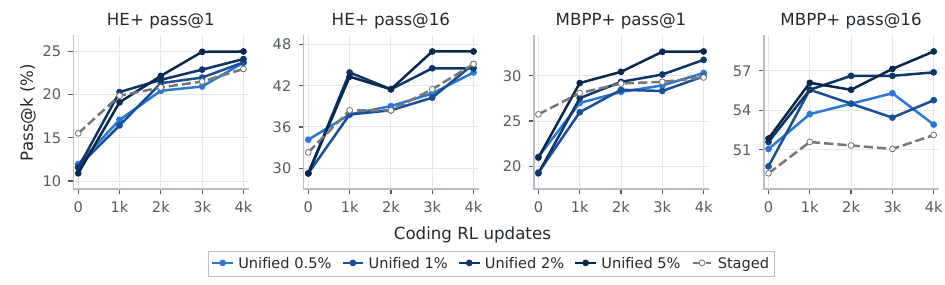}
    \caption{\textbf{Coding RL after mathematical training.}
        Curve labels identify the preceding mathematical training, with percentages denoting constant midtraining proportions.
    All models then use the same coding RL configuration without midtraining data.
    We report pass@1 and pass@16 on HumanEval+ (HE+) and MBPP+ over 4000 coding RL updates.
    \pref{app:real-continual} gives the setup and additional evaluations of mathematics and general knowledge.}
    \label{fig:real-continual}
\end{figure}

We next ask whether unified training improves subsequent learning (\pref{q:continual}).
\pref{fig:real-continual} compares the staged mathematical model with unified models trained at constant 0.5, 1, 2, and 5\% mixing, all using the same coding RL without midtraining data (\pref{app:real-continual}).

Despite lower initial coding accuracy, all unified models achieve higher final accuracy on both benchmarks and higher MBPP+ coverage than staged training.
The 5\% model performs best on all four metrics, while HumanEval+ coverage at smaller mixtures is similar to or below staged training.
These results support unified training as a better starting point for continual learning, with the strongest benefits at the largest midtraining proportion tested.

The finding is further verified in a continual learning setting in a general reasoning domain with WebInstruct~\citep{ma2026general} (\pref{app:real-webinstruct}).
\subsubsection{Additional Analyses}

\paragraph{\maxrl versus \reinforce}
Our formulation accommodates both \maxrl and \reinforce (\pref{sec:unified}), allowing us to examine how the RL objective affects learning from the same data mixture.
At constant 2\% mixing, \reinforce achieves higher Math accuracy, while \maxrl achieves higher Math coverage.
\reinforce also yields lower held-out loss and higher SciQ and ARC-Challenge~\citep{clark2018arc} accuracy.
The coverage and retention differences persist across the additional proportions and schedules in \pref{app:math-weighting-details}.
Thus, the RL objective affects the balance between mathematical coverage and retention even when the data mixture is fixed.

\paragraph{Comparison with KL regularization}
We ask whether regularizing toward the base model can reproduce the benefits of mixing midtraining data (\pref{app:kl-regularization}).
Increasing the KL penalty reduces the deterioration in held-out loss, but the strongest penalty also substantially reduces Math accuracy and coverage.
In this setting, none of the tested penalties reproduces the combination of mathematical performance, lower corpus loss, and higher SciQ and MMLU accuracy achieved by unified training.
\paragraph{Midtraining data composition}
To understand whether changing the midtraining corpus can reduce the coding accuracy cost of mixing, we remove or replace its designated code sources (\pref{app:math-content-details}).
Removing these sources improves Coding accuracy under constant mixing but has little effect under annealing.
Replacing them with verified coding examples improves coverage under annealing without improving accuracy.
Changing the corpus can therefore reduce the coding accuracy cost, but its effect depends on the mixing schedule.

\section{Related Work}
\label{sec:related-work}

\subsection{Foundation-Model Training}
Large-scale maximum likelihood pretraining established the basis for broadly capable language models~\citep{brown2020gpt3,zhang2022opt,chowdhery2023palm,bigscience2024bloom}.
Subsequent systems combine this foundation with supervised learning, preference optimization, and RL through increasingly elaborate training pipelines~\citep{openai2023gpt4,touvron2023llama2,grattafiori2024llama3,deepseek2025v3,guo2025deepseekr1}.
Open research efforts have made training data, recipes, and intermediate models more accessible, supporting closer examination of data curation, mixture schedules, and post-training procedures~\citep{groeneveld2024olmo,allal2025smollm2,lambert2025tulu3,olmo2025olmo3}.
Our work studies the organization of this training process through a common formulation and implementation for static and generated data, allowing their composition and timing to vary without requiring a separate training algorithm for each stage.

\subsection{Understanding Foundation-Model Training}
Empirical scaling studies characterize how performance varies with model size, data, and compute~\citep{hestness2017scaling,rosenfeld2020constructive,kaplan2020scaling,hoffmann2022computeoptimal}.
Subsequent work examines the value of repeated data when fresh data are limited~\citep{muennighoff2023dataconstrained} and the scaling of RL post-training, where initialization, model size, and optimization choices affect learning efficiency~\citep{nimmaturi2025predictive,khatri2026art,tan2026rlscaling}.

Understanding the full training process also requires examining how earlier training shapes subsequent learning.
Studies of transfer scaling and domain adaptation characterize how downstream performance depends on preceding training data~\citep{gururangan2020dontstop,hernandez2021transfer,isik2025downstream}, while recent reasoning studies examine how pretraining and midtraining affect subsequent RL and capability retention~\citep{zhang2026interplay,runwal2026prism,shen2026understanding}.
Complementary work optimizes data mixtures and studies the timing of domain exposure and the benefits of later replay~\citep{xie2023doremi,ye2025datamixing,liu2026midtrainingbridges,baek2026finetuner,kotha2026replaying}.
Our work complements these empirical accounts with a common likelihood formulation that accommodates both static and RL-generated data, enabling the composition, timing, and reuse of these sources to be studied across the training process.

\subsection{Hybrid Offline and Online Reinforcement Learning}
Combining demonstrations or other prior experience with online interaction has long been used to improve exploration and sample efficiency in RL~\citep{ross2012agnostic,hester2018demonstrations,nair2018exploration,rajeswaran2018dexterous}.
This principle underlies a substantial literature on offline-to-online and hybrid RL, including practical algorithms for incorporating prior data into online learning~\citep{nair2020awac,lee2022offlinetoonline} and theoretical analyses of when access to both sources improves learning efficiency~\citep{xie2021policyfinetuning,song2023hybridrl,wagenmaker2023leveragingoffline,zhou2024offlineenhanced,song2024hybridobservation}.

Language-model post-training likewise combines static corpora or demonstrations with online RL.
InstructGPT adds a pretraining likelihood term to the \ppo objective to reduce regressions on public benchmarks~\citep{ouyang2022instructgpt}, and recent work provides growing evidence that combining demonstrations with RL can improve learning and preserve capabilities~\citep{yan2025luffy,liu2025uft,fu2026srft,zhang2026chord}.
These methods typically add a separately weighted supervised term to the RL loss, with a coefficient that is tuned or scheduled~\citep{ouyang2022instructgpt,zhang2026chord}, or adapt the RL objective to off-policy data~\citep{yan2025luffy}.
In unified training, both sources share one likelihood objective, and their relative weight is set by the data mixture; \pref{app:e2e-sftrl} compares this with a separately weighted supervised loss.
Our work further studies the joint use of these sources across the full foundation-model training process, focusing on how their composition, timing, and reuse affect learning across pretraining, midtraining, and RL.

\subsection{Connections between Reinforcement Learning and Supervised Learning}
Connections between RL, supervised learning, and probabilistic inference have long motivated methods based on expectation-maximization, weighted regression, and constrained policy fitting~\citep{dayan1997expectation,peters2007regression,peters2010reps,abdolmaleki2018mpo,levine2018inference,peng2019awr}.
Reward-guided self-training methods such as STaR, ReST-EM, and RAFT iteratively train models by maximum likelihood on selected generated responses~\citep{zelikman2022star,gulcehre2023rest,singh2024beyond,dong2023raft}.
Related work also develops unified analyses of language-model post-training~\citep{lv2025unified}.
Our formulation builds directly on the likelihood objectives of \maxrl and \tailrl~\citep{tajwar2026maxrl,ramasubramanian2026tailrl}.
We use this connection to place static data and suitable RL targets within a common likelihood formulation, justifying mixtures of their training distributions and supporting a shared data format, loss, and trainer.

\subsection{Asynchronous Reinforcement Learning}
Our practical implementation is related to asynchronous RL, which separates acting from learning to make use of parallel hardware~\citep{mnih2016asynchronous}. Distributed actor-learner systems build on this separation by passing experience from many actors to a central learner through a replay buffer or a trajectory queue~\citep{horgan2018distributedprioritizedexperiencereplay,espeholt2018impalascalabledistributeddeeprl,espeholt2019seed}. RL systems for language models follow the same design, running generation and training on separate resources connected by a buffer or a stream~\citep{slime_github,zhong2025streamrlscalableheterogeneouselastic}. These systems differ mainly in how they limit policy lag: AReaL and StaleFlow bound the number of policy versions between generation and training~\citep{fu2026areallargescaleasynchronousreinforcement,li2026staleflowstalenessawaredatamanagement}, whereas PipelineRL updates the generator weights during generation~\citep{piche2025pipelinerlfasteronpolicyreinforcement}.

\section{Discussion}
\label{sec:discussion}

This work presents a common mathematical formulation and training infrastructure for foundation models, bringing static data and suitable RL updates together through maximum likelihood learning. The resulting flexibility allows data sources to overlap and their proportions to change throughout training. Our experiments demonstrate practical benefits from these choices and show that the value of a data source depends on both when it is introduced and how it interacts with subsequent learning.

A central open question is how to select these mixtures systematically. Our experiments reveal tradeoffs among accuracy, coverage, and retention, but do not identify a universally preferred mixture or schedule. The compute study likewise characterizes the evaluated configurations without establishing scaling laws with predictive power. Extending these analyses across model sizes and data budgets could support more predictive choices about when to reuse static data, generate new responses, and change their relative contributions. In addition, our current results are not established with in depth study on how optimization should change with different data schedule and mixture design~\citep{wen2026data}, and we leave this as an important direction for future work. 

The scope of the experiments also remains small-scale. Most runs optimize a single RL task at a time, whereas joint training across tasks may require accounting for competing learning dynamics and unequal rates of progress. Furthermore, our work does not fully eliminate the notion of stages as downstream adaptation of a foundation model to a new task will inevitably require a separate training stage. Our continual learning results provide initial evidence that the preceding training process can improve this adaptation; studying longer task sequences would clarify how that advantage interacts with retention. Together, these questions motivate studying data allocation throughout training as a coupled learning problem.

\section*{Acknowledgments}
This work was supported by an award from the ASCR Leadership Computing Challenge (ALCC) under
project ERCAP0034861.
This research also used resources of the Oak Ridge Leadership Computing Facility (OLCF)
and Argonne Leadership Computing Facility (ALCF) which are a DOE Office of Science User Facility. The authors also acknowledge the use of the DeltaAI advanced computing and data resource~\citep{deltaai}, which is supported by the National Science Foundation (award OAC 2320345) and the State of Illinois. DeltaAI is a joint effort of the University of Illinois Urbana-Champaign and its National Center for Supercomputing Applications. Its computing resources were used as part of ACCESS-approved compute grants~\citep{access_compute}.
The authors thank the CMU FLAME center and the CMU Babel Compute Cluster for compute support for
this project. 
YS and AS acknowledge and thank the support of NSF AI Institute for Societal Decision
Making AI-SDM grant IIS2229881. FT gratefully acknowledges the support of Qualcomm Innovation Fellowship North America 2026 and Bosch Research and Technology Center. DA is supported by the National Science Foundation under Grants CCF-2106778. 
DK is supported by ONR award  (N00014-24-1-2089). AZ acknowledges support from Impossible, Jane Street and Amazon AGI.

\clearpage
\label{sec:main-text-end} 
\safebibliography{refs}

\clearpage
\appendix
\addtocontents{toc}{\protect\setcounter{tocdepth}{2}}
\begingroup
    \renewcommand{\contentsname}{Appendix Contents}
    \renewcommand{\cfttoctitlefont}{\Large\bfseries}
    \setlength{\cftbeforetoctitleskip}{0pt}
    \setlength{\cftaftertoctitleskip}{12pt}
    \pdfbookmark[1]{Appendix Contents}{appendix-contents}
    \tableofcontents
\endgroup
\clearpage
\section{Data Mixing and Compute Calculation}
\label{app:accounting}

This appendix explains how we construct training batches from multiple data sources and account for the resulting training compute.
\pref{app:data-mixing} describes the amount of data drawn from each source, how that amount changes during training, and how the losses are combined.
\pref{app:compute-accounting} then converts the training budgets into corpus exposure and model-FLOP estimates.
The appendices for each experiment provide the remaining setup details, evaluation protocols, and results.

\subsection{Data Mixing}
\label{app:data-mixing}

A data mixture specifies how much data from each source enters a training batch.
In the reasoning-puzzle experiments, we set the number of packed sequences added during pretraining and the number of prompt--response examples added during RL.
In the real-world data experiments, we instead specify a target fraction of tokens from midtraining data.
Because examples differ in length, these two measures differ slightly but we do not expect this to impact the overall results, as the number of tokens in each example is relatively consistent in the controlled settings, and we present most of our results in terms of converted token numbers or FLOPs.
We use mean length estimates to convert the desired token fraction into the number of examples added to each batch, as detailed below and in \pref{sec:impl-buffer}.

\begingroup
\paragraph{Midtraining data during pretraining}
During the final 8147 pretraining update steps, each reasoning-puzzle run retains the original batch of 256 pretraining sequences of pretraining data and adds $k\in\{8,16,32,64,128\}$ midtraining sequences.
All sequences are packed to length 2048.
There are $256+k$ sequences in the combined batch, giving a midtraining token fraction of $k/(256+k)$. We do not vary the existing learning-rate schedule or other hyperparameters even though the batch size changes, and we defer a study of the optimal learning rate and batch size choice to future work.
Across the 8147 updates, the model therefore processes $8147\times k\times2048$ additional midtraining tokens.
\pref{tab:e2e-pretraining-mixtures} gives these quantities for each choice of $k$.

Here a packed sequence is a fixed-length token block that can contain more than one prompt--response example.
The midtraining sequences are drawn from a fixed corpus of 796175 reasoning traces, containing 509942713 tokens in total.
Prompt, response, and end-of-sequence tokens all contribute to the loss during this phase.
Increasing $k$ increases exposure to the same corpus while preserving the original pretraining-data stream.

\begin{table}[htbp]
    \centering
    \small
    \begin{tabular}{rrr}
        \toprule
        \shortstack{Midtraining sequences\\per update, $k$} & \shortstack{Midtraining token fraction\\$k/(256+k)$} \\
        \midrule
        8   & 3.03\%  \\
        16  & 5.88\%  \\
        32  & 11.11\% \\
        64  & 20.00\% \\
        128 & 33.33\% \\
        \bottomrule
    \end{tabular}
    \caption{Midtraining exposure during the final 8147 pretraining updates for the Base and Extended reasoning-puzzle settings.}
    \label{tab:e2e-pretraining-mixtures}
\end{table}

\paragraph{Midtraining data during RL}
During RL, we specify the mixture by the number of complete midtraining examples added to each update.
Unified training adds $m\in\{4,8,16,32,64\}$ prompt--response examples from the same 796175-example corpus alongside the RL generations.
We vary this count for the five pretraining mixtures in \pref{tab:e2e-pretraining-mixtures}.
The $m=0$ controls start from the same initializations but receive no midtraining examples during RL, allowing us to distinguish the effect of the initialization from that of further data mixing.
In this phase, prompt tokens are masked: only response and end-of-sequence tokens contribute to the midtraining loss. Note that this is a different masking choice than during pretraining, where all tokens contribute to the loss. This design choice is made due to the nature of the RL phase, but we do not expect it to impact the overall results.
\pref{app:compute-accounting} explains how the number of examples per update determines the total number of corpus passes.
\endgroup

\paragraph{Pretraining data during RL}
\pref{app:e2e-pretraining-replay} adds pretraining text alongside the RL generations and midtraining examples.
We use the $k\in\{32,128\}$ initializations, corresponding to 534M and 2.14B midtraining tokens during pretraining, with $m\in\{8,32\}$ midtraining examples per RL update.
Each update adds four pretraining-text sequences, drawn from the original 75\% ClimbMix and 25\% Dolmino mixture.
Their lengths are chosen using the mean midtraining response length so that the added text contributes approximately one loss-bearing pretraining token for every ten midtraining response tokens, including end-of-sequence tokens.
This proportion is relative to the midtraining data, not to all tokens in the batch.
The pretraining tokens receive unit weight in the shared loss, alongside the unchanged numbers of RL generations and midtraining examples.

\paragraph{Token proportions in the real-world data study}
\begingroup
In the real-world data experiments, we control the mixture through a target fraction of midtraining tokens rather than a fixed number of midtraining examples.
Let $T_{\mathrm{mid}}$ count the midtraining tokens and $T_{\mathrm{rl}}$ count the RL response tokens in a batch.
The latter count includes unsuccessful responses as well as successful ones.
The midtraining fraction is
\begin{align*}
    f = \frac{T_{\mathrm{mid}}}{T_{\mathrm{mid}}+T_{\mathrm{rl}}}.
\end{align*}
The percentages reported in the figures are the configured targets for this fraction.
Each update retains 512 RL responses and adds midtraining examples.
To determine how many examples to add, we use the target fraction and the mean example lengths measured before training.
Because the resulting count need not be an integer, rounding is applied to cumulative counts across updates.
The length estimates remain fixed throughout training.
Consequently, a constant configured fraction does not guarantee a constant fraction of tokens in the batches actually processed: that fraction can change as RL responses become longer or shorter. We believe this tradeoff is necessary to avoid complicated design choices in the RL training loop.

The schedule experiments vary the configured fraction as training proceeds.
Let $u\in[0,1]$ be the fraction of optimizer updates completed, and let $a=0.005$, corresponding to an average configured proportion of 0.5\%.
The four schedules in \pref{fig:math-schedules} are
\begin{align*}
    f_{\mathrm{constant}}(u)&=a,
    &f_{\mathrm{full}}(u)&=2a(1-u),\\
    f_{\mathrm{growth}}(u)&=2au,
    &f_{\mathrm{half}}(u)&=4a\max\{1-2u,0\}.
\end{align*}
The constant schedule remains at 0.5\% throughout training.
Full annealing starts at 1\% and decreases linearly to zero, while growth increases linearly from zero to 1\%.
Half annealing starts at 2\%, reaches zero halfway through training, and remains zero thereafter.
Thus, each schedule has the same average configured proportion $a$, but allocates midtraining data differently over time.
Their realized token proportions can still differ because the response lengths can differ across runs.
\pref{app:math-schedules} reports these comparisons and additional schedules, including three-quarter annealing, which reaches zero after three quarters of training.
\endgroup

\paragraph{Loss weighting and normalization}
\begingroup
The comparison in \pref{app:e2e-sftrl} keeps the initialization and example counts fixed while changing how the RL and midtraining losses are combined.
For an RL batch, let $N_R$ be the sum of response-token negative log-likelihoods weighted by their RL weights.
For the midtraining examples, let $N_D$ be the corresponding sum of response-token negative log-likelihoods before applying any midtraining weight.
Let $T_R$ and $T_D$ be the respective response-token counts, including end-of-sequence tokens; $T_R$ also includes tokens from unsuccessful RL responses.
The two objectives, each optimized with a single optimizer, are
\begin{align*}
    \mathcal{L}_{\mathrm{unified}} &= \frac{N_R+wN_D}{T_R+T_D},
    &
    \mathcal{L}_{\mathrm{separate}} &= \frac{N_R}{T_R}+\alpha\frac{N_D}{T_D}.
\end{align*}
Unified training divides the combined weighted loss by the total token count $T_R+T_D$; $w$ is the weight on midtraining tokens.
The alternative first computes a token mean for each source and then multiplies the midtraining mean by a fixed coefficient $\alpha$.

Writing the unified objective in terms of these two source means gives a relative coefficient $wT_D/T_R$ on the midtraining mean and an overall scale $T_R/(T_R+T_D)$.
This makes its dependence on batch composition explicit: the relative contribution of midtraining data changes with the ratio of midtraining to RL response tokens.
In the separate-loss objective, $\alpha$ remains fixed throughout a run.
\endgroup

\subsection{Compute Calculations in the Controlled Experiments}
\label{app:compute-accounting}
\label{app:e2e-compute}

We distinguish the amount of data processed from the compute required to process it.
Training exposure counts examples or tokens each time they are presented to the model, including repeated presentations of the same data.
The model-FLOP estimates additionally account for the cost of generating RL responses and training on the resulting batches.
These estimates provide the horizontal axes in \pref{fig:teaser,fig:e2e-compute,fig:e2e-compute-expanded}.

\begingroup
\paragraph{Pretraining, midtraining, and RL budgets}
The full RL budgets are 6000 updates in Base and 12000 in Extended, with 512 requested generations per update.
A run reaching these horizons therefore requests 3072000 or 6144000 generations, respectively.
Midtraining exposure during RL depends on the number of examples added to each update.
For $U$ updates and $m$ examples per update, the model processes $Um$ midtraining examples.
Dividing by the 796175 examples in the corpus gives $Um/796175$ corpus passes.
At the full training horizons, $m\in\{4,8,16,32,64\}$ gives approximately 0.03--0.48 passes in Base and 0.06--0.96 passes in Extended.

Staged training instead processes 1, 2, or 4 passes before RL and uses no midtraining data during RL.
For unified training, both the amount mixed into pretraining and the amount added during RL contribute to the total midtraining exposure.
Increasing either quantity reuses the same corpus but requires additional training compute.
Thus, matching the number of RL updates does not match total compute.
\endgroup

\paragraph{Model-FLOP estimates}
We use the dense-model approximation of \citet{shen2026understanding}: processing $T$ model-input tokens costs approximately $6NT$ FLOPs for training and $2NT$ FLOPs for generation, where $N=123587328$ is the parameter count.
We track four token totals to account for the full training process.
$T_{\mathrm{pre}}$ is the number of shared pretraining tokens, and $T_{\mathrm{mid}}$ is the number of additional midtraining tokens processed before RL.
During RL, $T_{\mathrm{rollout}}$ counts the cumulative prompt and response tokens in the RL data, and $T_{\mathrm{mix}}$ counts the midtraining tokens added to the RL batches.
The compute estimates are
\begin{align*}
    C_{\mathrm{total}}
    &=6N(T_{\mathrm{pre}}+T_{\mathrm{mid}})
      +8NT_{\mathrm{rollout}}+6NT_{\mathrm{mix}},\\
    C_{\mathrm{additional}}
    &=C_{\mathrm{total}}-6NT_{\mathrm{pre}}.
\end{align*}
The coefficient $8N$ on the RL token total accounts for generation ($2N$) followed by one policy-training pass ($6N$).
Both prompt and response tokens count toward this estimate because the model processes both, even when prompt positions are masked from the loss.
We likewise count all nonpadding input tokens for static data.
These compute totals therefore differ from the token counts used to normalize response losses in \pref{app:data-mixing}.
Subtracting $6NT_{\mathrm{pre}}$ removes the pretraining cost shared by every run and leaves the compute allocated to midtraining and RL.
The estimates omit attention-length corrections, padding overhead, activation recomputation, evaluation, unused generations, and work discarded after restarts.

For \pref{fig:teaser,fig:e2e-compute,fig:e2e-compute-expanded}, all runs retain the same 20000014336-token pretraining stream.
Unified initializations add approximately 133M--2.14B midtraining tokens before RL.
Staged initializations process 1, 2, or 4 passes over the 509942713-token corpus, at each of the two learning rates in \pref{app:e2e-staged-sweep}.
During RL, the training logs record batch response-token counts every ten updates.
We use these measurements to estimate cumulative response tokens and subtract the estimated midtraining response tokens to obtain the RL response total.
The estimates for RL prompt tokens and complete midtraining examples use the mean lengths of their respective corpora.
At each checkpoint, we average compute and performance over the same two RL training seeds, so the horizontal and vertical coordinates describe the same pair of models.
For the before-RL points, we use the shared initialization evaluations in \pref{app:e2e-initialization} and count only the pretraining and midtraining compute already incurred.

\pref{fig:e2e-compute,fig:e2e-compute-expanded} compare unified training with all six staged configurations.
For each setting, we display checkpoints up to the largest final compute budget among its staged runs.
Both figures show pass@1 in the top row and pass@16 in the bottom row.
For each metric, we order the evaluated checkpoints by compute and retain those that improve the best score observed so far.
These checkpoints form the observed frontier.
The black curve connects them; the connecting segments are visual guides, not additional evaluations.
After the final improvement, the curve remains at the best observed value up to the displayed budget limit.
The two figures use the same checkpoints and frontiers.
\pref{fig:e2e-compute} includes all training compute, whereas \pref{fig:e2e-compute-expanded} subtracts $6NT_{\mathrm{pre}}$ from every horizontal coordinate to show compute beyond shared pretraining.
The right panel of \pref{fig:teaser} summarizes the Extended results using the same evaluations and compute limit as \pref{fig:e2e-compute}.
It constructs separate curves for unified and staged training: at each budget, each curve gives the highest observed accuracy or coverage among that approach's tested configurations.
A curve increases only when a better checkpoint becomes available within the budget.
The curves begin at the first evaluated initialization for each approach, while the horizontal axis begins at the common pretraining cost.

\paragraph{Real-world data experiments}
The real-world data experiments in \pref{sec:real} compare runs with the same number of optimizer updates, rather than matching total compute.
Adding midtraining data changes the amount of work in each update, and asynchronous training can reuse available RL responses.
\pref{app:math-zero-mixing} separately reports the measured time per update and describes the hardware and policy-staleness differences between implementations.
Those timings measure optimization throughput; they do not measure model FLOPs or the rate of generating new RL responses.

\clearpage
\begingroup
\tcbset{nl example/.style={colback=white,colframe=PaperBorder,
    coltext=black,coltitle=black,colbacktitle=PaperPretraining!45,
    fonttitle=\small\bfseries,fontupper=\small,
    boxrule=0.6pt,arc=1mm,left=2.5mm,right=2.5mm,top=2mm,bottom=2mm,
    before skip=8pt,after skip=8pt}}

\section{Reasoning Puzzles: Construction and Examples}
\label{app:nltower-construction}

We construct reasoning puzzles with verifiable solutions to study how a data design supports learning throughout the whole training pipeline.
The construction lets us control both the reasoning patterns required to solve a puzzle and the number of people whose locations must be determined.
We use a symbolic oracle to produce worked solutions for the midtraining corpus, then introduce additional combinations of the same reasoning operations during RL.
Below, we describe the puzzles, explain which patterns are included in midtraining, and give complete examples of the resulting model inputs and training responses.

Our construction is most closely related to Einstein-style (zebra) logic puzzles, in which relational clues determine an assignment of attributes to entities~\citep{pmlr-v267-lin25i}.
We similarly ask for an assignment of people to floors, using numerical constraints and references through friendship and relative position.
Knights and Knaves puzzles also require deductions from interdependent statements, but the unknowns are which speakers always tell the truth and which always lie~\citep{xie-etal-2025-memorization}.
In our puzzles, all clues are true; the challenge is to resolve their references and find an assignment that satisfies them jointly.

\subsection{Puzzle Construction}
\label{app:nltower-puzzles}
The task is to assign each of a set of named people to a distinct floor.
The occupied floors are sampled from the integers 1--40 and listed in the prompt, together with each person's best friend and a set of natural-language clues.
The model must recover the complete assignment from this information.
The friendship relation is a permutation without self-links: each person names one other person as their best friend, and each person is named exactly once.
Friendship need not be reciprocal.

The clues constrain the floor numbers through differences, sums, comparisons, parity, and integer multiples.
A clue may identify a person directly by name or indirectly through friendship, floor rank, or another person's location.
For example, ``the person directly above Ava'' refers to the occupant of the next \emph{occupied} floor above Ava, even if unoccupied floors lie between them.
``Directly below'' is defined in the same way.

To generate a puzzle, we first sample a hidden assignment and construct a chain of clues whose interpretation depends on deductions from earlier clues.
We then add two redundant clues and use a symbolic solver to verify that the assignment is unique.
The solver also checks that the assignment can be recovered by constraint propagation, so that the oracle can provide a deductive solution without searching over alternative assignments.

\subsection{Reasoning Patterns and Midtraining Data}
\label{app:nltower-patterns}
A \emph{reference} is a phrase that identifies a person or their floor within a clue.
Interpreting a reference can require one or more operations, such as looking up a person's best friend or finding the occupant of the highest floor.
We control which combinations of these operations appear in the midtraining corpus, \emph{so the model is only taught how to perform reasoning and deductions, instead of overfit to all kinds of references}.

For example, midtraining includes references to a named person's best friend, the person on the highest floor, and the best friend of the person on an explicitly specified floor.
It omits ``the best friend of the person on the highest floor,'' which first identifies the highest-floor occupant and then looks up that person's best friend.
\emph{The constituent operations therefore appear in midtraining, while this particular composition does not.}

\pref{tab:nltower-references} lists the demonstrated patterns and the three families of compositions omitted from midtraining.
RL prompts introduce these omitted compositions, and held-out questions evaluate performance on the corresponding distributions.
\emph{This design allows us to study whether reasoning skills acquired from midtraining support subsequent learning through RL.}
The omitted compositions are new at the start of RL, but they are included in RL training rather than reserved exclusively for evaluation.

\begin{table}[H]
    \centering
    \small
    \renewcommand{\arraystretch}{1.12}
    \begin{tabular}{p{0.26\linewidth}p{0.66\linewidth}}
        \toprule
        Availability & Example reference \\
        \midrule
        Included in midtraining & Ava's best friend \\
        & The best friend of Ava's best friend \\
        & The person directly above Ava \\
        & The person on the second-highest floor \\
        & The best friend of the person on floor 19 \\
        \midrule
        Introduced during RL & The best friend of the person on the highest floor \\
        & The best friend of the person directly above Ava \\
        & The best friend of the person directly above Ava's best friend \\
        \bottomrule
    \end{tabular}
    \caption{Reference patterns in the reasoning puzzles.
    Each row gives a schematic example of how a clue can identify a person or floor.
    The final three rows combine operations that occur in midtraining but whose composition is omitted from that corpus; the last row represents a family of longer nested references.
    \pref{fig:nltower-reference} illustrates two such compositions in an actual question.}
    \label{tab:nltower-references}
\end{table}

\paragraph{Oracle reasoning traces}
We construct the midtraining corpus from questions that use only the demonstrated patterns.
For each question, a symbolic oracle maintains the set of feasible floors for every person and repeatedly applies the clues to eliminate inconsistent assignments.
As these sets narrow, the oracle can resolve indirect references and use the resulting information in further deductions.
It records the deductions in natural language and returns the complete assignment without branching into alternative solutions.

We generate up to four traces for each question by varying the order in which constraints are applied and how the deductions are verbalized.
After removing duplicate traces, each remaining prompt--response pair becomes one midtraining example.
The response contains the oracle's deductions followed by the complete assignment in a box.
\pref{app:e2e-data} specifies the corpus size and its use across training phases.

\clearpage
\subsection{Examples}
\label{app:nltower-examples}

The first example shows both a complete midtraining prompt and the oracle response used as its training target.
The subsequent examples show evaluation prompts from the Base and Extended settings defined in \pref{app:e2e-data}.
All prompts and the oracle response are reproduced in full from stored records; only typography and line wrapping are changed.
The final \texttt{Answer:} cue belongs to the model input and indicates where generation begins.
For the evaluation examples, we show the correct assignments separately to explain the task; these answers are not supplied to the model.

\paragraph{A midtraining question and its oracle response}
This six-person example uses reference patterns demonstrated in the corpus.
\begin{tcolorbox}[nl example,title={Midtraining example: question}]
Hugo, Maya, Ethan, Zoe, Ella and Mia live on different floors of the same tower. The occupied floors are \{6, 13, 23, 28, 31, 38\}. Best friends: Hugo's best friend is Ethan, Maya's best friend is Ella, Ethan's best friend is Mia, Zoe's best friend is Hugo, Ella's best friend is Maya, Mia's best friend is Zoe. Here is what is known:
\begin{itemize}[leftmargin=1.3em,itemsep=0pt,parsep=0pt,topsep=3pt]
    \item Zoe lives 3 floors above Ella.
    \item Ethan lives 10 floors above Ethan's best friend.
    \item Ethan lives 17 floors above the person on the lowest floor.
    \item Zoe's best friend lives 25 floors above Mia.
    \item Mia lives on an odd floor.
    \item Ethan lives on an odd floor.
\end{itemize}
Work out which floor each person lives on. Answer with a comma-separated list like "Hugo=NUMBER, ..." covering everyone.\par
Answer:\par
\end{tcolorbox}
\begin{tcolorbox}[nl example,title={Midtraining example: oracle response}]
The clue Zoe - Ella = 3 leaves these pairs for (Zoe,Ella): \{(31,28)\}.\par
That pins Zoe to 31, so 31 is taken.\par
That pins Ella to 28, so 28 is taken.\par
The lowest floor is 6.\par
Using Ethan - floor 6 = 17: Ethan is now in \{23\}.\par
That pins Ethan to 23, so 23 is taken.\par
Zoe's best friend is Hugo.\par
The clue Zoe's best friend - Mia = 25 leaves these pairs for (Hugo,Mia): \{(38,13)\}.\par
That pins Hugo to 38, so 38 is taken.\par
That pins Mia to 13, so 13 is taken.\par
That pins Maya to 6, so 6 is taken.\par
\smallskip
\noindent\fbox{\strut Hugo=38, Maya=6, Ethan=23, Zoe=31, Ella=28, Mia=13}
\end{tcolorbox}
\clearpage
\paragraph{Base setting}
The following eight-person question includes two compositions omitted from midtraining, highlighted in bold. \pref{fig:nltower-reference} resolves these references using the gold assignment.
\begin{tcolorbox}[nl example,title={Base setting: evaluation question}]
Zoe, Maya, Ava, Nora, Hugo, Felix, Mia and Ruby live on different floors of the same tower. The occupied floors are \{2, 12, 19, 20, 22, 29, 37, 40\}. Best friends: Zoe's best friend is Hugo, Maya's best friend is Ruby, Ava's best friend is Nora, Nora's best friend is Mia, Hugo's best friend is Zoe, Felix's best friend is Ava, Mia's best friend is Felix, Ruby's best friend is Maya. Here is what is known:
\begin{itemize}[leftmargin=1.3em,itemsep=0pt,parsep=0pt,topsep=3pt]
    \item The floor numbers of Maya's best friend and Mia add up to 42.
    \item Hugo lives on an odd floor.
    \item Maya lives 8 floors above \textbf{the best friend of the best friend of the person on the highest floor}.
    \item Hugo lives 17 floors above Hugo's best friend.
    \item The floor numbers of \textbf{the best friend of the person on the highest floor} and Zoe add up to 24.
    \item The floor numbers of Ruby's best friend and Ruby add up to 57.
    \item Ruby lives on an even floor.
    \item Nora lives 21 floors above Hugo.
\end{itemize}
Work out which floor each person lives on. Answer with a comma-separated list like "Zoe=NUMBER, ..." covering everyone.\par
Answer:\par
\end{tcolorbox}
\noindent\textbf{Gold answer:} Zoe=2, Maya=37, Ava=12, Nora=40, Hugo=19, Felix=29, Mia=22, Ruby=20.
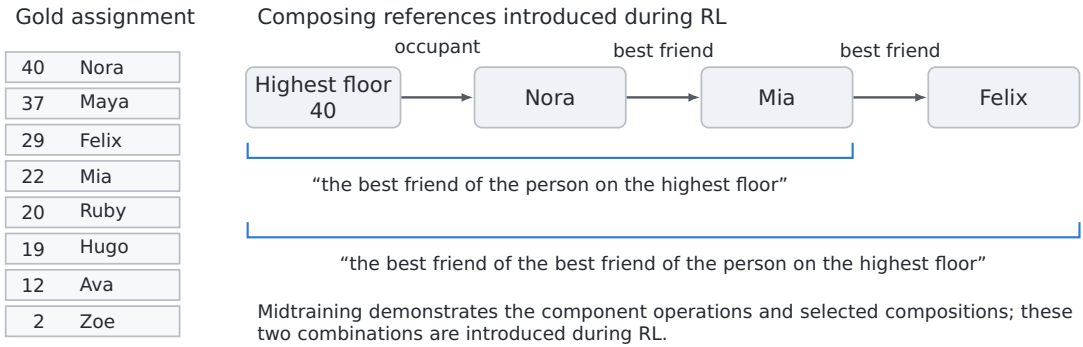
\begin{figure}[H]
    \centering
    \begin{tikzpicture}[paper figure,text=PaperInk,
    refnode/.style={draw=PaperBorder,fill=PaperPretraining!45,
        rounded corners=1mm,minimum height=8mm,minimum width=20mm,align=center},
    reflabel/.style={font=\paperfigsmall,text=PaperInk,align=center}]
    \node[anchor=west] at (0,2.45) {Gold assignment};
    \foreach \floor/\person/\y in {40/Nora/1.8,37/Maya/1.32,29/Felix/.84,22/Mia/.36,20/Ruby/-.12,19/Hugo/-.60,12/Ava/-1.08,2/Zoe/-1.56}{
        \draw[draw=PaperBorder,fill=PaperPretraining!25] (0,\y-.20) rectangle (2.3,\y+.20);
        \node[anchor=east,font=\paperfigsmall] at (.65,\y) {\floor};
        \node[anchor=west,font=\paperfigsmall] at (.85,\y) {\person};
    }
    \node[anchor=west] at (3.2,2.45) {Composing references introduced during RL};
    \node[refnode] (floor) at (4.2,1.4) {Highest floor\\40};
    \node[refnode] (nora) at (7.2,1.4) {Nora};
    \node[refnode] (mia) at (10.2,1.4) {Mia};
    \node[refnode] (felix) at (13.2,1.4) {Felix};
    \draw[paper arrow] (floor)--node[above=4mm,reflabel]{occupant}(nora);
    \draw[paper arrow] (nora)--node[above=4mm,reflabel]{best friend}(mia);
    \draw[paper arrow] (mia)--node[above=4mm,reflabel]{best friend}(felix);
    \draw[draw=PaperUnified,line width=.8pt] (3.2,.8)--(3.2,.6)--(11.2,.6)--(11.2,.8);
    \node[anchor=north,reflabel] at (7.2,.48)
        {``the best friend of the person on the highest floor''};
    \draw[draw=PaperUnified,line width=.8pt] (3.2,-.25)--(3.2,-.45)--(14.2,-.45)--(14.2,-.25);
    \node[anchor=north,reflabel,text width=11cm] at (8.7,-.57)
        {``the best friend of the best friend of the person on the highest floor''};
    \node[anchor=north west,reflabel,text width=10.8cm,align=left] at (3.2,-1.2)
        {Midtraining demonstrates the component operations and selected compositions; these two combinations are introduced during RL.};
\end{tikzpicture}
    \caption{Resolving composed references in the eight-person evaluation example.
    The diagram shows the gold assignment, with occupied floors spaced evenly for readability.
    The highest floor is occupied by Nora, whose best friend is Mia; Mia's best friend is Felix.
    The two highlighted clues therefore reduce to $37-29=8$ and $22+2=24$.
    This illustration uses the known answer to explain the task; the assignment is not part of the model input.}
    \label{fig:nltower-reference}
\end{figure}
\clearpage
\paragraph{Extended setting}
This setting retains the same composition rules while increasing the number of people. In this ten-person example, the highlighted references require identifying the best friend of the occupant of an extreme floor.
\begin{tcolorbox}[nl example,title={Extended setting: evaluation question}]
Owen, Ruby, Liam, Ava, Ethan, Hugo, Mia, Iris, Ella and Felix live on different floors of the same tower. The occupied floors are \{1, 2, 4, 5, 6, 8, 10, 13, 14, 20\}. Best friends: Owen's best friend is Felix, Ruby's best friend is Iris, Liam's best friend is Owen, Ava's best friend is Ruby, Ethan's best friend is Mia, Hugo's best friend is Ethan, Mia's best friend is Ella, Iris's best friend is Ava, Ella's best friend is Liam, Felix's best friend is Hugo. Here is what is known:
\begin{itemize}[leftmargin=1.3em,itemsep=0pt,parsep=0pt,topsep=3pt]
    \item Ava lives on an even floor.
    \item The best friend of the person on floor 5 lives 1 floor above Ella.
    \item \textbf{The best friend of the person on the lowest floor} lives 4 floors above Ruby.
    \item The person on the second-highest floor lives 1 floor above Felix.
    \item The floor numbers of Ava's best friend and Ethan add up to 6.
    \item Hugo lives 4 floors above \textbf{the best friend of the person on the highest floor}.
    \item Ruby's best friend lives 10 floors above Ruby.
    \item Mia lives on an even floor.
    \item The floor numbers of Iris's best friend and Liam add up to 18.
    \item Mia lives 15 floors above Hugo.
\end{itemize}
Work out which floor each person lives on. Answer with a comma-separated list like "Owen=NUMBER, ..." covering everyone.\par
Answer:\par
\end{tcolorbox}
\noindent\textbf{Gold answer:} Owen=6, Ruby=4, Liam=8, Ava=10, Ethan=2, Hugo=5, Mia=20, Iris=14, Ella=1, Felix=13.

\clearpage
\endgroup
\clearpage
\section{Pretraining, Midtraining, and RL: Setup and Additional Results}
\label{app:e2e-setup}
\label{app:nltower-setup}

This appendix explains the experimental design for \pref{sec:e2e} and provides the additional results used to interpret the main comparisons.
We first describe the data, model initializations, RL procedure, and evaluation protocol, then examine performance before and after RL, compute allocation, loss aggregation, and retention.

\begingroup

\subsection{Tasks and Midtraining Data}
\label{app:e2e-data}
\label{app:nltower-data}

We use the reasoning puzzles constructed in \pref{app:nltower-construction} to vary when the model sees midtraining data and how often it reuses that data.
All comparisons use the same fixed corpus of oracle solutions.
We first describe this corpus and the question distributions used for RL and evaluation, then give the training and evaluation protocols.

\paragraph{Oracle traces in the midtraining corpus}
The midtraining corpus contains 200000 questions: 100000 with 6--7 people and 100000 with 8--9 people.
All questions use only the demonstrated reference patterns described in \pref{app:nltower-patterns}.
The oracle produces up to four reasoning traces per question; removing duplicate traces leaves 796175 prompt--response examples.
Together, the prompts, responses, and end-of-sequence tokens contain 509942713 tokens, or approximately 510M tokens.
Staged midtraining and unified training draw from this same corpus, including when unified training mixes midtraining data into pretraining or RL.
Increasing midtraining exposure therefore means reusing existing examples rather than introducing new questions.
\pref{app:nltower-examples} gives a complete corpus example and evaluation prompts.

\paragraph{Two RL settings}
We vary the reasoning required during RL in two settings.
In \emph{Base}, all puzzles contain 8--9 people, so their size remains within the range covered by midtraining.
The RL prompt pool includes both questions using only demonstrated patterns and questions that require compositions omitted from midtraining.
The primary test set contains the latter type, allowing us to measure learning of these compositions without also increasing puzzle size.

In \emph{Extended}, the RL pool additionally includes puzzles with 10--11 people, and the primary test set uses this larger size.
The reference families remain the same as in Base, but the model must now recover more people's locations.
Puzzles containing omitted compositions require two to four steps that use them in the generated clue chain.
We call these \emph{omitted steps}: the term counts uses of compositions absent from midtraining, not all deductions required to solve the puzzle.
\pref{tab:nltower-splits} gives the composition of both RL pools and their test sets.

Finally, \emph{Extended generalization} evaluates the same Extended models on a separate test set with 12 people, a size absent from both midtraining and RL.
It requires no additional training.
All test questions are disjoint from the midtraining corpus and the RL prompt pools.

\begin{table}[htbp]
    \centering
    \small
    \renewcommand{\arraystretch}{1.12}
    \begin{tabular}{lccrr}
        \toprule
        & People & \shortstack{Omitted\\steps} & Base & Extended \\
        \midrule
        RL prompts: demonstrated patterns & 8--9 & 0 & 10000 & 10000 \\
        RL prompts: omitted compositions & 8--9 & 2--4 & 18570 & 11666 \\
        RL prompts: larger puzzles & 10--11 & 2--4 & -- & 10000 \\
        \midrule
        Primary test questions & 8--9 & 2--4 & 1024 & -- \\
        Primary test questions & 10--11 & 2--4 & -- & 1024 \\
        Size-generalization test questions & 12 & 2--4 & -- & 1024 \\
        RL updates & -- & -- & 6000 & 12000 \\
        \bottomrule
    \end{tabular}
    \caption{Question pools and training horizons for the two reasoning-puzzle settings.
    ``Omitted steps'' counts uses of compositions absent from midtraining in each puzzle's clue chain.
    These compositions are introduced during RL, while the 12-person puzzle size is used only for evaluation.
    Test questions are distinct from training questions.}
    \label{tab:nltower-splits}
    \label{tab:e2e-difficulties}
\end{table}

\subsection{Model and Pretraining}
\label{app:e2e-pretraining}

\paragraph{Model and original pretraining}
We train a 124M-parameter model from scratch on a mixture of 75\% ClimbMix~\citep{diao2025climb}\footnote{\url{https://huggingface.co/datasets/karpathy/climbmix-400b-shuffle}} and 25\% Dolmino~\citep{olmo2025olmo3}.\footnote{\url{https://huggingface.co/datasets/allenai/dolma3_dolmino_mix-100B-1025}}
The model has 12 transformer layers, hidden dimension 768, feedforward dimension 2048, and 12 attention heads.
\pref{tab:e2e-pretraining-hyperparameters} gives the architecture and pretraining settings.
The original pretraining run uses 38147 updates, each with 256 packed sequences of length 2048, and processes 20000014336 tokens in total.
All comparisons start from the same checkpoint after the first 30000 updates.
For the remaining 8147 updates, they preserve the same web-data stream; unified training also introduces midtraining examples, as described below.

\begin{table}[htbp]
    \centering
    \begingroup
\small
\renewcommand{\arraystretch}{1.08}
\begin{tabular}{p{0.29\linewidth}p{0.65\linewidth}}
    \toprule
    Setting & Value \\
    \midrule
    Parameters / layers & 123587328 / 12 \\
    Hidden / feedforward size & 768 / 2048 \\
    Attention heads & 12 \\
    Embeddings and normalization & Tied embeddings, rotary position embeddings, RMSNorm \\
    Tokenizer / context length & GPT-2 / 2048 tokens \\
    \midrule
    Global batch size & 256 packed sequences of 2048 tokens \\
    Training updates & 38147 \\
    Optimizer / betas & AdamW / $(0.9,0.95)$ \\
    Peak / minimum learning rate & $6\times10^{-4}$ / $6\times10^{-5}$ \\
    Weight decay / gradient clip & 0.1 / 1.0 \\
    Learning-rate schedule & 200-update warmup; constant through update 30518; cosine decay over the final 7629 updates \\
    \bottomrule
\end{tabular}
\endgroup

    \caption{\textbf{Model and original pretraining hyperparameters.}
    Mixed-pretraining branches resume from update 30000 with the saved optimizer state and complete the remaining 8147 updates of the same learning-rate schedule.}
    \label{tab:e2e-pretraining-hyperparameters}
\end{table}

\paragraph{Unified Pretraining and Midtraining}
During the final 8147 pretraining updates, unified training retains all 256 web-text sequences in each batch and adds 8, 16, 32, 64, or 128 packed midtraining sequences, each of length 2048.
These choices contribute approximately 133M, 267M, 534M, 1.07B, or 2.14B midtraining tokens, respectively; \pref{tab:e2e-pretraining-mixtures} gives the exact counts.
These additions increase the batch size and total compute while preserving the web-data exposure and number of optimizer updates.
They also allow multiple passes over the fixed corpus at the larger mixtures.
During mixed pretraining, the language-model loss applies to every token in the packed midtraining sequences, including the prompts, reasoning traces, and end-of-sequence tokens.
The five resulting models serve directly as initializations for RL.

\clearpage
\subsection{Staged Initialization and RL Training}
\label{app:e2e-rl}
\label{app:nltower-training}
  
\paragraph{Staged initialization}
Staged training first completes the original 20B-token pretraining run without mixing in the reasoning corpus.
It then trains on that corpus for 1, 2, or 4 epochs at a peak learning rate of $2\times10^{-4}$ or $5\times10^{-5}$, giving six candidate initializations.
Here, each epoch is one pass over the midtraining corpus.
The prompt conditions the response, but its tokens are masked from the loss; only response and end-of-sequence tokens contribute to the midtraining objective.

We run RL from each candidate.
Unless otherwise specified, a single staged reference is selected by the highest mean final RL pass@1, separately for Base and Extended, as reported in \pref{app:e2e-staged-sweep}.
Both settings select two epochs at $2\times10^{-4}$.
The Extended generalization evaluation uses the same Extended checkpoints and does not enter configuration selection.

\paragraph{Shared RL configuration}
Both approaches use \maxrl with the optimizer and generation settings in \pref{tab:e2e-training-hyperparameters}.
Each update samples 32 prompts and generates 16 responses per prompt, yielding 512 RL responses.
We make one optimization pass over this batch.
The training budget is 6000 updates for Base and 12000 for Extended.

For each prompt, we estimate its success probability $\hat p$ as the fraction of its 16 generations that receive reward one.
Correct generations receive weight $1/\hat p$, while unsuccessful generations receive zero weight.
All prompt groups remain in the batch, including those with no correct generation.
The token-mean normalization also counts the unsuccessful response tokens, even though their loss terms have zero weight.
During training, a response that reaches the generation limit without an end-of-sequence token receives zero reward.

\begin{table}[htbp]
    \centering
    \begingroup
\small
\renewcommand{\arraystretch}{1.12}
\begin{tabular}{p{0.27\linewidth}p{0.28\linewidth}p{0.35\linewidth}}
    \toprule
    Setting & Staged midtraining & RL (both approaches) \\
    \midrule
    Optimizer & AdamW & AdamW \\
    Adam betas & $(0.9,0.95)$ & $(0.9,0.999)$ \\
    Learning rate & $2\times10^{-4}$ peak & $10^{-5}$ \\
    Schedule & 5\% warmup, then cosine decay & Constant, no warmup \\
    Weight decay & 0.01 & 0.01 \\
    Gradient clip & 1.0 & 1.0 \\
    Context length & 2048 tokens & 2048 tokens \\
    \midrule
    Training budget & 2 epochs & 6000 updates (Base); 12000 (Extended) \\
    Global batch & 32 examples & 32 prompts, 16 generations each \\
    Added midtraining data & --- & 0 for staged; 4--64 examples for unified \\
    Maximum response length & --- & 1280 tokens \\
    Temperature / top-$p$ & --- & 1.0 / 1.0 \\
    Top-$k$ truncation & --- & Disabled \\
    KL penalty / entropy bonus & --- & None / none \\
    \bottomrule
\end{tabular}
\endgroup

    \caption{\textbf{Staged midtraining and shared RL hyperparameters.}
    The midtraining column gives the selected configuration for both Base and Extended; \pref{app:e2e-staged-sweep} reports the full sweep.
    Unified training uses the same RL settings and adds midtraining examples to each batch.
    To assess the contribution of this mixing during RL, we also run controls that start from the same mixed-pretraining initializations but add no midtraining examples.}
    \label{tab:e2e-training-hyperparameters}
    \label{tab:nltower-hyperparameters}
\end{table}

\paragraph{Unified training}
Unified runs start from the five mixed-pretraining initializations described above.
During RL, each update adds 4, 8, 16, 32, or 64 examples from the full midtraining corpus to a batch of 512 RL responses, preserving the number of RL generations used in staged training.
For these midtraining examples, the prompt tokens are masked from the loss, and response and end-of-sequence tokens receive weight one.
We combine these losses with the weighted RL losses using the shared token-mean normalization in \pref{sec:implementation}.

We also train controls from the same mixed-pretraining initializations without adding midtraining data during RL.
Comparing against these controls isolates the contribution of mixing during RL from that of the initialization.
The main sweep does not include pretraining data during RL; \pref{app:e2e-pretraining-replay} examines this addition separately.
\pref{app:data-mixing,app:compute-accounting} give the mixture counts and corresponding training exposure.

\paragraph{Combining midtraining and RL losses}
To examine the choice of normalization, we compare unified training with an alternative that averages the RL and midtraining losses separately, then adds the midtraining loss with a fixed coefficient $\alpha\in\{0.1,0.5\}$.
For Extended, we also test $\alpha=1$ with the initialization trained on 2.14B midtraining tokens during pretraining, across all five midtraining example counts during RL.
The comparison holds the initialization, training data, example counts, optimizer, and RL update budget fixed.
Unified training instead uses the shared token mean with unit midtraining weight.
\pref{app:data-mixing} defines both objectives, and \pref{app:e2e-sftrl} reports the comparisons at each available common training horizon.

\clearpage
\subsection{Evaluation Protocol}
\label{app:e2e-evaluation}
\label{app:nltower-evaluation}

\paragraph{Accuracy and coverage}
\label{app:passk-estimator}
Each test set contains 1024 held-out questions.
For each question, we independently sample 64 responses at temperature 1.0 and top-$p$ 1.0, without top-$k$ truncation.
Generation uses a 2048-token context and a maximum of 1280 response tokens.
The grader checks the final boxed answer against the complete correct person-to-floor assignment and gives no partial credit.

We report sampled pass@1 as \emph{accuracy} and pass@16 as \emph{coverage}.
Accuracy estimates the probability that one sampled response is correct; coverage estimates the probability that at least one of 16 sampled responses is correct.
We compute both from the 64 responses per question using the unbiased estimator of \citet{chen2021evaluating}.
For $c$ correct responses among $n$ samples, the per-question estimator is
\begin{align*}
    \widehat{\mathrm{pass@}k}
    =1-\frac{\binom{n-c}{k}}{\binom{n}{k}},
    \qquad 1\leq k\leq n,
\end{align*}
where $\binom{n-c}{k}=0$ for $n-c<k$.
We average the per-question estimates over the test set.
Coverage does not assume that the model can identify which of its responses is correct.

\paragraph{Averaging over training seeds}
Each RL configuration is trained with two seeds from the same pre-RL initialization.
We first compute accuracy and coverage separately for each trained model, then average the two scores.
In particular, we do not pool responses from different models when estimating pass@16.
Training curves compare the two models at the same optimizer update and include a point only when both evaluations are available.
We apply this procedure to both the primary test set and, for Extended models, the 12-person generalization test.

The evaluations before RL follow a different convention because the two RL seeds share one initialization.
We evaluate that initialization once with the same sampling protocol (\pref{app:e2e-initialization}); its reported score is therefore a single model evaluation rather than an average over subsequent training seeds.

\paragraph{Pretraining retention}
We assess retention before RL to measure how incorporating the reasoning corpus changes capabilities acquired during pretraining.
We use two measures: held-out language-model loss and performance on general benchmarks.
For held-out loss, we evaluate approximately 400000 fixed tokens from each of ClimbMix and Dolmino in 1024-token windows.
We then combine the two losses with weights 0.75 and 0.25, matching the pretraining mixture.
The general-capability score averages MMLU~\citep{hendrycks2021mmlu} accuracy, ARC-Easy~\citep{clark2018arc} normalized accuracy, HellaSwag~\citep{zellers2019hellaswag} normalized accuracy, TruthfulQA MC2~\citep{lin2022truthfulqa}, and WinoGrande~\citep{sakaguchi2020winogrande} accuracy.
All benchmarks use zero-shot likelihood evaluation without a chat template: the model is scored by the likelihood it assigns to candidate answers rather than by sampled generations.
Each shared initialization is evaluated once, so these scores are not averages over the subsequent RL training seeds.
\pref{app:e2e-retention} reports the full results for the five unified and six staged initializations, together with the pretraining-only reference.
For the comparison in \pref{app:e2e-pretraining-replay}, we apply the same retention evaluations after RL and average the scores over the two training seeds.

\paragraph{Training exposure and compute}
Within each setting, the methods have the same prescribed number of RL updates and requested generations per update.
This does not make their total compute equal: midtraining exposure before and during RL varies across configurations, and the lengths of generated responses can also differ.
We therefore estimate the compute used throughout training under the model-FLOP convention in \pref{app:compute-accounting}.
This estimate includes training on web and midtraining data, generating RL responses, and training on those responses.

For the reasoning-puzzle runs, we estimate RL token counts from sampled training logs and the mean prompt length in the corresponding pool.
Midtraining counts use the corpus token totals and the number of examples processed.
Both prompt and response tokens count toward compute, including prompt positions that are masked from the loss.
The cumulative-compute plots include all of these costs.
The plots of compute beyond shared pretraining subtract the common 20B-token web-pretraining cost from every run, making differences in the remaining training budget easier to compare.

\FloatBarrier
\endgroup

\clearpage
\subsection{Reasoning Performance before RL}
\label{app:e2e-initialization}

\begingroup

We first evaluate what the models have learned from the midtraining corpus before they receive any RL updates.
For unified training, this evaluation follows pretraining with midtraining data mixed into its final part; for staged training, it follows the separate midtraining phase.
These evaluations distinguish the reasoning ability acquired before RL from the improvements obtained during RL.

\pref{fig:e2e-cpt-passk} compares the five unified initializations with the model trained on pretraining data alone and the selected staged initialization.
The staged reference uses two midtraining epochs at a peak learning rate of $2\times10^{-4}$, selected by final RL accuracy rather than by its score before RL (\pref{app:e2e-staged-sweep}).
Both approaches use the same fixed midtraining corpus, although they differ in how much they reuse it and in their total compute.
The figure therefore compares the reasoning ability of the initializations; the compute comparison appears in \pref{app:e2e-staged-compute}.

\begin{figure}[H]
    \centering
    \includegraphics[width=\linewidth]{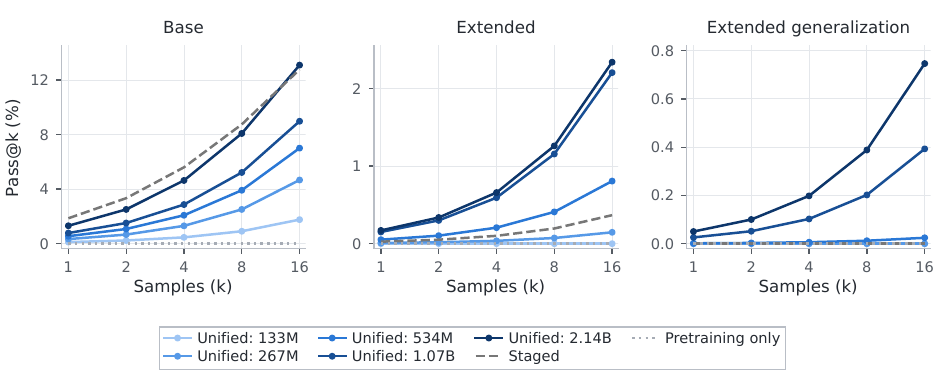}
    \caption{\textbf{Reasoning accuracy and coverage before RL.}
    Each curve shows pass@$k$, the estimated probability of obtaining a correct answer in $k$ attempts, for an initialization evaluated before RL.
    Colors indicate the number of midtraining tokens mixed during pretraining; the dashed gray curve shows the selected staged initialization.
    Midtraining puzzles contain 6--9 people, while the three panels use the Base, Extended, and Extended generalization test sets defined in \pref{app:e2e-data}.
    Estimates use 64 responses per question on 1024 questions in each test set.
    Vertical scales differ across panels; zero-valued curves overlap.}
    \label{fig:e2e-cpt-passk}
\end{figure}

Increasing the amount of midtraining data used during pretraining improves unified accuracy and coverage on Base.
The selected staged initialization has higher pass@1 than all five unified initializations, while the largest unified mixture reaches similar pass@16.
Increasing the unified mixture also improves performance on Extended and Extended generalization, but transfer to larger puzzles remains limited before RL: even the largest mixture reaches only 2.34\% and 0.75\% pass@16, respectively.

To distinguish learning the demonstrated reasoning patterns from applying them in new compositions, \pref{tab:reasoning-prerl} also evaluates held-out 8--9-person puzzles containing only demonstrated patterns.
The table includes all five unified and all six staged initializations.
Both approaches perform substantially better on the demonstrated patterns than on Base, whose puzzles have the same sizes but include compositions omitted from midtraining.
The models therefore learn the demonstrated patterns, but applying them in new compositions remains difficult before RL.
Comparing this table with \pref{tab:e2e-staged-sweep} also shows that the strongest staged initialization on Base before RL does not yield the best final RL performance.
This motivates selecting the staged reference by its performance after RL.

\begin{table}[H]
    \centering
    \begingroup
\small
\setlength{\tabcolsep}{5pt}
\renewcommand{\arraystretch}{1.10}
\begin{tabular}{l*{8}{r}}
\toprule
 & \multicolumn{2}{c}{\shortstack{Demonstrated\\patterns}} & \multicolumn{2}{c}{Base} & \multicolumn{2}{c}{Extended} & \multicolumn{2}{c}{\shortstack{Extended\\generalization}} \\
\cmidrule(lr){2-3}\cmidrule(lr){4-5}\cmidrule(lr){6-7}\cmidrule(lr){8-9}
Midtraining & pass@1 & pass@16 & pass@1 & pass@16 & pass@1 & pass@16 & pass@1 & pass@16 \\
\midrule
None & 0.00 & 0.00 & 0.00 & 0.00 & 0.00 & 0.00 & 0.00 & 0.00 \\
\midrule
\multicolumn{9}{l}{\textit{Unified: mixed during pretraining}} \\
133M tokens & 0.92 & 12.11 & 0.12 & 1.76 & 0.00 & 0.00 & 0.00 & 0.00 \\
267M tokens & 7.23 & 55.98 & 0.34 & 4.67 & $<0.01$ & 0.15 & 0.00 & 0.00 \\
534M tokens & 16.74 & 79.67 & 0.55 & 7.01 & 0.05 & 0.81 & $<0.01$ & 0.02 \\
1.07B tokens & 35.60 & 94.92 & 0.78 & 8.98 & 0.15 & 2.21 & 0.03 & 0.39 \\
2.14B tokens & 51.98 & 98.83 & 1.31 & 13.09 & 0.17 & 2.34 & 0.05 & 0.75 \\
\midrule
\multicolumn{9}{l}{\textit{Staged: midtraining at peak LR $2\times10^{-4}$}} \\
1 epoch & 78.76 & 99.95 & 1.19 & 8.34 & 0.02 & 0.37 & 0.00 & 0.00 \\
\rowcolor{hl}
2 epochs & 86.97 & 99.99 & 1.86 & 12.77 & 0.03 & 0.37 & 0.00 & 0.00 \\
4 epochs & 90.70 & 100.00 & 2.02 & 12.77 & 0.05 & 0.57 & 0.00 & 0.00 \\
\midrule
\multicolumn{9}{l}{\textit{Staged: midtraining at peak LR $5\times10^{-5}$}} \\
1 epoch & 65.15 & 99.79 & 2.09 & 15.97 & 0.04 & 0.61 & 0.00 & 0.00 \\
2 epochs & 78.83 & 99.93 & 2.05 & 13.43 & 0.05 & 0.64 & 0.00 & 0.00 \\
4 epochs & 87.68 & 100.00 & 1.85 & 11.88 & 0.02 & 0.21 & 0.00 & 0.00 \\
\bottomrule
\end{tabular}
\endgroup

    \caption{\textbf{Reasoning performance across all initializations before RL.}
    Values are percentages; positive values below 0.01\% are shown as $<0.01$, and zeros indicate no observed correct responses.
    The demonstrated-pattern test contains held-out 8--9-person puzzles using only patterns present in midtraining; Base uses the same puzzle sizes but introduces omitted compositions.
    The shaded row is the staged configuration selected by final RL accuracy for both Base and Extended.
    Each row evaluates one initialization before RL, which is subsequently shared by the two RL training seeds.}
    \label{tab:reasoning-prerl}
\end{table}

\endgroup

\clearpage
\subsection{Staged Performance after RL}
\label{app:e2e-staged-sweep}

\begingroup

We evaluate six staged midtraining configurations to select a strong baseline for the full training pipeline.
Within each setting, we select the configuration with the highest final RL accuracy.
The sweep combines 1, 2, or 4 passes over the fixed corpus with a peak learning rate of either $2\times10^{-4}$ or $5\times10^{-5}$.
All configurations start from the same pretrained model and are followed by RL without midtraining data.
\pref{tab:e2e-staged-sweep} reports their final accuracy and coverage after 6000 RL updates in Base and 12000 in Extended.
For each configuration, we compute the metrics separately for two RL training seeds and report their mean.

We select the configuration with the highest mean final pass@1 on the primary test set, independently for Base and Extended.
The same configuration is selected in both settings: two midtraining epochs at a peak learning rate of $2\times10^{-4}$.
It supplies the staged references in \pref{fig:e2e-demo-rate}.
For Extended generalization, we evaluate the selected Extended models on the 12-person test set without making a separate configuration choice using those results.
The compute comparisons in \pref{fig:e2e-compute,fig:e2e-compute-expanded} retain all six staged configurations, since the best choice at a smaller compute budget need not be the configuration with the highest final accuracy.

\begin{table}[H]
    \centering
    \begingroup
\small
\setlength{\tabcolsep}{5pt}
\renewcommand{\arraystretch}{1.12}
\begin{tabular}{rr*{6}{r}}
\toprule
 & & \multicolumn{2}{c}{Base} & \multicolumn{2}{c}{Extended} & \multicolumn{2}{c}{\shortstack{Extended\\generalization}} \\
\cmidrule(lr){3-4}\cmidrule(lr){5-6}\cmidrule(lr){7-8}
Epochs & Peak LR & pass@1 & pass@16 & pass@1 & pass@16 & pass@1 & pass@16 \\
\midrule
1 & $2\times10^{-4}$ & 21.79 & 72.40 & 16.14 & 63.51 & 6.58 & 41.78 \\
1 & $5\times10^{-5}$ & 12.24 & 54.54 & 9.88 & 51.30 & 3.18 & 26.31 \\
\rowcolor{hl}
2 & $2\times10^{-4}$ & 24.54 & 73.51 & 16.91 & 64.80 & 5.58 & 35.90 \\
2 & $5\times10^{-5}$ & 12.74 & 55.25 & 7.25 & 41.13 & 2.17 & 18.86 \\
4 & $2\times10^{-4}$ & 15.15 & 57.49 & 11.64 & 53.20 & 4.10 & 28.53 \\
4 & $5\times10^{-5}$ & 18.82 & 65.58 & 6.96 & 41.43 & 2.15 & 19.68 \\
\bottomrule
\end{tabular}
\endgroup

    \caption{\textbf{Staged performance after RL.}
    Each row specifies the number of midtraining passes and peak learning rate used before RL.
    Pass@1 and pass@16 are percentages averaged over two RL training seeds, after 6000 updates in Base and 12000 in Extended.
    The shaded row maximizes final pass@1 independently on the Base and Extended primary test sets.
    Extended generalization evaluates the same Extended checkpoints on 12-person puzzles and is not used to select the reference.}
    \label{tab:e2e-staged-sweep}
    \label{tab:e2e-staged-rl-selection}
\end{table}

Using more of the fixed corpus before RL does not necessarily improve the final result: the best four-epoch configuration remains below the best two-epoch configuration in both settings.
The effect of learning rate also depends on the number of epochs.
Together with the evaluations before RL in \pref{app:e2e-initialization}, these results show why the staged reference is selected after RL: strong performance immediately after midtraining does not by itself identify the most effective initialization for further learning.

\endgroup

\clearpage
\subsection{Additional Compute Comparisons}
\label{app:e2e-staged-compute}

\begingroup

\pref{fig:e2e-compute} performs the compute comparison with pretraining compute in to account.
To make differences in the RL stage easier to see, \pref{fig:e2e-compute-expanded} subtracts this shared cost from the horizontal coordinate of every checkpoint.
The remaining compute includes midtraining before RL, generation and training during RL, and any midtraining data mixed into RL updates; \pref{app:compute-accounting} gives the calculation.

The configurations and evaluations are identical to those in \pref{fig:e2e-compute}.
Each plot ends at the same total training budget as its main-text counterpart, with the common pretraining cost subtracted from the displayed value.
This subtraction changes the spacing of points on the logarithmic axis but does not change which checkpoints attain the highest observed performance within a budget.

\begin{figure}[H]
    \centering
    \includegraphics[width=\linewidth]{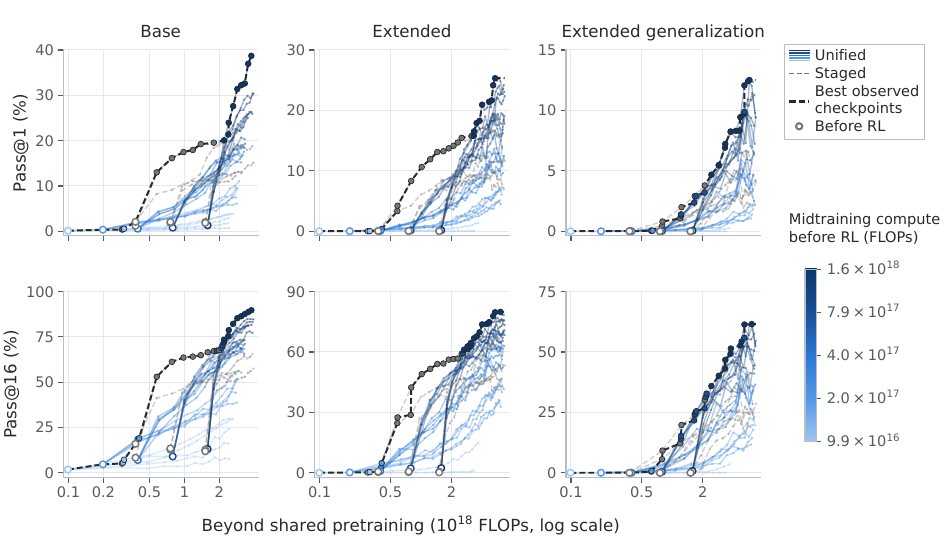}
    \caption{\textbf{Accuracy and coverage versus compute beyond shared pretraining.}
    The top and bottom rows show pass@1 and pass@16, respectively.
    Gray dashed curves show all six staged configurations; colored curves show unified configurations, with darker colors indicating more midtraining compute before RL.
    All checkpoints also appear in \pref{fig:e2e-compute}; only their horizontal coordinates change by subtracting the common pretraining cost.
    The black curve connects checkpoints that attain the highest observed value within a compute budget across both approaches, with marker colors identifying their source.
    Open circles mark evaluations of the shared initializations before RL.
    Subsequent points use evaluations available for both RL seeds at the same update, averaging performance and estimated compute over those two seeds.
    Curves extend through available evaluations up to 6000 updates in Base and 12000 in Extended, subject to the displayed compute budget.
    Extended generalization evaluates the Extended checkpoints on 12-person puzzles and uses the same training costs.}
    \label{fig:e2e-compute-expanded}
\end{figure}

Subtracting the shared cost makes the tradeoffs at smaller additional budgets easier to see.
Staged training contributes to the observed frontier at mid-level budgets during in distribution eval, while unified training achieves higher accuracy and coverage at smaller and larger budgets in all three evaluations. Unified training achieves the best-compute accuracy and coverage for generalization evaluation for almost all compute budgets. 
This comparison allows each approach to use its best observed checkpoint within the budget, even if that checkpoint precedes the end of its run.
It therefore accounts for choices of both midtraining configuration and RL duration, rather than comparing only the final checkpoints.

\endgroup

\clearpage
\subsection{Retaining Pretraining Data during RL}
\label{app:e2e-pretraining-replay}

\begingroup

The main reasoning experiments mix midtraining data with RL generations.
Here we ask whether also retaining pretraining data helps preserve general capabilities, and how this affects reasoning performance.
Each comparison uses the same initialization, RL and midtraining example counts, optimizer, and training horizon.
One variant additionally receives pretraining text from the original corpus mixture, with approximately 10\% as many loss-bearing tokens as the midtraining responses (\pref{app:data-mixing}).
The added text contributes to the shared loss; the comparison holds the RL update budget fixed rather than matching total compute.

We use initializations trained with 534M or 2.14B midtraining tokens during pretraining, and mix 8 or 32 midtraining examples into each RL update.
We report all four combinations for Base after 6000 updates and Extended after 12000 updates.
The same Extended models are evaluated on Extended generalization.
All results average two training seeds using the protocol in \pref{app:e2e-evaluation}.

\pref{fig:e2e-pretraining-replay} shows that adding pretraining text reduces accuracy and coverage in most comparisons.
The only exceptions are small gains in Extended coverage and generalization accuracy with the 2.14B-token initialization and 32 midtraining examples per RL update.
Held-out pretraining loss decreases in all eight training configurations (\pref{tab:reasoning-replay-retention}).
The general benchmark average improves in seven of the eight configurations, although all changes are smaller than 0.7 percentage points.
At the tested proportion, retaining pretraining data therefore improves retention, usually at the expense of reasoning performance.

\begin{figure}[H]
    \centering
    \includegraphics[width=\linewidth]{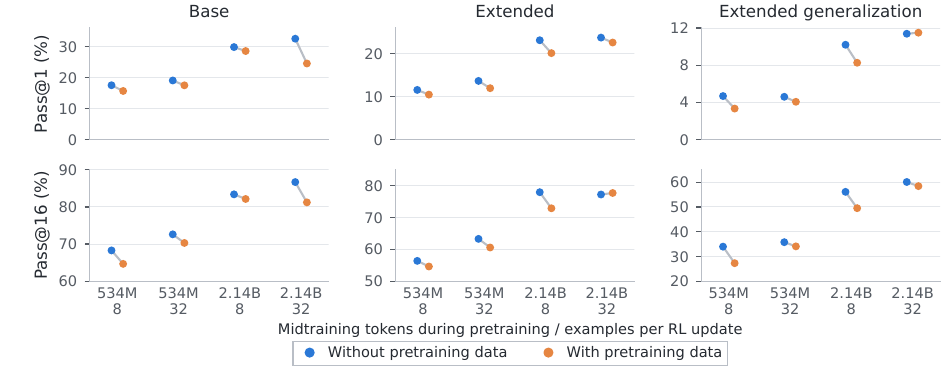}
    \caption{\textbf{Effect of retaining pretraining data during RL.}
    Final accuracy (top) and coverage (bottom) average two training seeds after 6000 updates in Base and 12000 in Extended.
    Horizontal labels give the midtraining tokens used during pretraining (top line) and midtraining examples per RL update (bottom line).}
    \label{fig:e2e-pretraining-replay}
\end{figure}

\begin{table}[H]
    \centering
    \begingroup
\small
\setlength{\tabcolsep}{5pt}
\begin{tabular}{lrrrrrr}
\toprule
& \multicolumn{2}{c}{Midtraining} & \multicolumn{2}{c}{Pretraining loss $\downarrow$} & \multicolumn{2}{c}{Benchmark mean (\%)} \\
\cmidrule(lr){2-3}\cmidrule(lr){4-5}\cmidrule(lr){6-7}
Setting & \shortstack{Tokens during\\pretraining} & \shortstack{Examples per\\RL update} & Without & With & Without & With \\
\midrule
Base & 534M & 8 & 2.570 & 2.542 & 41.66 & 42.13 \\
Base & 534M & 32 & 2.575 & 2.535 & 41.65 & 41.98 \\
Base & 2.14B & 8 & 2.572 & 2.545 & 41.93 & 42.08 \\
Base & 2.14B & 32 & 2.571 & 2.537 & 42.01 & 41.85 \\
\midrule
Extended & 534M & 8 & 2.613 & 2.546 & 41.60 & 42.25 \\
Extended & 534M & 32 & 2.617 & 2.537 & 41.24 & 41.90 \\
Extended & 2.14B & 8 & 2.605 & 2.549 & 41.63 & 42.00 \\
Extended & 2.14B & 32 & 2.601 & 2.539 & 41.67 & 41.95 \\
\bottomrule
\end{tabular}
\endgroup

    \caption{\textbf{Retention after RL with and without added pretraining data.}
    Pretraining loss is measured in nats and weights held-out ClimbMix and Dolmino by 75/25; the benchmark mean averages the five tasks in \pref{app:e2e-evaluation}.
    Scores use the same final checkpoints as \pref{fig:e2e-pretraining-replay} and average two training seeds.}
    \label{tab:reasoning-replay-retention}
\end{table}

\endgroup

\clearpage
\subsection{Combining Midtraining and RL Losses}
\label{app:e2e-sftrl}

\begingroup

We compare two ways of combining midtraining data and RL generations in a training update.
Unified training normalizes the combined weighted loss by the total number of response tokens, so the relative contribution of the sources follows the token mixture.
The alternative first normalizes the loss from each source separately, then multiplies the midtraining loss by a fixed coefficient $\alpha$ before adding it to the RL loss.
\pref{app:data-mixing} defines both objectives.
This comparison tests how the choice of loss aggregation affects learning when the model receives the same kinds and amounts of data.

Within each comparison, we hold the initialization, training data, example counts, optimizer, and RL update budget fixed.
Unified training uses unit weight for midtraining tokens; the separately normalized alternative uses $\alpha\in\{0.1,0.5\}$.
For the 2.14B-token initialization in Extended, we additionally test $\alpha=1$ at all five midtraining example counts and evaluate the same models on Extended generalization.
\pref{fig:reasoning-loss-base,fig:reasoning-loss-extended,fig:reasoning-loss-generalization} show the comparisons for initializations trained with 534M, 1.07B, and 2.14B midtraining tokens during pretraining.
Every displayed objective has evaluations through the full training horizon: 6000 RL updates in Base and 12000 in Extended.
We estimate accuracy and coverage separately for each of two RL training seeds, then average their scores at each update.

The relative performance depends on the task, initialization, and amount of midtraining data included during RL.
Separately normalized losses often achieve higher accuracy and coverage in Extended, and the better coefficient varies across configurations.
Across the configurations compared here, unified training attains the highest observed final accuracy and coverage in Base and Extended generalization.
The $\alpha=1$ results do not improve the highest final accuracy or coverage observed on either Extended test set.
However, unified training does not consistently outperform the alternative when the initialization and data mixture are matched.

\begin{figure}[H]
    \centering
    \includegraphics[width=\linewidth]{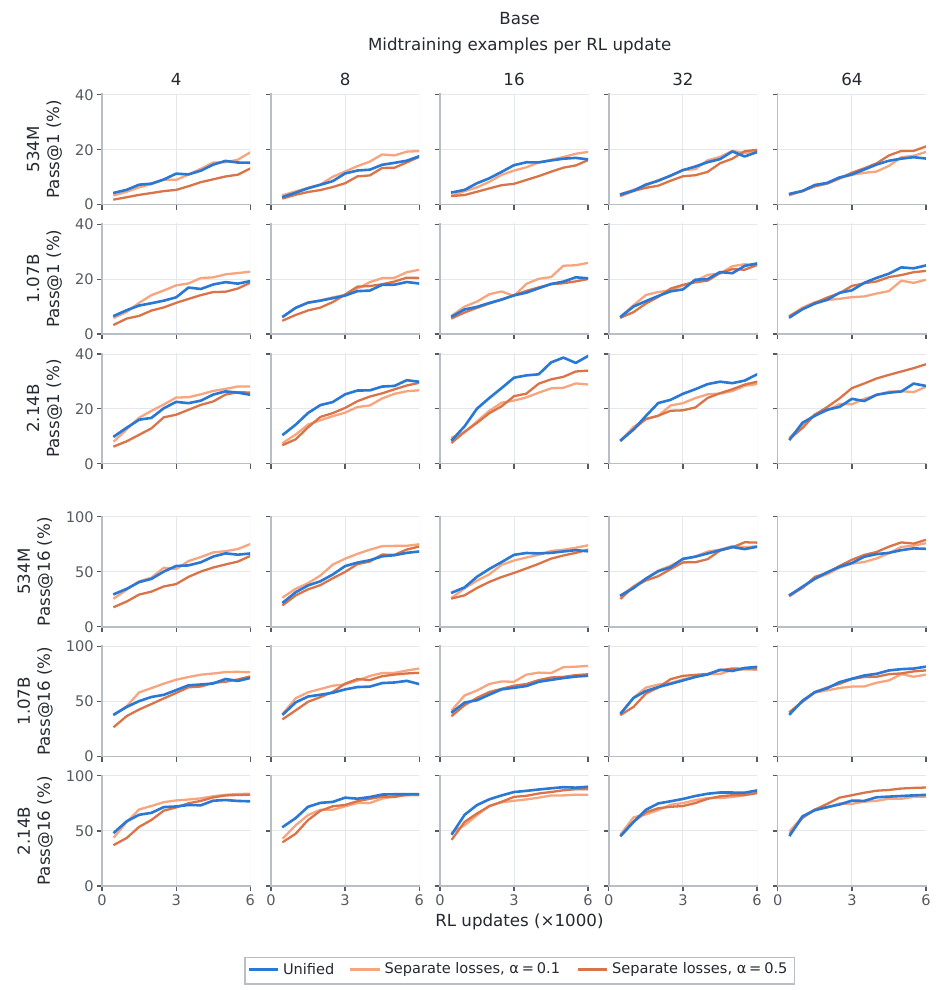}
    \caption{\textbf{Loss aggregation in Base.}
    The upper three rows show pass@1, and the lower three show pass@16, through 6000 RL updates.
    Within each group, rows vary the number of midtraining tokens used during pretraining; columns vary the number of midtraining examples added to each RL update.
    Each panel compares unified training with two separately normalized objectives, holding the initialization and example counts fixed.
    Curves average the scores of two RL training seeds at each update.}
    \label{fig:reasoning-loss-base}
\end{figure}

\clearpage
\begin{figure}[H]
    \centering
    \includegraphics[width=\linewidth]{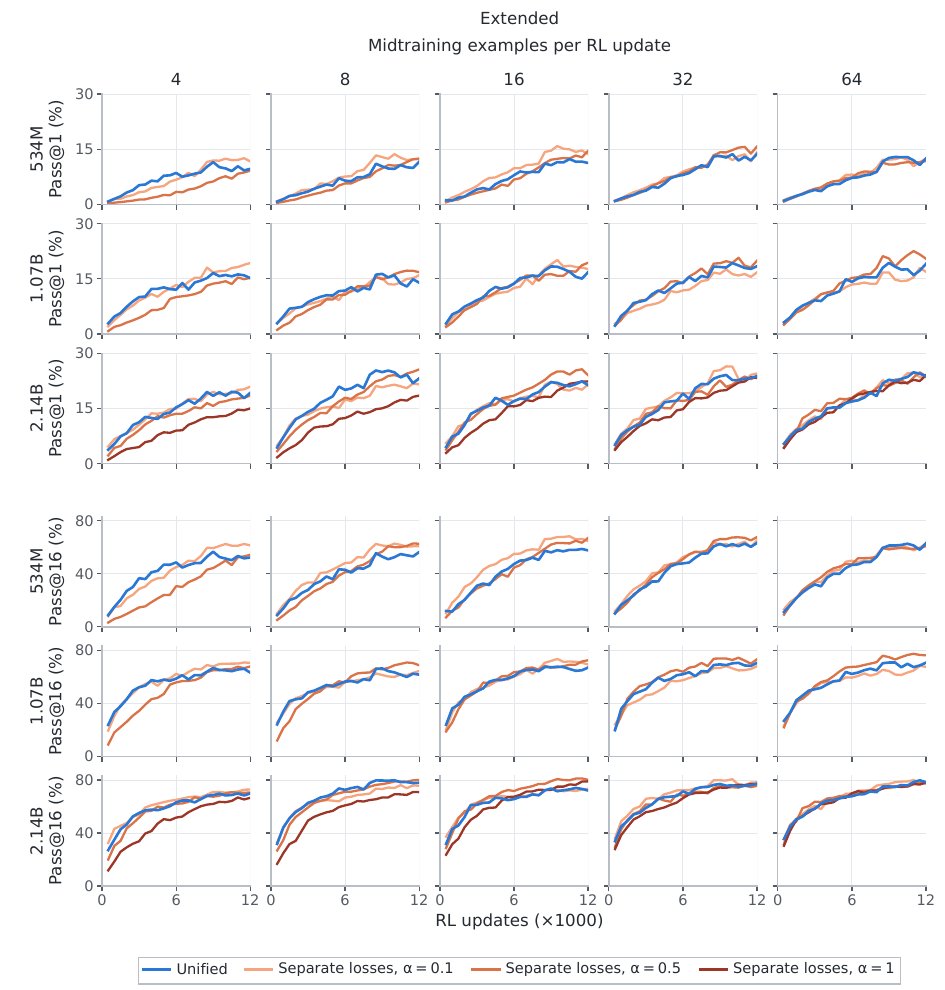}
    \caption{\textbf{Loss aggregation in Extended.}
    Pass@1 (upper three rows) and pass@16 (lower three rows) are measured on the primary Extended test set through 12000 RL updates.
    Rows and columns follow \pref{fig:reasoning-loss-base}, using the same three initializations and five midtraining example counts.
    The $\alpha=1$ curves are included for the 2.14B-token initialization.
    Each panel holds the initialization and data mixture fixed while varying the loss aggregation.
    Curves average the scores of two RL training seeds.}
    \label{fig:reasoning-loss-extended}
\end{figure}

\clearpage
\begin{figure}[H]
    \centering
    \includegraphics[width=\linewidth]{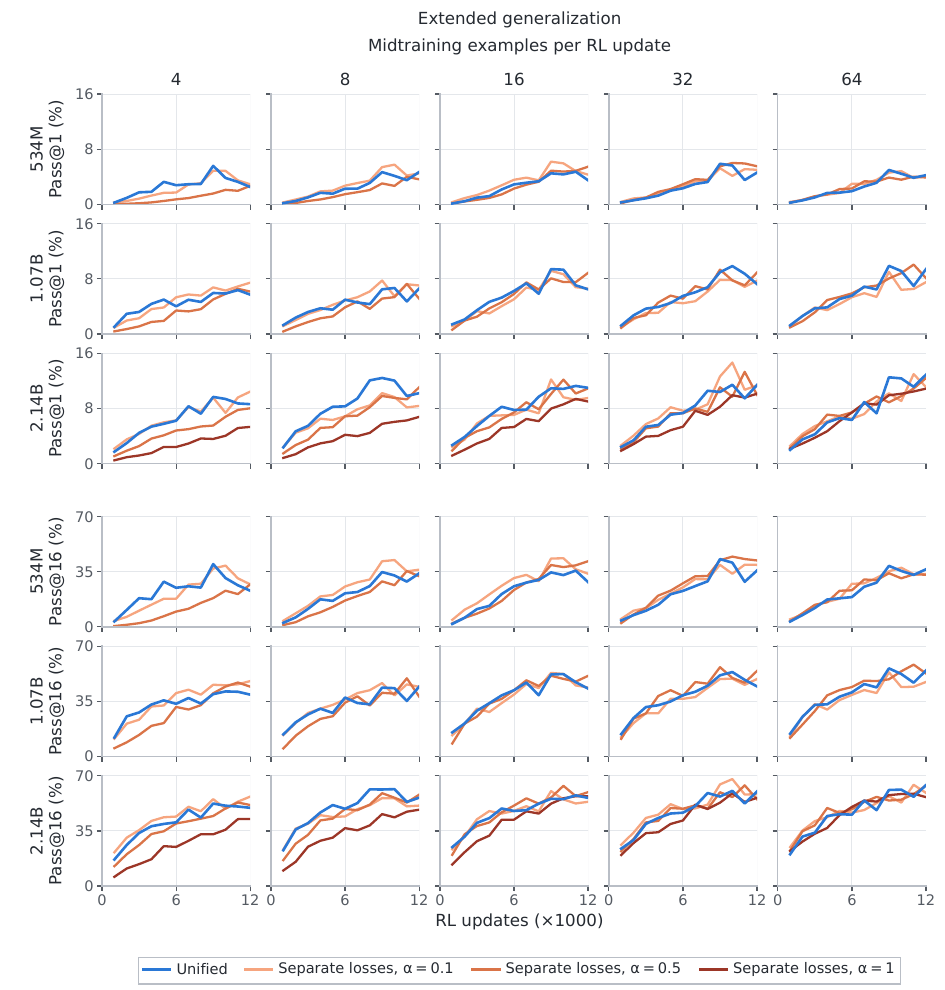}
    \caption{\textbf{Loss aggregation and Extended generalization.}
    The models in \pref{fig:reasoning-loss-extended} are evaluated on 12-person puzzles, a size absent from RL training.
    The upper three rows show pass@1 and the lower three show pass@16, with the same arrangement of initializations and midtraining example counts.
    As in \pref{fig:reasoning-loss-extended}, $\alpha=1$ is evaluated for the 2.14B-token initialization.
    Curves average the scores of two RL training seeds through 12000 updates.}
    \label{fig:reasoning-loss-generalization}
\end{figure}

\clearpage
\paragraph{Smaller pretraining mixtures}
The remaining figures examine the two initializations trained with less midtraining data during pretraining: 133M and 267M tokens.
For Base, we evaluate both initializations through 6000 RL updates.
For Extended and Extended generalization, we report the 267M-token initialization through 12000 updates.
Within each panel, all three objectives are shown over the same training horizon, so their endpoints are directly comparable.

\begin{figure}[H]
    \centering
    \includegraphics[width=\linewidth]{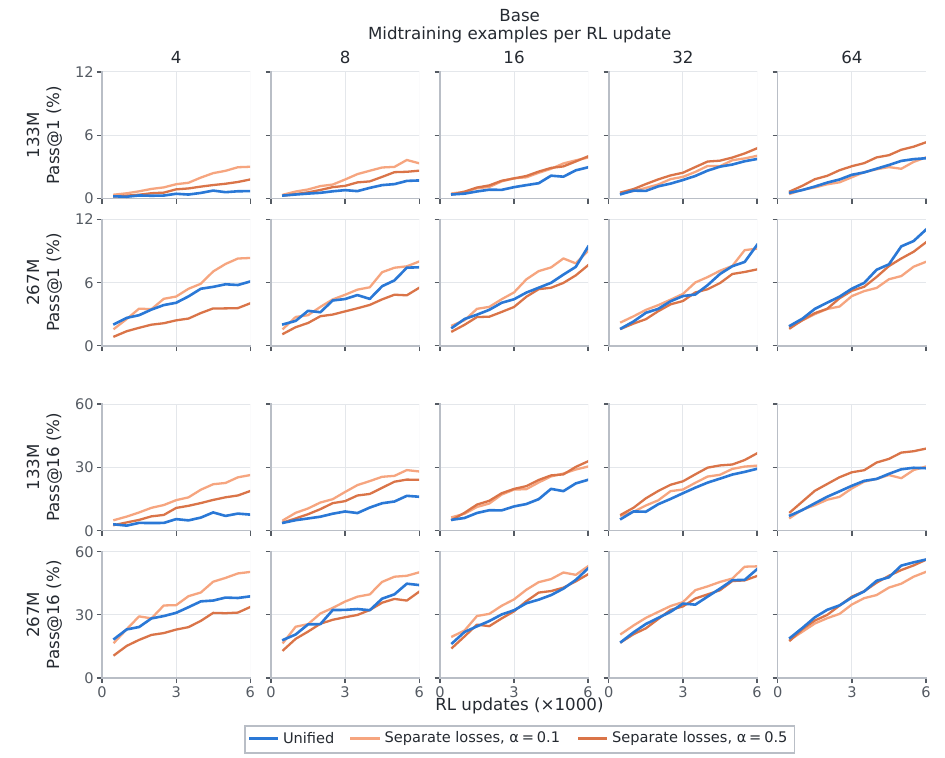}
    \caption{\textbf{Loss aggregation with smaller pretraining mixtures in Base.}
    The upper two rows show pass@1 and the lower two show pass@16 for the 133M- and 267M-token initializations, through 6000 RL updates.
    Columns vary the number of midtraining examples added to each RL update.
    Each panel compares the three objectives with the same initialization and example counts, averaging scores over two RL training seeds.}
    \label{fig:reasoning-loss-base-low}
\end{figure}

\clearpage
\begin{figure}[H]
    \centering
    \includegraphics[width=\linewidth]{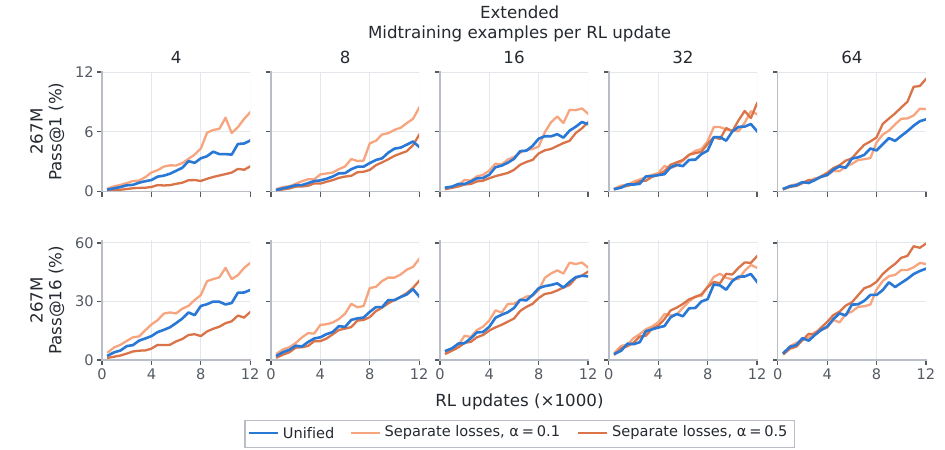}
    \caption{\textbf{Loss aggregation with a smaller pretraining mixture in Extended.}
    The top row shows pass@1 and the bottom row shows pass@16 for the 267M-token initialization through 12000 RL updates.
    Columns vary the number of midtraining examples per RL update, with the initialization and example counts held fixed within each comparison.
    Curves average scores over two RL training seeds.}
    \label{fig:reasoning-loss-extended-low}
\end{figure}

\begin{figure}[H]
    \centering
    \includegraphics[width=\linewidth]{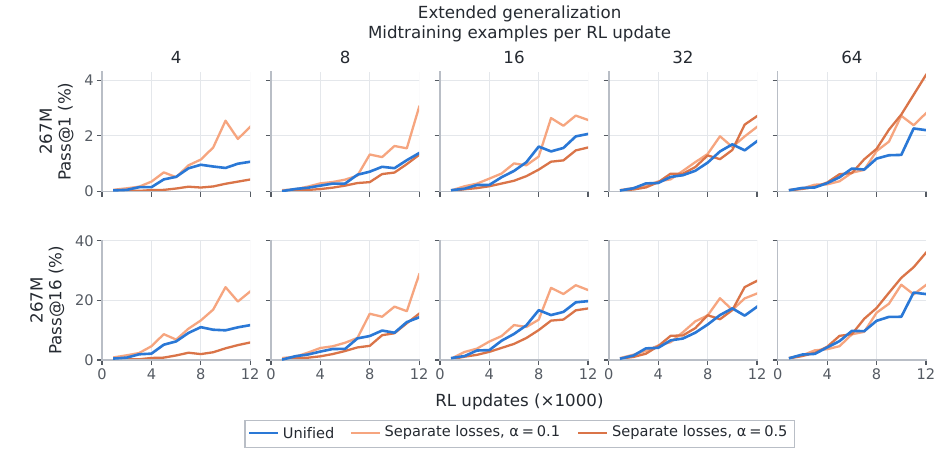}
    \caption{\textbf{Extended generalization with a smaller pretraining mixture.}
    The models in \pref{fig:reasoning-loss-extended-low} are evaluated on the 12-person test set through 12000 RL updates.
    The top row shows pass@1 and the bottom row shows pass@16 for the 267M-token initialization.
    Each curve averages the per-model score over two RL training seeds, without pooling generations from different models.}
    \label{fig:reasoning-loss-generalization-low}
\end{figure}

\endgroup

\clearpage
\begingroup

\subsection{Retention before RL}
\label{app:e2e-retention}

For a for focus study on the effect of unified training before the RL stage, we evaluate the models on pretraining data and general benchmarks before RL.
We include all five unified initializations, all six staged midtraining configurations, and the model trained on pretraining data alone.
Unified initializations learn from the reasoning corpus while retaining the pretraining data stream; staged initializations learn from this corpus in a separate phase after pretraining.
Evaluating at this point measures retention before any further changes caused by RL.
The same initializations are used in Base and Extended, so one set of retention evaluations applies to both settings.

\pref{tab:reasoning-retention} reports held-out language-model loss on ClimbMix and Dolmino, together with scores on the five general benchmarks specified in \pref{app:e2e-evaluation}.
The mean loss weights the two corpora by their pretraining proportions, while the mean benchmark score gives equal weight to the five tasks.
Lower loss and higher benchmark scores indicate better retention relative to the model trained on pretraining data alone.

Across the unified mixtures, mean pretraining loss remains between 2.537 and 2.540, close to the reference value of 2.537, and the mean benchmark score remains within 0.52 percentage points of the reference.
Staged midtraining increases loss on both pretraining corpora and lowers the mean benchmark score at every tested epoch budget and learning rate.
Using the lower learning rate reduces both changes at each epoch budget, but does not eliminate them.
The one-epoch configuration at a peak learning rate of $5\times10^{-5}$ has both the lowest mean loss and the highest mean benchmark score among staged variants, so we use it as the retention reference in \pref{fig:e2e-cpt-retention}.
The two-epoch configuration at $2\times10^{-4}$, selected for the best final RL accuracy in \pref{app:e2e-staged-sweep}, has higher mean loss and a lower mean benchmark score.

\begin{table}[H]
    \centering
    \begingroup
\small
\setlength{\tabcolsep}{3pt}
\renewcommand{\arraystretch}{1.12}
\begin{tabular}{l*{9}{r}}
\toprule
 & \multicolumn{3}{c}{Held-out loss $\downarrow$} & \multicolumn{6}{c}{Benchmark score (\%)} \\
\cmidrule(lr){2-4}\cmidrule(lr){5-10}
Midtraining & ClimbMix & Dolmino & Mean & MMLU & ARC-E & HellaSwag & TQA & WinoG & Mean \\
\midrule
None & 2.784 & 1.794 & 2.537 & 28.65 & 55.09 & 35.39 & 39.15 & 53.59 & 42.37 \\
\midrule
\multicolumn{10}{l}{\textit{Unified: mixed during pretraining}} \\
133M tokens & 2.784 & 1.795 & 2.537 & 28.62 & 54.67 & 35.55 & 39.09 & 53.20 & 42.23 \\
267M tokens & 2.786 & 1.796 & 2.538 & 28.79 & 54.97 & 35.17 & 39.17 & 53.99 & 42.42 \\
534M tokens & 2.785 & 1.797 & 2.538 & 28.22 & 54.67 & 35.67 & 39.57 & 51.14 & 41.85 \\
1.07B tokens & 2.787 & 1.797 & 2.539 & 28.30 & 54.71 & 35.31 & 39.43 & 52.96 & 42.14 \\
2.14B tokens & 2.788 & 1.798 & 2.540 & 27.85 & 54.88 & 35.47 & 39.46 & 52.33 & 42.00 \\
\midrule
\multicolumn{10}{l}{\textit{Staged: midtraining at peak LR $2\times10^{-4}$}} \\
1 epoch (510M) & 12.421 & 11.340 & 12.150 & 24.98 & 24.92 & 26.64 & 48.09 & 50.36 & 35.00 \\
2 epochs (1.02B) & 12.426 & 12.147 & 12.356 & 25.34 & 24.96 & 26.49 & 48.57 & 49.88 & 35.05 \\
4 epochs (2.04B) & 15.324 & 14.443 & 15.104 & 27.12 & 25.04 & 26.27 & 47.68 & 51.93 & 35.61 \\
\midrule
\multicolumn{10}{l}{\textit{Staged: midtraining at peak LR $5\times10^{-5}$}} \\
\rowcolor{hl}
1 epoch (510M) & 4.272 & 3.722 & 4.135 & 23.59 & 41.25 & 32.86 & 42.35 & 52.41 & 38.49 \\
2 epochs (1.02B) & 5.362 & 5.131 & 5.305 & 23.43 & 36.66 & 31.18 & 42.82 & 50.91 & 37.00 \\
4 epochs (2.04B) & 7.005 & 7.203 & 7.054 & 23.91 & 33.50 & 29.08 & 46.23 & 49.41 & 36.42 \\
\bottomrule
\end{tabular}
\endgroup

    \caption{\textbf{Pretraining retention across all initializations.}
    Every row evaluates one initialization before RL; these results are shared by the Base and Extended experiments.
    Loss is in nats per token, with a mean weighted by 75\% ClimbMix and 25\% Dolmino.
    Benchmark scores are percentages, with an unweighted mean over the five tasks.
    ARC-E denotes ARC-Easy normalized accuracy, TQA denotes TruthfulQA MC2, and WinoG denotes WinoGrande accuracy; HellaSwag uses normalized accuracy and MMLU uses accuracy.
    Midtraining token counts include prompts and responses; staged configurations also indicate the number of passes over the corpus.
    The shaded row is the staged initialization with the best retention on both aggregate measures, used in \pref{fig:e2e-cpt-retention}.}
    \label{tab:reasoning-retention}
\end{table}

The individual benchmark scores also help interpret the aggregate result.
Staged midtraining improves TruthfulQA MC2 relative to pretraining alone, while reducing MMLU, ARC-Easy, HellaSwag, and WinoGrande performance.
Thus, the retention advantage shown in \pref{fig:e2e-cpt-retention} concerns mean benchmark performance and pretraining loss; it does not hold for every individual benchmark.

\endgroup

\clearpage
\section{Details of Training with Real-world Midtraining Data}
\label{app:math}

This section describes the unified training with real-world midtraining experiments in \pref{sec:real}.
We first give the common training and evaluation setup, then examine the amount, timing, and composition of midtraining data and the choice of RL objective.
The final two subsections study subsequent learning on coding and WebInstruct tasks.
Throughout the appendix, comparisons at the end of training use each run's final checkpoint; we do not select a different checkpoint for each metric.

\subsection{Experimental Setup}
\label{app:math-design}

\paragraph{Base model}
We use a model with 756819456 parameters, 24 layers, hidden width 1536, 12 attention and key/value heads, and context length 2048.
We train it from scratch on 354.53B ClimbMix pretraining tokens~\citep{diao2025climb}, followed by 17.73B OLMo-3 Dolmino midtraining tokens~\citep{olmo2025olmo3}. We do not perform study on unified training during pretraining due to computational constraints.
We refer to the resulting checkpoint as the base model.
\pref{tab:real-base-hyperparameters} gives the pretraining and Dolmino midtraining settings.

\begin{table}[htbp]
\centering
\small
\renewcommand{\arraystretch}{1.08}
\begin{tabular}{@{}lcc@{}}
\toprule
Setting & Pretraining & Midtraining \\
\midrule
Optimizer & \multicolumn{2}{c}{AdamW} \\
Adam $(\beta_1,\beta_2)$; $\epsilon$ & \multicolumn{2}{c}{$(0.9,0.95)$; $10^{-8}$} \\
Weight decay; gradient clipping norm & \multicolumn{2}{c}{0.1; 1} \\
Global batch size (sequences) & \multicolumn{2}{c}{2048} \\
Sequence length (tokens) & \multicolumn{2}{c}{2048} \\
Peak / minimum learning rate & \multicolumn{2}{c}{$4\times10^{-4}$ / $4\times10^{-5}$} \\
Learning-rate schedule & \multicolumn{2}{c}{Warmup, constant, cosine decay} \\
Warmup updates & 845 & 177 \\
Final decay updates & 16905 & 845 \\
Total updates & 84527 & 4226 \\
\bottomrule
\end{tabular}
\caption{\textbf{Pretraining and midtraining hyperparameters for the 0.76B base model.}
The midtraining column describes training on the Dolmino corpus.}
\label{tab:real-base-hyperparameters}
\end{table}

\paragraph{Mathematical RL and additional midtraining data}
Math12K provides the mathematical RL tasks, with 11977 prompts remaining after length filtering.
We use the reward function from veRL~\citep{Sheng_2025}.
The additional midtraining corpus contains 240M tokens from 24 components, including web text, mathematics, code, question answering, and reasoning traces.
Unified training uses this corpus during RL, whereas staged training uses it before RL.
Examples from the corpus are trained as document continuation with unit token weights, except in the code-data replacement experiment described in \pref{app:math-content-details}.

\paragraph{Data mixing}
A mixing percentage specifies the target fraction of midtraining tokens in the combined midtraining and RL response tokens.
We convert this target into example counts using lengths estimated before training, so the realized token fraction can vary as RL response lengths change.
\pref{app:data-mixing} gives the conversion, batch construction, and schedule definitions; \pref{app:compute-accounting} explains how we count training exposure.

\paragraph{Training}
The RL-only baseline uses the unified trainer with no midtraining data.
It provides a control for adding midtraining data while keeping the trainer fixed.
The staged reference instead trains the base model on the additional 240M-token corpus before beginning on-policy \maxrl with baseline subtraction.

All mathematical RL runs use 12000 optimizer updates, with 512 RL responses per update and 16 responses generated per prompt.
Unified training adds midtraining examples to this RL batch according to the configured token proportion.
Both approaches initialize a fresh AdamW optimizer for RL, using learning rate $10^{-6}$, betas $(0.9,0.999)$, and zero weight decay, with no warmup or learning-rate decay.

\paragraph{Response weighting and loss normalization}

For RL responses, we compute token likelihoods under the current policy's top-$k$/top-$p$ distribution.
Tokens outside this distribution's support are excluded from both the loss and the token count used to normalize it.
Loss normalization uses the total number of remaining RL response tokens and midtraining tokens, including response tokens from unsuccessful generations.
Midtraining token likelihoods use the full vocabulary.

Generation and optimization run asynchronously in unified training.
The trainer publishes updated model weights every four updates, and the buffer keeps the newest generation round from each generator even when that round is older than four updates.
\pref{app:math-zero-mixing} discusses the resulting data staleness.
The optimizer, batch, and generation settings are listed in \pref{tab:real-math-hyperparameters}.

\begin{table}[htbp]
\centering
\small
\renewcommand{\arraystretch}{1.08}
\begin{tabular}{@{}lcc@{}}
\toprule
Setting & Unified & Staged RL \\
\midrule
Optimizer & AdamW & AdamW \\
Adam $(\beta_1,\beta_2)$ & $(0.9,0.999)$ & $(0.9,0.999)$ \\
Adam $\epsilon$ & $10^{-8}$ & $10^{-8}$ \\
Weight decay & 0 & 0 \\
Gradient clipping norm & 1 & 1 \\
Learning rate & $10^{-6}$ & $10^{-6}$ \\
Learning-rate schedule & Constant, no warmup & Constant, no warmup \\
Optimizer updates & 12000 & 12000 \\
RL responses per update & 512 & 512 \\
Responses per prompt & 16 & 16 \\
RL data collection & Asynchronous generators & 32 fresh prompts per update \\
Prompt + response limit (tokens) & 2048 & 2048 \\
Maximum response length (tokens) & 1024 & 1024 \\
Temperature / top-$k$ / top-$p$ & 1 / 50 / 0.95 & 1 / 50 / 0.95 \\
Baseline subtraction & No & Yes \\
\bottomrule
\end{tabular}
\caption{RL hyperparameters.}
\label{tab:real-math-hyperparameters}
\end{table}

\paragraph{Task evaluation}
We evaluate mathematical performance on MATH-500~\citep{hendrycks2021math,lightman2023verify} and GSM8K~\citep{cobbe2021training}, and coding performance on the EvalPlus HumanEval+ and MBPP+ suites~\citep{chen2021evaluating,austin2021program,liu2023evalplus}.\footnote{MBPP+ was released through EvalPlus: \url{https://github.com/evalplus/evalplus/releases/tag/v0.2.0}.}
Evaluation covers all 500 MATH-500, 1319 GSM8K, 164 HumanEval+, and 378 MBPP+ problems.
For each metric, \emph{Math} is the equally weighted average of the two mathematical benchmark scores, and \emph{Coding} is the corresponding average of the two coding benchmark scores.

We estimate pass@1 from 16 sampled responses per problem.
To measure coverage, we use a separate evaluation with 256 responses per mathematical problem to estimate pass@32, pass@64, and pass@256, and 32 responses per coding problem to estimate pass@32.
All estimates use the pass@$k$ estimator in \pref{app:passk-estimator}~\citep{chen2021evaluating}.
Sampling uses temperature 1, top-$k$ 50, top-$p$ 0.95, and context length 2048, with generation caps of 1024 tokens for math and 768 for coding.

\paragraph{General-knowledge evaluation}
SciQ~\citep{welbl2017sciq} and ARC-Challenge~\citep{clark2018arc} are evaluated by greedy decoding with the answer choices included in the prompt.
We allow at most eight generated tokens and score the leading answer label strictly.
MMLU~\citep{hendrycks2021mmlu} uses zero-shot answer-choice likelihoods.
Mathematics, coding, SciQ, and ARC-Challenge are evaluated every 1000 updates; MMLU and held-out corpus loss are evaluated every 2000 updates.

\paragraph{Held-out loss}
To measure retention of the midtraining distribution, we evaluate token loss separately on each of the 24 corpus components and give each component equal weight in the reported average.
The evaluation contains 3608122 scored tokens in total.
We report both absolute loss and its change from the base model, whose loss is 1.4352 nats/token; a negative change indicates lower loss than the base model.
The evaluation documents are excluded from the midtraining corpus.

\paragraph{Base-model performance}
\pref{tab:real-base-performance} reports the base model's performance before mathematical RL, using the evaluation protocol above.

\begin{table}[H]
    \centering
    \small
    \setlength{\tabcolsep}{3pt}
    \begin{minipage}[t]{0.52\linewidth}
        \vspace{0pt}
        \begin{tabular*}{\linewidth}{@{\extracolsep{\fill}}lrr@{}}
            \toprule
            Benchmark & Pass@1 & Pass@32 \\
            \midrule
            MATH-500 & 6.54 & 43.76 \\
            GSM8K & 17.84 & 71.63 \\
            Math (average) & 12.19 & 57.69 \\
            \midrule
            HumanEval+ & 11.97 & 32.32 \\
            MBPP+ & 21.05 & 57.14 \\
            Coding (average) & 16.51 & 44.73 \\
            \bottomrule
        \end{tabular*}
    \end{minipage}\hfill
    \begin{minipage}[t]{0.43\linewidth}
        \vspace{0pt}
        \begin{tabular*}{\linewidth}{@{\extracolsep{\fill}}lr@{}}
            \toprule
            Metric & Value \\
            \midrule
            MMLU accuracy & 37.00 \\
            SciQ accuracy & 51.00 \\
            ARC-Challenge accuracy & 31.23 \\
            \midrule
            Held-out loss $\downarrow$ & 1.4352 \\
            \bottomrule
        \end{tabular*}
    \end{minipage}
    \caption{\textbf{Base-model performance before mathematical RL.}
    Scores are percentages, except held-out loss in nats/token.
    Pass@1 uses 16 responses per problem; pass@32 uses 256 responses for math and 32 for coding.
    Math and Coding are the equally weighted benchmark averages defined in \pref{app:math-design}.}
    \label{tab:real-base-performance}
\end{table}

\newpage
\subsection{Effect of the Midtraining Data Proportion}
\label{app:math-dose-details}

\begin{figure}[!htbp]
    \centering
    \includegraphics[width=\linewidth]{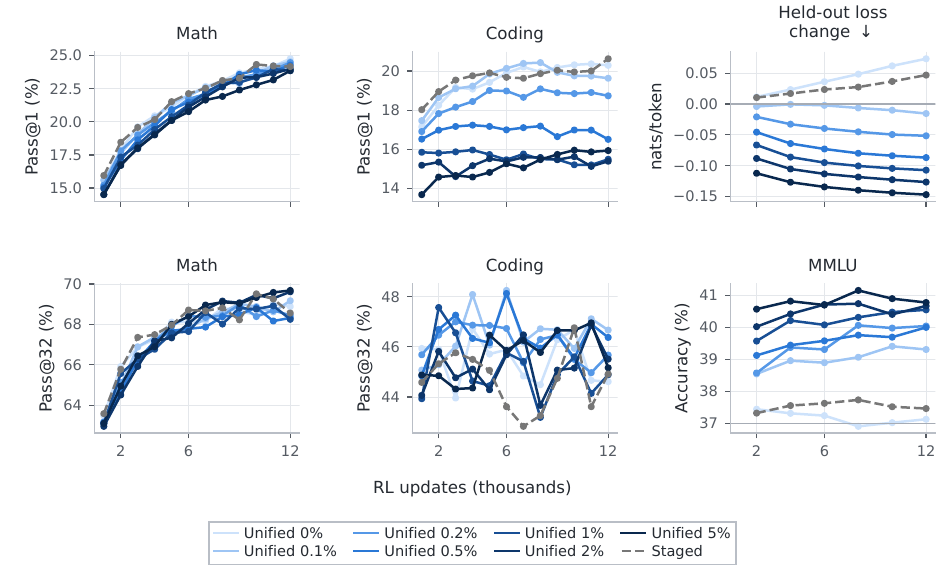}
    \caption{\textbf{Training curves for different midtraining data proportions.}
    Each unified curve uses a constant configured proportion of 0, 0.1, 0.2, 0.5, 1, 2, or 5\%; the dashed curves show staged training.
    Math and Coding average the two benchmarks in each domain, as defined in \pref{app:math-design}, which also gives the evaluation sample counts.}
    \label{fig:math-trajectories}
\end{figure}

This section describes comparisons that vary the amount of midtraining data while keeping its configured proportion constant throughout mathematical RL.
We evaluate proportions of 0, 0.1, 0.2, 0.5, 1, 2, and 5\%, where 0\% is the RL-only with unified infrastructure ablation described above.
\pref{fig:math-trajectories} shows the training curves for all seven proportions.
The tables below report final results for the full sweep, together with the base model and staged reference.

\begin{table}[H]
    \centering
    \small
    \renewcommand{\arraystretch}{0.9}
    \setlength{\tabcolsep}{3pt}
    \begin{tabular*}{\linewidth}{@{\extracolsep{\fill}}lrrrrrr@{}}
        \toprule
        Mixing & \shortstack{MATH\\pass@1} & \shortstack{GSM\\pass@1} & \shortstack{MATH\\pass@64} & \shortstack{GSM\\pass@64} & \shortstack{MATH\\pass@256} & \shortstack{GSM\\pass@256} \\
        \midrule
        0\% & 16.83 & 32.68 & 66.73 & 83.88 & 78.20 & 91.81 \\
        0.1\% & 16.90 & 32.14 & 65.75 & 85.56 & 77.00 & 93.03 \\
        0.2\% & 16.56 & 32.35 & 65.03 & 84.74 & 76.20 & 92.49 \\
        0.5\% & 16.83 & 31.71 & 65.60 & 84.06 & 77.20 & 91.81 \\
        1\% & 16.11 & 31.61 & 65.85 & 83.71 & 76.00 & 91.74 \\
        2\% & 16.68 & 31.30 & 67.51 & 84.90 & 78.40 & 93.03 \\
        5\% & 16.91 & 30.80 & 67.45 & 85.15 & 77.20 & 92.72 \\
        \midrule
        Base model & 6.54 & 17.84 & 53.67 & 80.15 & 70.20 & 91.51 \\
        Staged & 16.98 & 31.35 & 66.17 & 84.50 & 78.00 & 92.34 \\
        \bottomrule
    \end{tabular*}
    \caption{\textbf{Mathematical accuracy and coverage at different midtraining proportions.}
    The Base model row reports performance before RL; all other rows report final results after RL.
    MATH and GSM denote MATH-500 and GSM8K, and all scores are percentages.
    Pass@1 is estimated from 16 responses per problem; pass@64 and pass@256 use a separate evaluation with 256 responses per problem.}
    \label{tab:math-dose-full}
\end{table}

\begin{table}[H]
    \centering
    \small
    \renewcommand{\arraystretch}{0.9}
    \setlength{\tabcolsep}{3pt}
    \begin{tabular*}{\linewidth}{@{\extracolsep{\fill}}lrrrrrr@{}}
        \toprule
        Mixing & $\Delta$loss $\downarrow$ & MMLU & SciQ & ARC-C & MBPP+ & HumanEval+ \\
        \midrule
        0\% & $+0.0737$ & 37.12 & 50.20 & 27.90 & 25.41 & 15.17 \\
        0.1\% & $-0.0154$ & 39.30 & 54.60 & 29.69 & 24.82 & 14.44 \\
        0.2\% & $-0.0516$ & 40.04 & 54.30 & 31.06 & 23.71 & 13.76 \\
        0.5\% & $-0.0871$ & 39.99 & 62.00 & 32.34 & 21.05 & 11.97 \\
        1\% & $-0.1076$ & 40.54 & 66.10 & 33.53 & 19.31 & 11.66 \\
        2\% & $-0.1269$ & 40.66 & 67.00 & 34.22 & 19.23 & 11.55 \\
        5\% & $-0.1474$ & 40.77 & 65.10 & 30.12 & 20.97 & 10.90 \\
        \midrule
        Base model & $0$ & 37.00 & 51.00 & 31.23 & 21.05 & 11.97 \\
        Staged & $+0.0472$ & 37.46 & 53.00 & 31.48 & 25.74 & 15.51 \\
        \bottomrule
    \end{tabular*}
    \caption{\textbf{Held-out loss and performance beyond math.}
    The Base model row gives performance before RL, and the remaining rows give final results after mathematical RL.
    Scores are percentages; $\Delta$loss is the change in held-out loss from the base model, in nats/token.
    The coding columns report pass@1 estimated from 16 responses per problem.
    ARC-C denotes ARC-Challenge.}
    \label{tab:math-dose-breadth}
\end{table}

\begin{figure}[H]
    \centering
    \includegraphics[width=\linewidth]{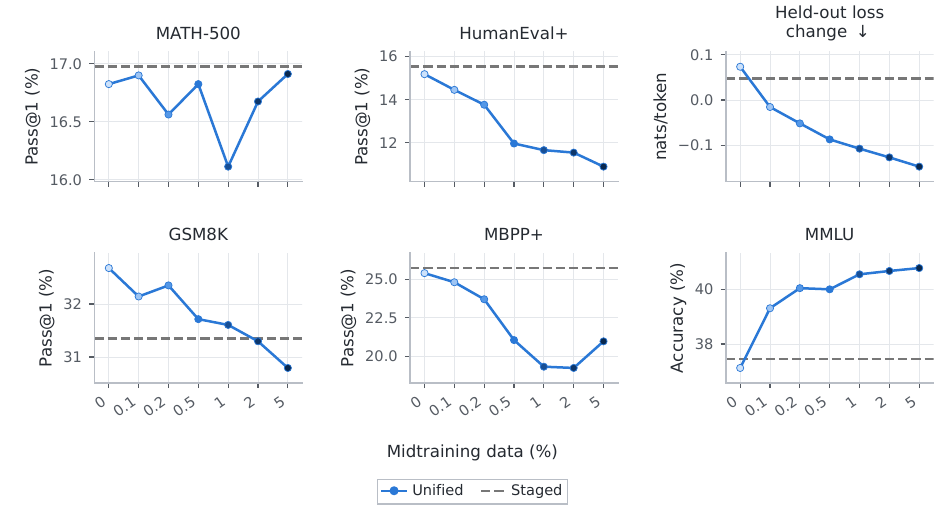}
    \caption{\textbf{Accuracy on individual benchmarks.}
    Final pass@1 on MATH-500, GSM8K, HumanEval+, and MBPP+ is estimated from 16 responses per problem.
    The dashed reference shows staged training.}
    \label{fig:math-highdose}
\end{figure}

\begin{figure}[H]
    \centering
    \includegraphics[width=\linewidth]{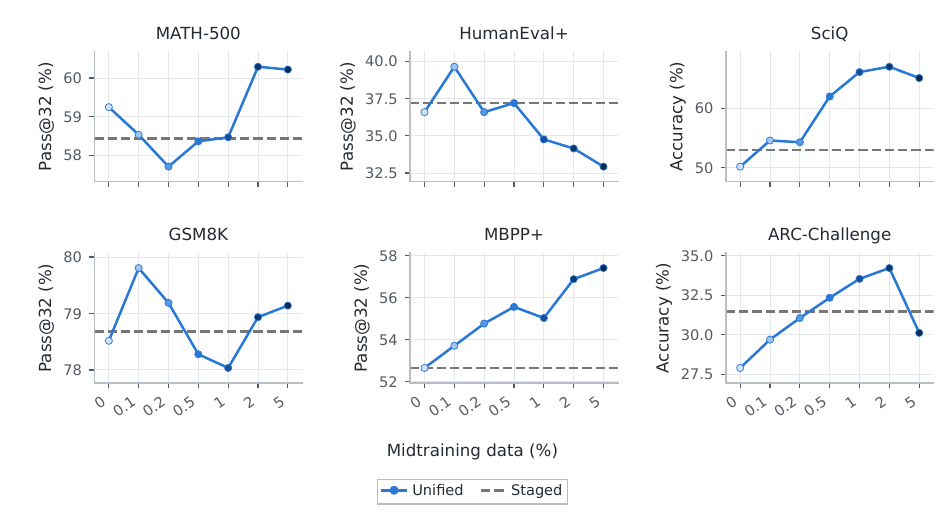}
    \caption{\textbf{Coverage and knowledge accuracy across midtraining proportions.}
    Final results are shown for all seven configured midtraining proportions.
    Mathematical pass@32 is estimated from 256 responses per problem, and coding pass@32 from 32 responses per problem.
    SciQ and ARC-Challenge use the greedy evaluation protocol in \pref{app:math-design}.}
    \label{fig:math-breadth}
\end{figure}

\paragraph{Individual benchmarks}
\pref{fig:math-highdose} and \pref{fig:math-breadth} separate the individual benchmarks that contribute to the reported Math and Coding averages.
Increasing the midtraining proportion consistently lowers held-out corpus loss, but does not produce a corresponding improvement on every benchmark.
Mathematical coverage is nonmonotonic: the highest proportions improve it modestly, while intermediate proportions show no consistent gain over zero mixing.
Coding accuracy remains below the RL-only baseline, whereas Coding coverage exceeds that baseline at every nonzero proportion.
For example, at 5\% mixing, Coding pass@32 is 45.17\%, compared with 44.62\% for the RL-only baseline.
Relative to the base model before RL, Coding accuracy is unchanged at 0.5\% mixing and lower at 1, 2, and 5\% (\pref{tab:real-base-performance}).

\clearpage
\subsection{Data Mixing Schedules}
\label{app:math-schedules}

\paragraph{Schedules with the same average proportion}
We next vary different midtraining schedules, holding the average configured proportion at 0.5\%.
The four schedules keep the proportion constant, decrease it linearly to zero at the end of training, increase it linearly from zero, or decrease it to zero halfway through training.
\pref{app:data-mixing} gives their definitions and explains why equal configured averages need not yield identical realized token proportions.

\begin{figure}[H]
    \centering
    \includegraphics[width=\linewidth]{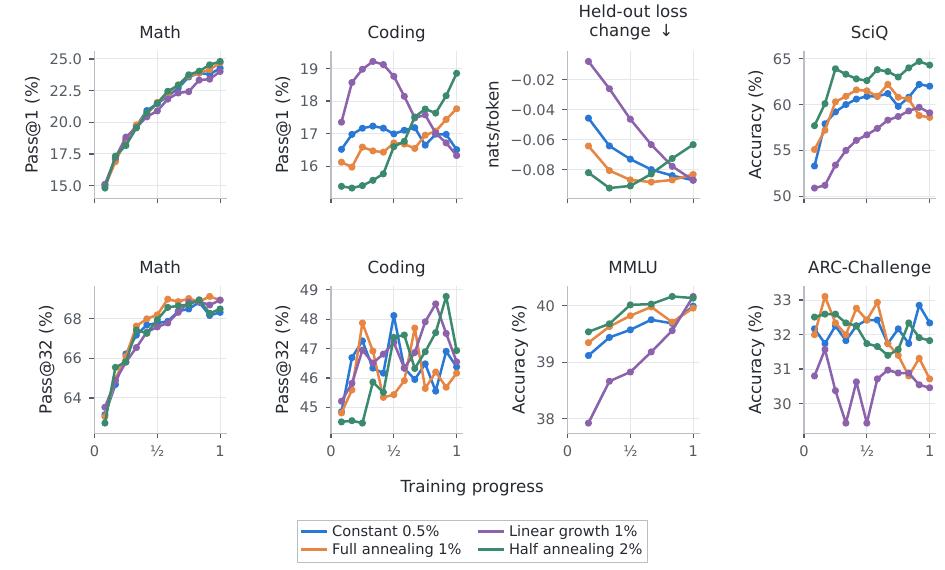}
    \caption{\textbf{Full results for the four mixing schedules.}
    These are the runs in \pref{fig:math-schedules}, with additional panels for Math and Coding accuracy and coverage, held-out loss change, MMLU, SciQ, and ARC-Challenge.
    The vertical line marks the midpoint of training, when the half-annealing schedule stops adding midtraining data.}
    \label{fig:math-schedule-outcomes}
\end{figure}

\pref{fig:math-schedule-outcomes} shows the training curves, and \pref{tab:math-schedule} reports the final results.
Ending midtraining exposure halfway through training allows Coding accuracy to improve further and preserves the SciQ gains, while held-out corpus loss increases.
Mathematical performance and MMLU accuracy are similar across the four schedules: Math pass@1 spans 24.01--24.80\%, Math pass@32 spans 68.32--68.94\%, and MMLU spans 39.96--40.17\%.
Coding pass@32 ranges from 46.16\% to 46.93\%, and the schedules rank differently on Coding pass@1 and pass@32.
ARC-Challenge accuracy also varies with the schedule.
These differences show why the schedule comparison must consider retention, accuracy, and coverage separately.

\begin{table}[H]
    \centering
    \small
    \setlength{\tabcolsep}{3pt}
    \begin{tabular*}{\linewidth}{@{\extracolsep{\fill}}lrrrrrr@{}}
        \toprule
        Schedule & \shortstack{Math\\pass@1} & \shortstack{Math\\pass@32} & \shortstack{Coding\\pass@1} & \shortstack{Coding\\pass@32} & $\Delta$loss $\downarrow$ & MMLU \\
        \midrule
        Constant 0.5\% & 24.27 & 68.32 & 16.51 & 46.38 & $-0.0871$ & 39.99 \\
        Full anneal 1\% & 24.70 & 68.94 & 17.76 & 46.16 & $-0.0833$ & 39.96 \\
        Ramp to 1\% & 24.01 & 68.94 & 16.33 & 46.55 & $-0.0872$ & 40.17 \\
        Half anneal 2\% & 24.80 & 68.49 & 18.85 & 46.93 & $-0.0634$ & 40.14 \\
        \bottomrule
    \end{tabular*}
    \caption{\textbf{Results for the four mixing schedules.} Each schedule has an average configured midtraining proportion of 0.5\%. Scores are percentages, except $\Delta$loss in nats/token. SciQ and ARC-Challenge appear in \pref{fig:math-schedule-outcomes}.}
    \label{tab:math-schedule}
\end{table}

\begin{figure}[H]
    \centering
    \includegraphics[width=\linewidth]{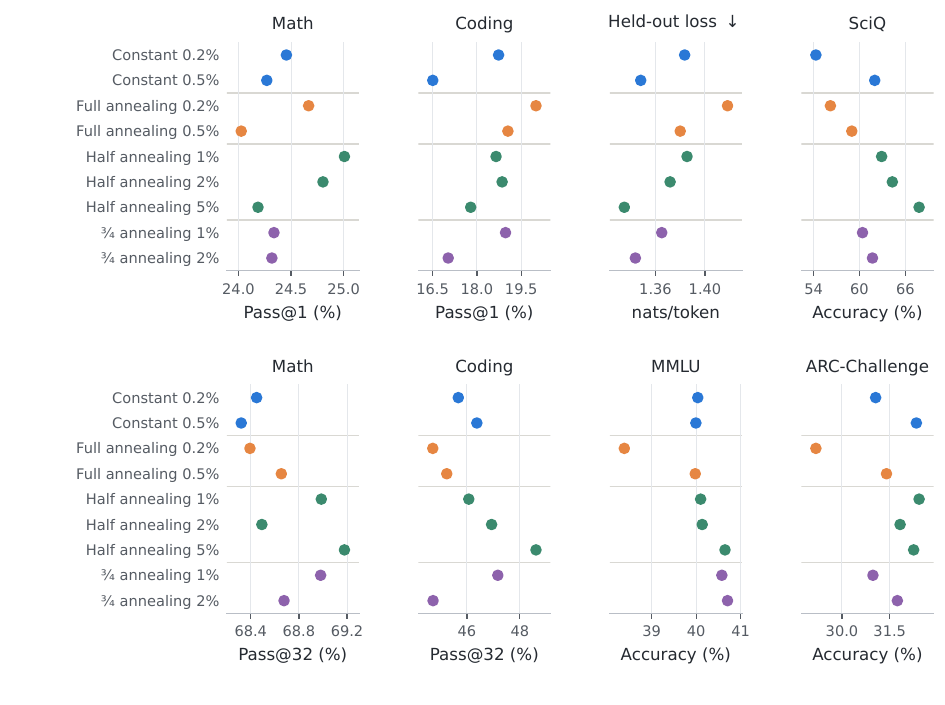}
    \caption{\textbf{Results for additional mixing schedules.}
    Full, half, and three-quarter annealing reach zero at the end, midpoint, and three-quarter point of training, respectively.
    The held-out loss panel reports loss in nats/token, rather than its change from the base model.}
    \label{fig:math-corrected-schedules}
\end{figure}

\clearpage
\paragraph{Additional schedules}
\pref{fig:math-corrected-schedules} extends the comparison to nine schedules: constant 0.2 or 0.5\%, full decay from 0.2 or 0.5\%, half decay from 1, 2, or 5\%, and three-quarter decay from 1 or 2\%.
For each decay schedule, the stated percentage is the initial proportion; the proportion then decreases linearly to zero at the specified point in training.

\clearpage
\subsection{Effect of Code Data}
\label{app:math-content-details}

We examine whether changing the code data in the midtraining corpus can reduce the loss of coding accuracy during mathematical RL.
The corpus has three designated code sources---CraneCode, StackEdu-FIM, and Code Meta-Reasoning---which together account for 24.8\% of its tokens.
We compare the original corpus with two variants.
The first removes these sources and preserves the relative proportions of the remaining data.
The second replaces their allocation with execution-tested Python answers from OpenCodeInstruct~\citep{ahmad2025opencodeinstruct}, using chat formatting and assistant-only loss.
Removing the designated sources does not exclude all possible code from the other components, and replacement changes formatting and loss masking as well as content.

\begin{figure}[H]
    \centering
    \includegraphics[width=\linewidth]{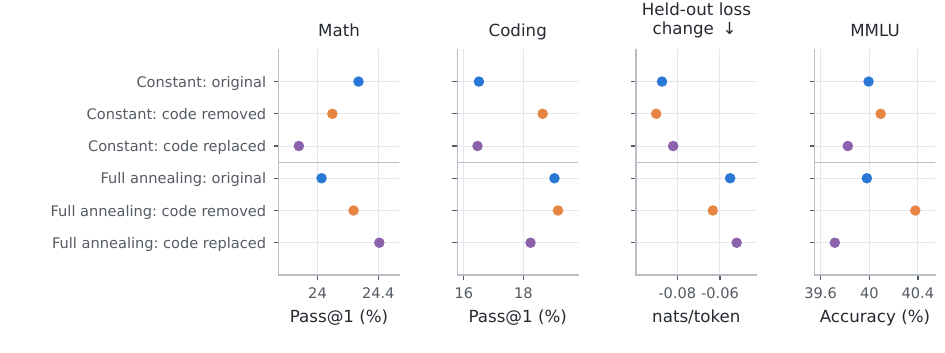}
    \caption{\textbf{Effect of removing or replacing code data.}
    We compare the original corpus with variants that remove its designated code sources or replace them with verified coding examples.
    Each corpus is evaluated with constant 0.5\% mixing and full annealing from 0.5\%.
    Math and Coding are the benchmark averages defined in \pref{app:math-design}.}
    \label{fig:math-content}
\end{figure}

We evaluate each corpus with constant 0.5\% mixing and with a proportion that decays linearly from 0.5\% to zero over the full training run.
Removing the designated code sources improves Coding accuracy under constant mixing, from 16.51\% to 18.65\%, but has little effect under full annealing (\pref{fig:math-content,tab:math-content}).
Replacing these sources does not improve Coding accuracy in either setting, although it improves Coding coverage under full annealing.
The effect of corpus composition therefore depends on the mixing schedule and the metric.

\begin{table}[H]
    \centering
    \small
    \setlength{\tabcolsep}{3pt}
    \begin{tabular*}{\linewidth}{@{\extracolsep{\fill}}lrrrrrr@{}}
        \toprule
        Setting & MATH & GSM & $\Delta$loss $\downarrow$ & SciQ & MBPP+ & HumanEval+ \\
        \midrule
        constant: original & 16.83 & 31.71 & $-0.0871$ & 62.00 & 21.05 & 11.97 \\
        constant: code removed & 16.05 & 32.15 & $-0.0898$ & 61.90 & 23.66 & 13.64 \\
        constant: code replaced & 16.54 & 31.22 & $-0.0819$ & 64.80 & 20.80 & 12.12 \\
        \midrule
        anneal: original & 15.50 & 32.55 & $-0.0552$ & 59.00 & 23.69 & 14.41 \\
        anneal: code removed & 16.40 & 32.07 & $-0.0633$ & 56.80 & 24.01 & 14.33 \\
        anneal: code replaced & 17.00 & 31.81 & $-0.0522$ & 61.70 & 23.38 & 13.11 \\
        \bottomrule
    \end{tabular*}
    \caption{\textbf{Results with different code data.}
    Each corpus variant is evaluated with constant 0.5\% mixing and with full decay from 0.5\% to zero at the end of training.
    The original corpus is compared with variants that remove or replace its designated code sources, as described in \pref{app:math-content-details}.
    Mathematical and coding scores are pass@1 (\%); SciQ reports accuracy (\%).
    MATH denotes MATH-500, GSM denotes GSM8K, and loss change is measured in nats/token relative to the base model.}
    \label{tab:math-content}
\end{table}

\clearpage
\subsection{\maxrl versus \reinforce}
\label{app:math-weighting-details}

In section~\pref{sec:framework}, we outline additional RL objectives that can also be fit into the unified training framework in addition to \maxrl.
We compare \maxrl and \reinforce at constant midtraining proportions of 0.5, 1, 2, and 5\%, and with proportions that decrease linearly from 1 or 2\% to zero halfway through training.
Each comparison pairs the algorithms at the same configured proportion and schedule.

\begin{figure}[H]
    \centering
    \includegraphics[width=\linewidth]{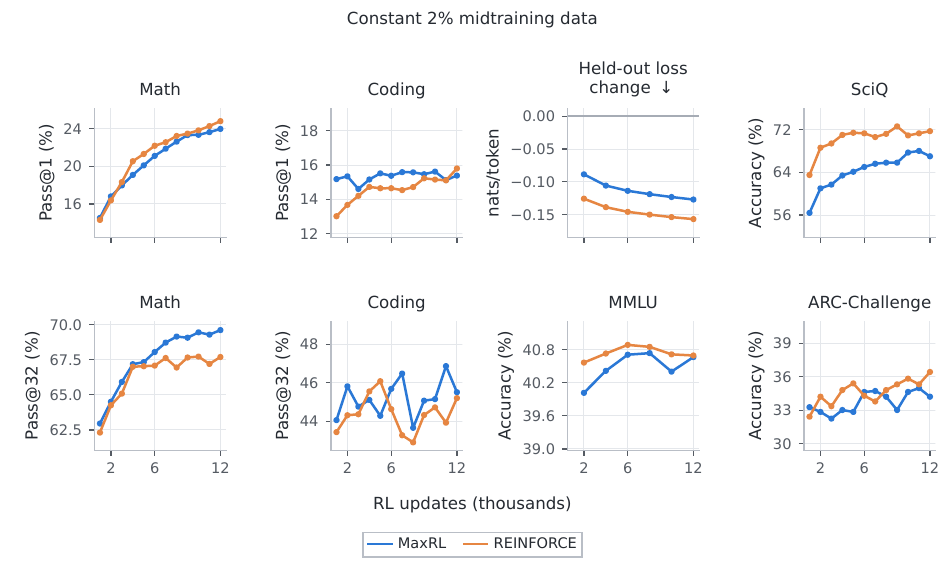}
    \caption{\textbf{\maxrl and \reinforce with a constant midtraining proportion of 2\%.}
    Training curves for the runs in \pref{tab:math-weighting}, evaluated using the protocol in \pref{app:math-design}.
    Both algorithms use unified training; blue curves use \maxrl and orange curves use \reinforce.}
    \label{fig:math-weighting-constant2}
\end{figure}

\paragraph{Constant 2\% mixing}
\pref{fig:math-weighting-constant2} gives the training curves for the comparison in \pref{tab:math-weighting}.
At the final checkpoint, \reinforce has higher Math accuracy than \maxrl, 24.81\% versus 23.99\%, while \maxrl has higher Math coverage.
The higher mathematical coverage of \maxrl and the lower held-out loss of \reinforce persist through the later part of training.

\begin{table}[H]
    \centering
    \small
    \setlength{\tabcolsep}{3pt}
    \begin{tabular*}{\linewidth}{@{\extracolsep{\fill}}lrrrrrrrr@{}}
        \toprule
        Method & \shortstack{Math\\pass@1} & \shortstack{Math\\pass@32} & \shortstack{Coding\\pass@1} & \shortstack{Coding\\pass@32} & $\Delta$loss $\downarrow$ & MMLU & SciQ & ARC-C \\
        \midrule
        \maxrl & 23.99 & 69.61 & 15.39 & 45.51 & $-0.127$ & 40.66 & 67.00 & 34.22 \\
        \reinforce & 24.81 & 67.70 & 15.80 & 45.21 & $-0.157$ & 40.69 & 71.70 & 36.43 \\
        \bottomrule
    \end{tabular*}
    \caption{\textbf{\maxrl and \reinforce with a constant midtraining proportion of 2\%.}
    Scores are percentages, except held-out loss change relative to the base model, measured in nats/token. ARC-C denotes ARC-Challenge.
    Training curves and comparisons across mixing settings appear in \pref{fig:math-weighting-constant2,fig:math-weighting}.}
    \label{tab:math-weighting}
\end{table}

\begin{figure}[H]
    \centering
    \includegraphics[width=\linewidth]{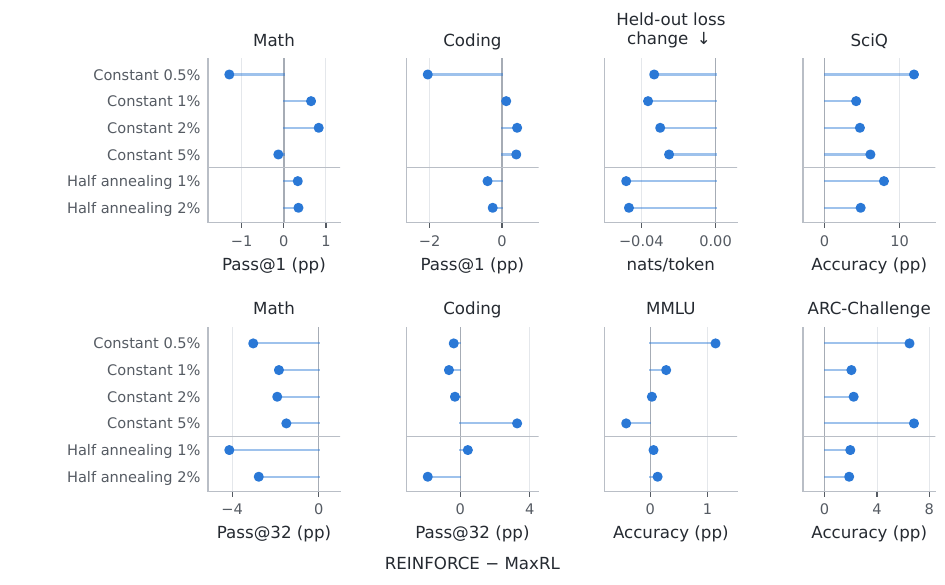}
    \caption{\textbf{Final \maxrl--\reinforce comparison across mixing settings.}
    Each difference subtracts the \maxrl result from the \reinforce result at the same mixing proportion and schedule.
    Positive values therefore indicate a higher value under \reinforce; for held-out loss, lower values are preferable.
    Differences are in percentage points for performance metrics and nats/token for loss.}
    \label{fig:math-weighting}
\end{figure}

\paragraph{Comparison across mixing settings}
Across the six settings in \pref{fig:math-weighting}, \maxrl consistently achieves higher mathematical coverage, while \reinforce yields lower held-out corpus loss and higher SciQ and ARC-Challenge accuracy.
The effects on mathematical accuracy and coding depend on the mixing proportion and schedule.

\pref{tab:math-weighting-full} reports the corresponding absolute results for both algorithms.

\begin{table}[H]
    \centering
    \small
    \setlength{\tabcolsep}{2.5pt}
    \begin{tabular*}{\linewidth}{@{\extracolsep{\fill}}llrrrrrrrr@{}}
        \toprule
        Mixing setting & Method & \shortstack{Math\\pass@1} & \shortstack{Math\\pass@32} & \shortstack{Coding\\pass@1} & \shortstack{Coding\\pass@32} & $\Delta$loss $\downarrow$ & MMLU & SciQ & ARC-C \\
        \midrule
        Constant 0.5\% & \maxrl & 24.27 & 68.32 & 16.51 & 46.38 & $-0.087$ & 39.99 & 62.00 & 32.34 \\
         & \reinforce & 22.98 & 65.29 & 14.45 & 46.00 & $-0.120$ & 41.13 & 73.90 & 38.82 \\
        \addlinespace[3pt]
        Constant 1\% & \maxrl & 23.86 & 68.25 & 15.49 & 44.89 & $-0.108$ & 40.54 & 66.10 & 33.53 \\
         & \reinforce & 24.51 & 66.41 & 15.60 & 44.24 & $-0.144$ & 40.82 & 70.30 & 35.58 \\
        \addlinespace[3pt]
        Constant 2\% & \maxrl & 23.99 & 69.61 & 15.39 & 45.51 & $-0.127$ & 40.66 & 67.00 & 34.22 \\
         & \reinforce & 24.81 & 67.70 & 15.80 & 45.21 & $-0.157$ & 40.69 & 71.70 & 36.43 \\
        \addlinespace[3pt]
        Constant 5\% & \maxrl & 23.85 & 69.68 & 15.93 & 45.17 & $-0.147$ & 40.77 & 65.10 & 30.12 \\
         & \reinforce & 23.72 & 68.18 & 16.32 & 48.44 & $-0.172$ & 40.35 & 71.20 & 36.95 \\
        \addlinespace[3pt]
        Half anneal 1\% & \maxrl & 25.01 & 68.99 & 18.65 & 46.07 & $-0.050$ & 40.10 & 62.90 & 32.42 \\
         & \reinforce & 25.34 & 64.85 & 18.25 & 46.51 & $-0.098$ & 40.16 & 70.80 & 34.39 \\
        \addlinespace[3pt]
        Half anneal 2\% & \maxrl & 24.80 & 68.49 & 18.85 & 46.93 & $-0.063$ & 40.14 & 64.30 & 31.83 \\
         & \reinforce & 25.15 & 65.72 & 18.59 & 45.06 & $-0.110$ & 40.26 & 69.10 & 33.70 \\
        \bottomrule
    \end{tabular*}
    \caption{\textbf{\maxrl and \reinforce across midtraining proportions and schedules.}
    Each pair of rows compares the two algorithms with the same configured proportion and schedule, giving the absolute final scores underlying the differences in \pref{fig:math-weighting}.
    Half annealing decreases the stated initial proportion linearly to zero halfway through training.
    Scores are percentages, except held-out loss change relative to the base model, which is measured in nats/token.
    ARC-C denotes ARC-Challenge.}
    \label{tab:math-weighting-full}
\end{table}

\clearpage
\subsection{Unified Training with Zero Mixing versus Staged RL}
\label{app:math-zero-mixing}

The 0\% mixing condition uses the same asynchronous trainer and reward weighting as the other unified runs, providing a control for the effect of adding midtraining data.
We compare this control with the staged RL reference to characterize the differences between the two implementations.
Both start from the same base model and train exclusively on mathematical RL data.
At zero mixing, the unified objective contains only the RL term, so the two approaches coincide when data generation and optimization are identical.
Our implementations differ in asynchronous execution and reward-baseline subtraction, as detailed below.

\paragraph{Asynchronous execution and data staleness}
The unified trainer consumes responses generated by earlier policies while generation continues in parallel, following \pref{sec:impl-buffer}.
Staged RL instead collects responses from the current policy before optimization.
Asynchronous execution removes this synchronization requirement, allowing the trainer to proceed with available responses at the cost of policy staleness.
We publish updated weights every four optimizer updates, but this interval is not a bound on the age of consumed data: the buffer retains the newest round from each generator even when its generating policy is more than four updates old.

To measure the staleness of available data, we inspected the newest completed round from each of eight generators at two points during training.
These measurements used constant or full-decay 0.2\% and 0.5\% mixing, with the same weight-publication interval and buffer recency settings as the comparison here.
The generating policies were 12--44 optimizer updates behind the latest logged trainer step.
This range describes the data available in the inspected rounds.
The policy versions of the examples actually consumed by the trainer were not logged, so these snapshots do not give the exact age distribution of the training data.

\paragraph{Baseline subtraction}
Unified training weights responses by $r/\hat p$, whereas staged \maxrl uses $(r-\hat p)/\hat p$, omitting numerical stabilizers here.
Replacing $\hat p$ by the population pass rate $p_\theta(x)>0$, the two weights yield the same expected on-policy gradient because
\begin{align*}
    \bbE_{y\sim\pi_\theta(\cdot\mid x)}\brk*{\nabla_\theta\log\pi_\theta(y\mid x)}=0.
\end{align*}
Thus, subtracting the prompt-dependent baseline leaves the population \maxrl gradient in \pref{sec:maxrl} unchanged.
The finite-sample updates nevertheless differ, and this on-policy identity does not apply unchanged to responses generated by an earlier policy.

\paragraph{Training speed}
Over 12000 updates, the production logs record approximately 3.7 seconds per update for unified training at 0\% mixing and 28.8 seconds for staged RL, a $7.8\times$ difference in update throughput.
These timings exclude launch and restart downtime and correspond to the implemented systems, which use 24 and 16 GPUs, respectively.
The asynchronous trainer can reuse available responses across updates, so this comparison measures optimization throughput rather than the rate of generating fresh rollouts.

\paragraph{Task performance and retention}
\pref{fig:math-dose} shows similar mathematical accuracy for zero mixing and staged RL: Math pass@1 is 24.75\% and 24.16\%, respectively.
Zero mixing has slightly higher mathematical coverage, with pass@32 of 68.88\% compared with 68.56\%, but a larger increase in held-out corpus loss, $0.074$ versus $0.047$ nats/token.
This comparison reflects the complete implementations and does not isolate the effects of staleness or baseline subtraction individually.

\clearpage
\subsection{Coding RL after Mathematical Training}
\label{app:real-continual}

\paragraph{Initialization}
We continue training on coding tasks using the staged mathematical model and four unified models trained with constant midtraining proportions of 0.5, 1, 2, and 5\%.
Each model starts from its final mathematical-training checkpoint.
Curve labels identify the models' mathematical training histories; all subsequently use the same coding RL configuration without midtraining data.
The comparison therefore tests how the preceding training affects adaptation to coding.

\paragraph{Data and training}
We use 20000 Python problems from KodCode-V1~\citep{xu2025kodcode},\footnote{\url{https://huggingface.co/datasets/KodCode/KodCode-V1}} comprising 10000 labeled medium difficulty and 10000 labeled hard difficulty.
The reward is one if the generated program passes all supplied unit tests and zero otherwise.
All models use the shared training settings in \pref{tab:real-coding-hyperparameters}.

\begin{table}[htbp]
\centering
\small
\renewcommand{\arraystretch}{1.08}
\begin{tabular}{@{}p{0.35\linewidth}p{0.61\linewidth}@{}}
\toprule
Setting & Value \\
\midrule
Algorithm & On-policy \maxrl with reward-baseline subtraction \\
Optimizer & Fresh AdamW \\
Adam $(\beta_1,\beta_2)$; $\epsilon$ & $(0.9,0.999)$; $10^{-8}$ \\
Weight decay; gradient clipping norm & 0; 1 \\
Learning rate / schedule & $10^{-6}$ / constant, no warmup \\
Prompts per update & 32 \\
Responses per prompt & 16 \\
Global batch size & 512 responses \\
Prompt + response limit (tokens) & 1024 \\
Temperature / top-$k$ / top-$p$ & 1 / 50 / 0.95 \\
KL penalty coefficient & 0 \\
\bottomrule
\end{tabular}
\caption{\textbf{Hyperparameters for coding and WebInstruct RL.}
All models use these same RL settings when trained on coding tasks or on WebInstruct after mathematical training.}
\label{tab:real-coding-hyperparameters}
\end{table}

\paragraph{Evaluation}
\pref{fig:real-continual} reports coding pass@1 and pass@16 at updates 0, 1000, 2000, 3000, and 4000.
Both metrics are estimated from 16 responses for each of the 164 HumanEval+ and 378 MBPP+ problems, using the EvalPlus tests and decoding settings in \pref{app:math-design}.
Step zero is the recorded evaluation of the mathematical checkpoint used to initialize coding RL; later points evaluate the corresponding coding checkpoints.
\pref{tab:real-continual-coding} reports the final coding scores.

To assess retention during this subsequent training, we also evaluate mathematics, general knowledge, and held-out midtraining loss.
For MATH-500 and GSM8K, we estimate pass@1 and pass@16 from 16 responses per problem.
MMLU uses zero-shot answer-choice likelihoods, while SciQ and ARC-Challenge use greedy decoding.
Held-out corpus loss uses the same equally weighted average over 24 components as in \pref{app:math-design}.

\begin{table}[htbp]
\centering
\small
\renewcommand{\arraystretch}{1.08}
\begin{tabular}{@{}lrrrr@{}}
\toprule
 & \multicolumn{2}{c}{HumanEval+} & \multicolumn{2}{c}{MBPP+} \\
\cmidrule(lr){2-3}\cmidrule(l){4-5}
Mathematical training & Pass@1 & Pass@16 & Pass@1 & Pass@16 \\
\midrule
Staged & 22.94 & 45.12 & 29.81 & 52.12 \\
Unified 0.5\% & 23.67 & 43.90 & 30.31 & 52.91 \\
Unified 1\% & 23.70 & 45.12 & 29.94 & 54.76 \\
Unified 2\% & 24.09 & 44.51 & 31.76 & 56.88 \\
Unified 5\% & \textbf{24.96} & \textbf{46.95} & \textbf{32.69} & \textbf{58.47} \\
\bottomrule
\end{tabular}
\caption{\textbf{Coding performance after subsequent RL.}
Scores are measured after 4000 coding RL updates and reported in percent.
Percentages in the model labels identify the midtraining proportions during mathematical training.
Both metrics use 16 responses per problem; bold indicates the highest score in each column.}
\label{tab:real-continual-coding}
\end{table}

\begin{figure}[!htbp]
    \centering
        \includegraphics[width=\linewidth]{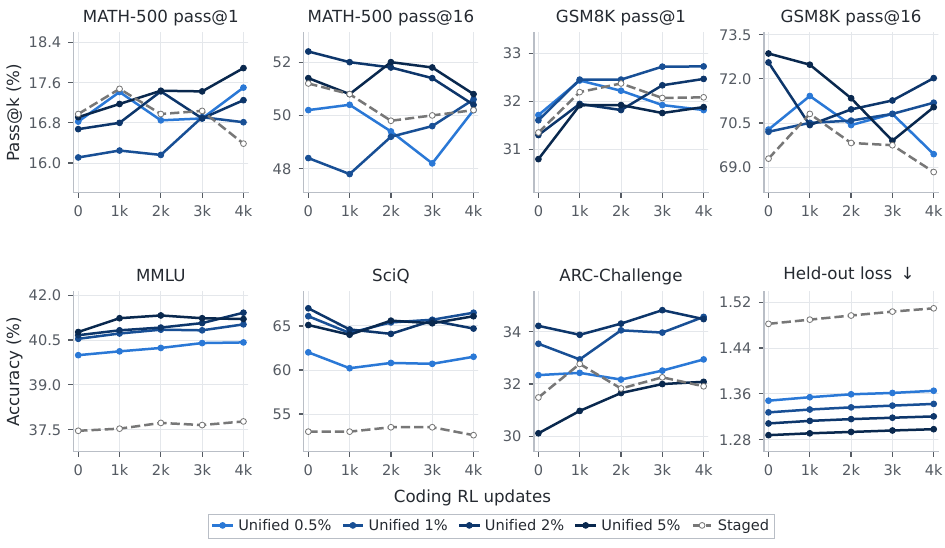}
    \caption{\textbf{Mathematical performance and retention during coding RL.}
    The initializations and coding updates are the same as in \pref{fig:real-continual}.
    Percentages denote constant midtraining proportions during the preceding mathematical training.
    Pass@1, pass@16, and benchmark accuracy are in percent; held-out loss is in nats/token.
    Mathematical accuracy remains close to its starting level, while changes in coverage vary across models.
    Held-out loss increases for every model.}
    \label{fig:real-continual-retention}
\end{figure}

\paragraph{Performance beyond coding}
\pref{fig:real-continual-retention} shows how mathematical performance and general knowledge change during coding RL.
Mathematical accuracy remains close to its starting level, while coverage can increase or decrease depending on the initialization.
All unified models retain higher MMLU and SciQ accuracy than staged training after 4000 updates.
Held-out corpus loss increases for every model, but the increase is smaller for the unified models, whose final loss also remains lower.

\clearpage
\subsection{WebInstruct RL after Mathematical Training}
\label{app:real-webinstruct}

\paragraph{Initialization and training}
To examine subsequent learning beyond coding, we train on WebInstruct starting from the staged mathematical model and the unified models trained with constant midtraining proportions of 2 and 5\%.
WebInstruct training is a separate continuation of mathematical training.
All three models use the same on-policy \maxrl procedure without midtraining data, with the hyperparameters in \pref{tab:real-coding-hyperparameters}.

\paragraph{Data and reward}
We randomly sample 20000 nonmathematical training questions from WebInstruct-verified and its unfiltered release~\citep{ma2026general}, covering fields such as physics, chemistry, business, and history.
We reserve disjoint sets of 512 validation and 1000 test questions.
The selected questions have numerical or free-form answers; decimal-number answer types are excluded.
The binary reward is one only when the extracted final answer is accepted by the General-Reasoner verifier and satisfies the answer-format checks.
Numerical answers must also pass an additional 1\% relative-error check.

\paragraph{Evaluation}
We monitor validation accuracy every 100 updates by generating 16 responses for each of the 512 validation questions and measuring the fraction of responses that receive reward one.
Separately, we evaluate test pass@1 and pass@64 at updates 0, 1000, 2000, and 3000, using 64 responses per question and the estimator in \pref{app:passk-estimator}.
Test sampling uses temperature 1, top-$k$ 50, top-$p$ 0.95, and up to 1024 response tokens within a 2048-token context.

Three test prompts exceed the 1024-token prompt budget and are excluded at every checkpoint, leaving a fixed set of 997 questions.
In addition to the overall scores, we report results separately for the 502 questions with numerical answers (integers, percentages, and fractions) and the 495 with free-form answers (strings, lists, and expressions).

\begin{figure}[!htbp]
    \centering
        \includegraphics[width=\linewidth]{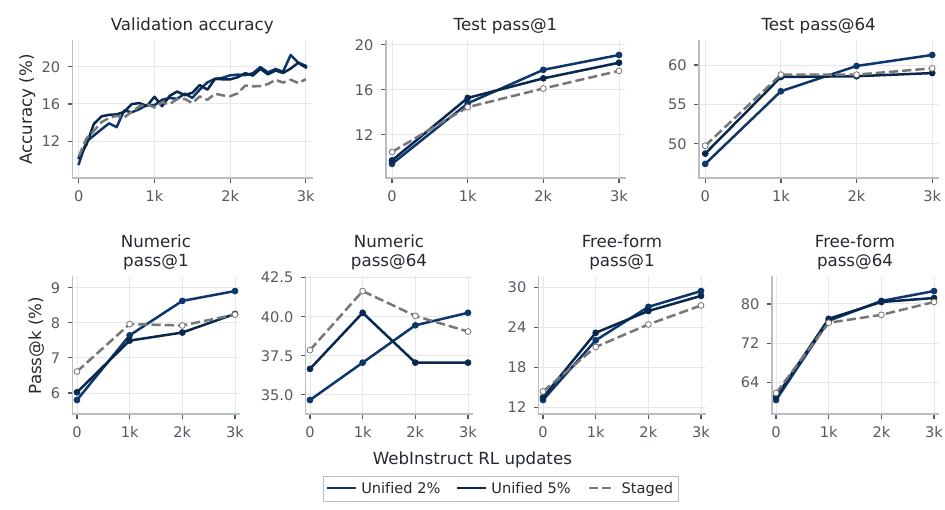}
    \caption{\textbf{WebInstruct RL after mathematical training.}
    Percentages identify the midtraining proportions during mathematical training; all models then use the same WebInstruct RL procedure without midtraining data.
    The top row shows validation accuracy and overall test performance; the bottom row separates test performance by numerical and free-form answers.
    All performance metrics are in percent.}
    \label{fig:real-webinstruct}
\end{figure}

\paragraph{Results}
\pref{fig:real-webinstruct} shows that both unified models begin with lower validation and test accuracy than staged training but achieve higher accuracy after 3000 updates.
The 2\% model also achieves higher test pass@64, whereas the 5\% model remains slightly below staged training on this metric.
The answer-type breakdown shows higher final free-form accuracy and coverage for both unified models than for staged training; on numerical answers, only the 2\% model achieves higher coverage.
These results provide further evidence that unified training can yield a better starting point for subsequent learning, with benefits that vary across answer types and metrics.

\clearpage
\section{Comparison with KL Regularization}
\label{app:kl-regularization}

Mixing midtraining data during RL may help preserve knowledge acquired before RL.
Here we ask whether a KL penalty that discourages changes from the base model can provide the same benefits.
Using the real-world data setting of \pref{sec:real}, we compare unified training with mathematical RL augmented by a sampled KL penalty.
The comparison measures both mathematical performance and retention, so that preserving the base model can be assessed alongside learning from RL.

\paragraph{Experimental setup}
All comparisons start from the same base model, use Math12K prompts for RL, and follow the same benchmark protocol.
For unified training, we use the constant-mixture runs in \pref{fig:math-dose}, whose midtraining proportions are 0.2, 0.5, 1, 2, and 5\%.
The KL runs receive no midtraining data during RL.
They use the asynchronous RL configuration and optimizer of the zero-mixing control, with 512 RL responses per update and 12000 updates.
Within this group, we vary only the penalty coefficient, $\beta\in\{0,0.001,0.01,0.1\}$; $\beta=0$ is the unregularized control.

\paragraph{KL penalty}
The reference policy is a frozen copy of the base model.
For each sampled RL response token, we subtract the current-policy log-probability from the reference log-probability and clip this difference to $[-20,20]$.
Writing the clipped difference as $u$, the token penalty is $e^u-u-1$.
The log-probabilities are computed over the full vocabulary.
We apply the penalty to tokens in both correct and incorrect responses, multiply it by $\beta$, and add it to the weighted token-loss numerator.
The resulting loss uses the same normalization as the RL term.

\begin{figure}[H]
    \centering
    \includegraphics[width=\linewidth]{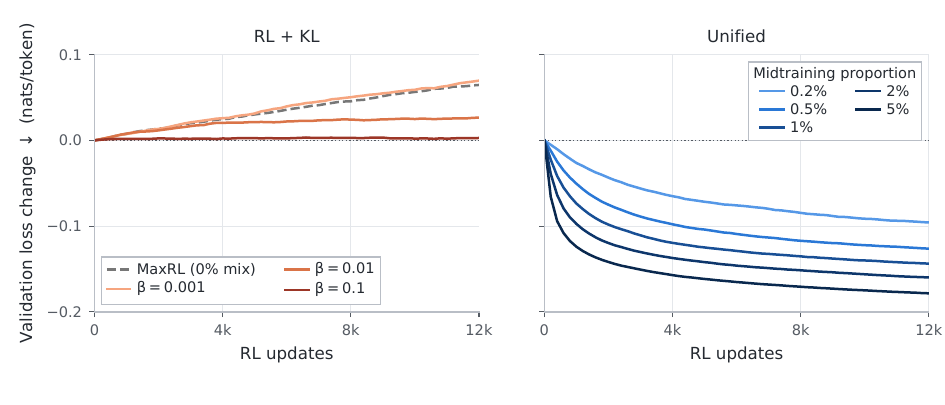}
    \caption{\textbf{Midtraining validation loss during KL-regularized RL and unified training.}
    Each curve shows the change from the common base-model value, measured on the trainer's fixed set of 512 held-out midtraining windows.
    This loss averages over tokens in the validation windows; it is distinct from the equally weighted average over corpus components in \pref{fig:kl-benchmark-curves}.
    Negative values indicate improvement over the base model.
    Curves show recorded evaluations without smoothing.}
    \label{fig:kl-training-loss}
\end{figure}

\begin{figure}[H]
    \centering
    \includegraphics[width=\linewidth]{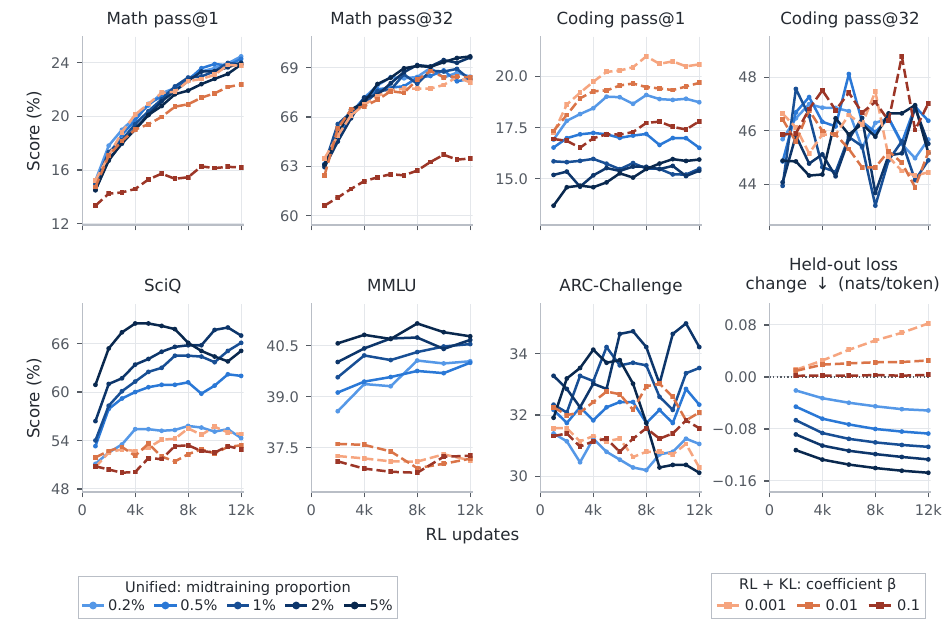}
    \caption{\textbf{Benchmark performance during KL-regularized RL and unified training.}
    Solid blue curves show unified training at five midtraining proportions; dashed orange curves show RL with three KL coefficients.
    Markers indicate recorded evaluations, joined without smoothing.
    Math and Coding average the two benchmarks in each domain, as defined in \pref{app:math-design}.
    Math pass@32 uses 256 responses per problem, and Coding pass@32 uses a separate 32-response evaluation.
    Held-out loss change gives equal weight to each of the 24 midtraining corpus components, with negative values indicating improvement over the base model.
    It is distinct from the token-averaged validation loss in \pref{fig:kl-training-loss}.}
    \label{fig:kl-benchmark-curves}
\end{figure}

\paragraph{Evaluation}
We use two held-out loss measurements, with different evaluation sets and averaging conventions.
During training, the trainer evaluates a fixed set of 512 held-out midtraining windows every 200 updates and at the final checkpoint.
This validation loss averages over the tokens in those windows and provides the curves in \pref{fig:kl-training-loss}.
The benchmark evaluation instead measures loss separately on the 24 midtraining corpus components and gives each component equal weight, following \pref{app:math-design}.
This component-averaged corpus loss appears alongside the task benchmarks in \pref{fig:kl-benchmark-curves}.
The two loss measurements should therefore be interpreted separately.

The task benchmarks use the evaluation protocol in \pref{app:math-design}.
The benchmark comparison includes all three nonzero KL coefficients and all five unified mixtures.
The training-time validation curves also include the unregularized control ($\beta=0$).

\clearpage
\paragraph{Results}
Increasing the nonzero KL coefficient reduces the deterioration in midtraining validation loss, as shown in \pref{fig:kl-training-loss}.
However, the strongest penalty ($\beta=0.1$) also substantially reduces mathematical accuracy and coverage relative to the weaker penalties (\pref{fig:kl-benchmark-curves}).
Unified training reduces the validation loss below the base-model value without this large reduction in mathematical performance.
The benchmark evaluation shows a similar retention benefit: at the final checkpoints, all five unified mixtures achieve higher MMLU accuracy than any of the KL-regularized runs.
The mixtures of 0.5\% and above also achieve higher SciQ accuracy; the 0.2\% mixture scores slightly below the weakest KL penalty on SciQ.

These benefits do not extend uniformly to every metric.
The weaker KL penalties yield higher coding accuracy, and the strongest KL penalty achieves the highest coding coverage among the tested settings.
ARC-Challenge accuracy shows no consistent advantage for either approach.
Nevertheless, none of the tested KL penalties reproduces the combination of mathematical performance, lower component-averaged corpus loss, and higher SciQ and MMLU accuracy obtained with unified training.
In this setting, regularizing toward the base policy therefore does not reproduce the full benefit of continuing to learn from midtraining data.

\clearpage
\section{Policy Gradient Objectives and Continuous Rewards}
\label{app:continuous-rewards}

We now show how the likelihood framework of \pref{sec:unified} recovers the policy gradients of \reinforce and \tailrl. For each method, we construct a target distribution and prompt weighting from a fixed generation policy and express the resulting update using the common likelihood loss $\cL(\theta;q)$. When the generation policy is the current model, these likelihood updates recover the corresponding policy gradients. \pref{tab:rl-objectives} summarizes these choices for on-policy updates.

\subsection{\reinforce}

For binary rewards, \reinforce and \maxrl fit the same distribution of correct completions. However, \reinforce weights each prompt by its success probability, giving greater weight to prompts on which the generation policy is more likely to succeed. 

We now show that, for general nonnegative rewards, \reinforce fits a reward-weighted target distribution, with each prompt weighted by its expected reward. For a fixed generation policy $\nu$ and rewards $r(x,y)\geq0$, let $p_\nu(x)\ldef\bbE_{y\sim\nu(\cdot\mid x)}[r(x,y)]$ denote the expected reward on prompt $x$. For $0<p_\nu(x)<\infty$, define the reward-weighted target
\begin{align*}
    \nu^+(y\mid x)=\frac{\nu(y\mid x)r(x,y)}{p_\nu(x)}.
\end{align*}
For binary rewards, this is $\nu$ conditioned on correctness.

\paragraph{The likelihood update}
We weight each prompt's likelihood loss by $p_\nu(x)$ to cancel the normalization in $\nu^+$:
\begin{align}
    \cL_{\mathrm{RF}}(\theta;\rho\circ\nu)
    \ldef -\bbE_{x\sim\rho}\,\bbE_{y\sim\nu^+(\cdot\mid x)}\brk*{p_\nu(x)\log\pi_\theta(y\mid x)}.
    \label{eq:reinforce-loss}
\end{align}
Differentiating gives
\begin{align*}
    -\nabla_\theta\cL_{\mathrm{RF}}(\theta;\rho\circ\nu)
    &=\bbE_{x\sim\rho}\,\bbE_{y\sim\nu^+(\cdot\mid x)}\brk*{p_\nu(x)\nabla_\theta\log\pi_\theta(y\mid x)}\\
    &=\bbE_{x\sim\rho}\,\bbE_{y\sim\nu(\cdot\mid x)}\brk*{r(x,y)\nabla_\theta\log\pi_\theta(y\mid x)}.
\end{align*}
Setting $\nu=\pi_{\theta_t}$ and evaluating at $\theta=\theta_t$ recovers the \reinforce policy gradient in \pref{eq:policy-gradient}. For a different generation policy, the likelihood update fits the reward-weighted target $\nu^+$ and need not match the current policy's gradient. 

\paragraph{Effective prompt distribution}
To express the weighted loss using the common loss $\cL(\theta;q)$, we absorb $p_\nu(x)$ into the prompt distribution.
Define
\begin{align*}
    Z\ldef\bbE_{x\sim\rho}[p_\nu(x)],\qquad
    \rho_{\mathrm{RF}}(x)\ldef\frac{\rho(x)p_\nu(x)}{Z},
\end{align*}
assuming $0<Z<\infty$. The weighted loss in \pref{eq:reinforce-loss} can therefore be written as $Z\cL(\theta;\rho_{\mathrm{RF}}\circ\nu^+)$. Here $\nu^+$ specifies the completion target, while $\rho_{\mathrm{RF}}$ reweights the original prompt distribution $\rho$ by the expected reward $p_\nu(x)$. Since the generation policy is fixed, $Z$ is independent of $\theta$ and scales the likelihood gradient. When combining this RL loss with the likelihood losses for pretraining, midtraining, or SFT, we absorb $Z$ into the RL source weight to preserve their relative weighting.

\paragraph{Baselines and reward normalization}
The likelihood formulation also applies to rewards with a finite lower bound, which can be shifted to be nonnegative. A fixed prompt-dependent baseline $b_t(x)$ contributes zero to the population on-policy gradient, since
\begin{align*}
    \bbE_{y\sim\pi_{\theta_t}(\cdot\mid x)}\brk*{\nabla_\theta\log\pi_\theta(y\mid x)}_{\theta=\theta_t}=0.
\end{align*}
Thus, for a lower bound $b(x)$ independent of $\theta$, the likelihood construction applies to the shifted reward $r(x,y)-b(x)\geq0$ without changing the \reinforce gradient. This baseline invariance relies on on-policy sampling and need not hold for uncorrected updates using data from an earlier policy.

Prompt-dependent reward scaling changes the relative prompt weights. Let $b_t(x)$ be a prompt-dependent baseline and $s_t(x)>0$ a prompt-dependent scale, both held fixed during the update. Subtracting the baseline contributes zero to the expected on-policy gradient. Thus, using $(r(x,y)-b_t(x))/s_t(x)$ gives the same expected
gradient as using $r(x,y)/s_t(x)$.

For nonnegative rewards, this gradient is recovered at $\theta=\theta_t$ by a likelihood update with target $\pi^+_{\theta_t}$ and prompt weight $p_{\theta_t}(x)/s_t(x)$. The target and prompt weight are defined using the original reward $r(x,y)$. The normalized reward itself need not be nonnegative. For binary rewards, choosing the population standard deviation $s_t(x)=\sqrt{p_{\theta_t}(x)(1-p_{\theta_t}(x))}$ as the scale gives prompt weight $\sqrt{p_{\theta_t}(x)/(1-p_{\theta_t}(x))}$ for $0<p_{\theta_t}(x)<1$, as in the population analysis of \grpo~\citep{tajwar2026maxrl}.

These identities concern expected on-policy gradients with fixed, prompt-dependent baselines and scales. Empirical advantages, policy ratios, clipping, and KL penalties in \ppo and \grpo need not preserve the fixed-target likelihood formulation~\citep{schulman2017ppo,shao2024deepseekmath}.

\subsection{Continuous Rewards with \tailrl}

\tailrl~\citep{ramasubramanian2026tailrl} extends \maxrl to continuous rewards by averaging over binary success events defined by reward thresholds. For rewards $r(x,y)\in[0,1]$, the probability of exceeding a threshold $\tau$ is
\begin{align*}
    q_\theta(x,\tau)\ldef\Pr_{y\sim\pi_\theta(\cdot\mid x)}\prn*{r(x,y)>\tau}.
\end{align*}
The objective averages the log-probability of success uniformly over thresholds:
\begin{align*}
    J_{\mathrm{Tail}}(\theta)\ldef\bbE_{x\sim\rho}\brk*{\int_0^1\log q_\theta(x,\tau)\,d\tau}.
\end{align*}
\paragraph{A target at each reward threshold}
For a fixed generation policy $\nu$, let $q_\nu(x,\tau)$ be its probability of exceeding threshold $\tau$. Assume that $q_\nu(x,\tau)>0$ for $\tau<1$ and that the \tailrl objective is finite.
The corresponding success-conditioned target is
\begin{align*}
    \nu^+_\tau(y\mid x)\ldef\frac{\nu(y\mid x)\,\mathbf{1}\{r(x,y)>\tau\}}{q_\nu(x,\tau)}.
\end{align*}
\paragraph{Combining the targets}
The likelihood target combines these conditional distributions using the same uniform weighting over thresholds as the objective:
\begin{align*}
    \nu_{\mathrm{Tail}}(y\mid x)\ldef\int_0^1\nu^+_\tau(y\mid x)\,d\tau.
\end{align*}
This mixture is a probability distribution because each conditional target is normalized. The threshold is integrated out, so the model continues to condition only on the prompt $x$.
Fitting the mixture gives
\begin{align*}
    -\nabla_\theta\cL(\theta;\rho\circ\nu_{\mathrm{Tail}})
    &=\bbE_{x\sim\rho}\,\bbE_{y\sim\nu_{\mathrm{Tail}}(\cdot\mid x)}\brk*{\nabla_\theta\log\pi_\theta(y\mid x)}\\
    &=\bbE_{x\sim\rho}\brk*{\int_0^1\bbE_{y\sim\nu^+_\tau(\cdot\mid x)}\brk*{\nabla_\theta\log\pi_\theta(y\mid x)}\,d\tau}.
\end{align*}
Setting $\nu=\pi_{\theta_t}$ and evaluating at $\theta=\theta_t$ recovers $\nabla_\theta J_{\mathrm{Tail}}(\theta_t)$. Thus, the corresponding likelihood update uses the joint target distribution $q=\rho\circ\nu_{\mathrm{Tail}}$. For a different generation policy, the likelihood update fits the mixture target induced by that policy and need not match the current policy's \tailrl gradient.

\paragraph{Binary rewards and finite samples}
For binary rewards, every threshold $\tau\in[0,1)$ selects the same success event, so the objective and target reduce to those of \maxrl.

The derivation of the \tailrl update concerns population gradients. The finite-sample \maxrl identity in \pref{sec:impl-generator} applies specifically to binary rewards. For continuous rewards, estimation must also account for the different reward thresholds. \citet{ramasubramanian2026tailrl} derive the corresponding estimators and the finite-order \tailrl objectives whose gradients they estimate.

\end{document}